%% file: main.tex
\PassOptionsToPackage{table}{xcolor}
\documentclass[11pt, a4paper, logo, copyright, nonumbering]{cls/Sci}

\usepackage[numbers, sort&compress, square]{natbib}
\usepackage{url}
\usepackage{amsmath}
\usepackage[table]{xcolor}
\usepackage{multirow}
\usepackage{makecell}
\usepackage{booktabs}
\usepackage{amsfonts}
\usepackage{nicefrac}
\usepackage{graphicx}
\usepackage{enumitem}
\usepackage{pifont}
\usepackage{pdflscape}
\usepackage{array}
\usepackage{caption}
\usepackage{subcaption}
\usepackage{wrapfig}
\usepackage{float}
\usepackage{placeins}
\usepackage{listings}
\usepackage{adjustbox}
\usepackage{fontawesome5}
\usepackage{textcomp}
\usepackage{etoolbox}
\usepackage{amsmath}  
\usepackage{amssymb}   
\usepackage{amsthm}  
\usepackage[most]{tcolorbox}
\usepackage{fvextra}

\newcommand{\cmark}{\textcolor{green}{\ding{51}}}
\newcommand{\xmark}{\textcolor{red}{\ding{55}}}

\newtheorem{theorem}{Theorem}[section]
\newtheorem{assumption}{Assumption}[section]

\lstdefinestyle{python}{
  language=Python,
  basicstyle=\ttfamily\small,
  keywordstyle=\color{blue}\bfseries,
  stringstyle=\color{red!70!black},
  commentstyle=\color{green!50!black}\itshape,
  showstringspaces=false,
  breaklines=true,
  frame=single,
  framerule=0.4pt,
  rulecolor=\color{black!30},
  backgroundcolor=\color{black!3},
  numbers=left,
  numberstyle=\tiny\color{black!40},
  numbersep=6pt,
  tabsize=4,
  literate=
    {->}{{\textrightarrow}}2
    {None}{{\textbf{None}}}1
    {True}{{\textbf{True}}}1
    {False}{{\textbf{False}}}1,
}
\title{Belief-State Management Meets Rubric Rewards for Agentic Forecasting}

\author{
{\normalfont\mdseries
Jingyi Chai\textsuperscript{1},
Yun Liao\textsuperscript{1},
Hao Lu\textsuperscript{1},
Kemeng Zhang\textsuperscript{1},
Bingyang Zheng\textsuperscript{1},
Yujie Ge\textsuperscript{1},
Xiangrui Liu\textsuperscript{1},
Linfeng Zhang\textsuperscript{1},
Siheng Chen\textsuperscript{1,\ensuremath{\dagger}}
}
\\
{\normalfont\bfseries
\textsuperscript{1}Shanghai Jiao Tong University
}
\\
{\normalfont\mdseries
\textsuperscript{\ensuremath{\dagger}}Corresponding author: sihengc@sjtu.edu.cn
}

}

\begin{document}

\begin{abstract}
Current large language model (LLM) agents still struggle to forecast real-world events. This stems from two primary limitations: i) standard ReAct harness accumulates reasoning and evidence within one growing context, and ii) outcome-based rewards in forecasting lead to a slow and noisy signal that cannot distinguish sound reasoning from lucky guesses. To address these, we introduce a forecasting training framework including harness design and post-training. 
For the harness side, we propose Proactive Belief-State Management (PBM), equipping agents with a fold tool to compress lengthy context into compact belief states and reset the context for subsequent exploration. 
For the training side, we develop SEIPH, a five-dimensional, outcome-blind rubric, providing process-quality rewards. Using the rubric-based reward, we post-train a 35B model with PPO and GRPO, yielding DeepForecaster-PPO and DeepForecaster-GRPO. 
Theoretically, we prove that idealized fold-based stopping converges and calibrated local rewards reduce segment-level gradient variance.
Empirically, DeepForecaster-PPO (35B) achieves the best Brier score on both ForecastBench and Polymarket-Sim, outperforming frontier agents based on DeepSeek-V4-Pro (1.6T), Kimi-K2.6 (1T), and GPT-5.5. DeepForecaster-GRPO ranks second on both FutureX metrics. Ablations reveal that PBM also improves the untrained base model without any training. Calibration analysis shows that DeepForecaster-PPO makes more accurate high-confidence predictions than Qwen3.6 and GPT-5.5. Trajectory analysis further shows that forecasts generally improve from the first to the final belief state as agents gather evidence and revise their judgments.
\end{abstract}

\maketitle

\input{secs/intro}
\input{secs/related}
\input{secs/method}
\input{secs/results}
\input{secs/conclusion}

\bibliography{main}

\appendix
\input{secs/app}

\end{document}

%% file: secs/intro.tex
\begin{figure}[b]
    \centering
    \includegraphics[width=1\linewidth]{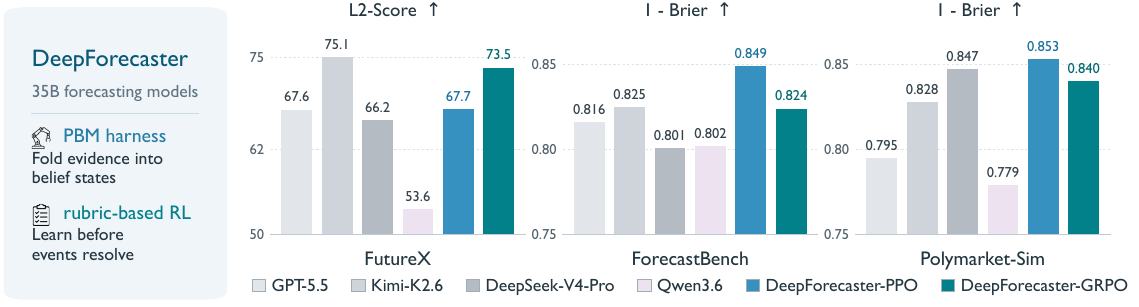}
    \caption{DeepForecaster models and forecasting performance. DeepForecaster-PPO achieves the best Brier scores among the evaluated agents on ForecastBench and Polymarket-Sim, while DeepForecaster-GRPO ranks second on FutureX L2-Score.}
    \label{fig:teaser}
\end{figure}

\section{Introduction}
Forecasting real-world events is an important but challenging task for large language model (LLM) agents~\citep{forecast_survey,futurex}. A forecasting agent needs to search for timely evidence from diverse sources, reason about competing scenarios under uncertainty, and produce a calibrated probability before the event resolves in the real world~\citep{atm}. Since live questions have no available label at prediction time, forecasting provides a demanding and contamination-free testbed for LLM agents~\citep{consistencychecks, beyondbinary, forecastbench}.

\begin{figure}[h]
    \centering
    \includegraphics[width=1\linewidth]{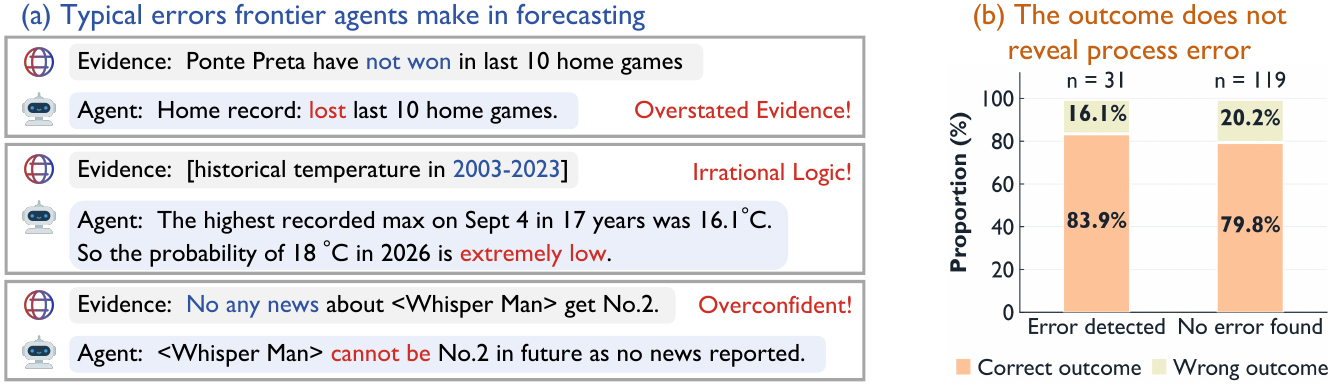}
    \caption{Analysis of forecasting errors in frontier agents and their forecast outcomes. \textbf{(a)} Three categories of manually verified errors derived from 150 trajectories across 30 live Polymarket questions. \textbf{(b)} Outcome accuracy with and without process error. Trajectories with detected errors predict the correct outcome at a rate comparable to error-free trajectories. This indicates that forecast outcomes cannot reveal any process error.}
    \label{fig:motivation}
\end{figure}

Despite rapid progress of LLMs in deep search and long-horizon reasoning, our experiments show that current advanced agents still perform poorly on forecasting. They often overstate evidence, treat forecasting as a search task, exhibit irrational logic, and update probabilities without sufficient support. Figure~\ref{fig:motivation}(a) illustrates three error examples of advanced agents in forecasting.
These failures reveal a fundamental mismatch between forecasting and the standard agent training paradigm, including the agent harness and training signal.

On the harness side, standard ReAct-style agents accumulate every tool observation and reasoning in a single growing context~\citep{react, agentfold}. As the trajectory lengthens, stale evidence and redundant search results crowd the context window and make it hard to integrate evidence into a coherent forecast. On the training side, recent works build live forecasting environments and train agents with outcome-based reinforcement learning~\citep{futureworld, openforecastor, teachfuture, outcome_forecast, future_as_label, forecastingrl}. However, outcome rewards have two drawbacks: (i) Resolution times are uncertain and often span weeks or months, so rewards are slow and costly to obtain. (ii) Outcome feedback is inherently noisy, as shown in Figure~\ref{fig:motivation} (b). Flawed logic can produce accurate outcome, and vice versa~\citep{worldreasoner}. Consequently, optimizing solely for outcome correctness risks rewarding spurious correlations rather than robust reasoning capabilities.

To fill this gap, we propose an LLM training framework for forecasting that combines an agent harness with rubric-based reinforcement learning (RL).
For the harness, we propose \emph{Proactive Belief-State Management} (PBM). PBM equips the agent with a "fold" tool to synthesize raw observations into a compact belief state, which defines current probability estimates, gathered evidence, and next search directions. 
After completing one line of evidence, the agent updates its belief state via the fold tool. The harness then discards the raw context and resets to the original question and the updated belief state. Consequently, the agent can pursue multiple independent lines of evidence without context overflow.
Building on this harness, we use supervised fine-tuning (SFT) to teach a 35B model to invoke the fold. To further optimize forecasting, we introduce the \emph{SEIPH rubric}, a five-dimensional, outcome-agnostic reward system. SEIPH evaluates each PBM segment by assessing search targeting, evidence verification, and the reasoning behind probability estimates and updates, without waiting for the event to resolve. It explicitly penalizes logical flaws like unsupported probability jumps or hallucinated evidence. Using this rubric as the reward signal, we train two model variants, \emph{DeepForecaster-PPO} and \emph{DeepForecaster-GRPO}, via PPO~\citep{ppo} and GRPO~\citep{grpo}, respectively.

We evaluate our framework on three live forecasting benchmarks, FutureX, ForecastBench, and Polymarket-Sim, comparing with GPT-5.5, DeepSeek-V4-Pro, Kimi-K2.6, and other strong baselines. DeepForecaster-PPO, with only 35B total parameters, achieves the best Brier score on both ForecastBench (0.151) and Polymarket-Sim (0.147), outperforming DeepSeek-V4-Pro (1.6T), Kimi-K2.6 (1T), and GPT-5.5. Experiments show that PBM improves the untrained ReAct baseline, and that our SFT and segment-level RL pipeline further improves forecasting performance. Calibration analysis shows that DeepForecaster-PPO maintains high confidence while making fewer incorrect predictions than Qwen3.6 and GPT-5.5. Behavioral analysis further indicates that both RL models fold more frequently, use fewer tool turns per segment than the SFT model, and generate longer reasoning traces. This pattern suggests that RL encourages more focused search cycles, with more reasoning between tool interactions and more frequent belief-state updates.

Our contributions are as follows:

(i) We introduce PBM, a forecasting harness where agents proactively fold context into structured belief states, which effectively prevents context overflow and enables focused search cycles.

(ii) We propose SEIPH, an outcome-blind rubric and hard-gate reward mechanism that provides segment-level supervision and penalizes critical violations, enabling robust PPO/GRPO training.

(iii) We prove that idealized fold-based stopping converges and calibrated local rewards reduce segment-level gradient variance. Experiments show that DeepForecaster outperforms larger advanced LLM agents in forecasting performance.

%% file: secs/related.tex
\section{Related Work}

\subsection{Harness for Long-Horizon Agents}

\textbf{General agent harness.}
Agent harnesses improve long-horizon LLM agents by organizing tool use, memory, and interaction histories~\citep{harness_survey}. A growing line of work focuses on context management for long-horizon tasks, including context engineering, structured trajectory folding, and training with context compression~\citep{contextengineering,agentfold,foldgrpo,contexttool}. However, general-purpose memory and compression methods do not explicitly track the uncertainty, competing hypotheses, and probability updates required for forecasting.

\textbf{Forecasting-specific harness.}
Existing forecasting harnesses mainly follow either a multi-agent or a single-agent design. Multi-agent forecasting works assign separate roles for search, evidence synthesis, and forecasting, enabling parallel exploration~\citep{miroflow,infodelphi,atm}. Single-agent systems keep the research loop unified with an explicit forecasting state. For example, the Bayesian Linguistic Forecaster~\citep{blf} maintains probability estimates and evidence summaries at every tool call, and Milkyway~\citep{mickyway} revisits unresolved questions and updates its forecasting state as new evidence arrives. 
While these demonstrate the importance of continuous state tracking, they ignore cumulative search results in one forecasting trajectory.

\subsection{Rewards in Reinforcement Learning for Forecasting}

\textbf{Outcome-based rewards.}
Recent work trains forecasting agents with realized events, market outcomes, or delayed labels~\citep{futureworld,openforecastor,teachfuture,outcome_forecast,future_as_label,forecastingrl}. These rewards are directly tied to the final correctness, but they are often delayed and noisy. A realized outcome is also an incomplete measure of reasoning quality, since a lucky guess may be correct and a well-supported forecast may be wrong~\citep{worldreasoner}. Outcome-only rewards therefore provide sparse credit assignment and may reinforce shortcuts.

\textbf{Rubric-based rewards.}
Recent work explores explicit rubrics as an alternative source of reward for open-ended agent tasks. Echo selects domain-specific behavioral rubrics for forecasting~\citep{echo}, while FinanceHarness uses expert-designed rubrics to evaluate financial research processes~\citep{financeharness}. Other methods study adaptive or reusable rubrics for agent reward learning~\citep{rar,zhang2026chaining}. These works suggest that rubrics can provide timely and interpretable feedback, but leave open how to design forecasting-specific criteria and assign reliable credit across long research trajectories.

%% file: secs/method.tex
\section{Problem Formulation}

We formulate forecasting as a long-horizon agent task. Each question $q_i$ describes an unresolved event, requiring a calibrated probability $p_i \in [0,1]$ before the outcome is known. The agent achieves this through extensive reasoning: searching for information, evaluating scenarios, and continuously updating its beliefs. These inherently long trajectories are structurally partitioned into distinct segments via a fold tool. Specifically, for question $q_i$, the model generates $N_i$ sibling trajectories. Let $F_{i,n}$ denote the number of Belief State Folds in the $n$-th trajectory, partitioning the sequence into $K_{i,n} = F_{i,n} + 1$ segments. Crucially, training is entirely outcome-blind: the ground-truth $a_i \in \{0,1\}$ is strictly excluded, ensuring our process reward evaluates reasoning independently of the final resolution. For notational brevity, we hereafter omit subscripts $i$ and $n$, denoting the question as $q$, the total segments as $K$, and the belief after segment $k$ as $D_k$.

The $k$-th local semantic episode is defined as$$\tau_k = (q, D_{k-1}, \xi_k, O_k), \quad k=1,\ldots, K$$where $D_0$ indicates an empty initial belief state, and the variable $\xi_k$ encapsulates the search, visit, observation, and reasoning process within the $k$-th stage. The stage output $O_k$ is defined as
$$O_k = \begin{cases}  D_k, & k < K \\  (D_K, p_K), & k = K  \end{cases},$$
yielding the updated belief state $D_k$ for intermediate stages, and appending the final probability prediction $p_K$ at the terminal stage.

\section{Method}\label{sec:method}

\begin{figure}[t]
    \centering
    \includegraphics[width=1\linewidth]{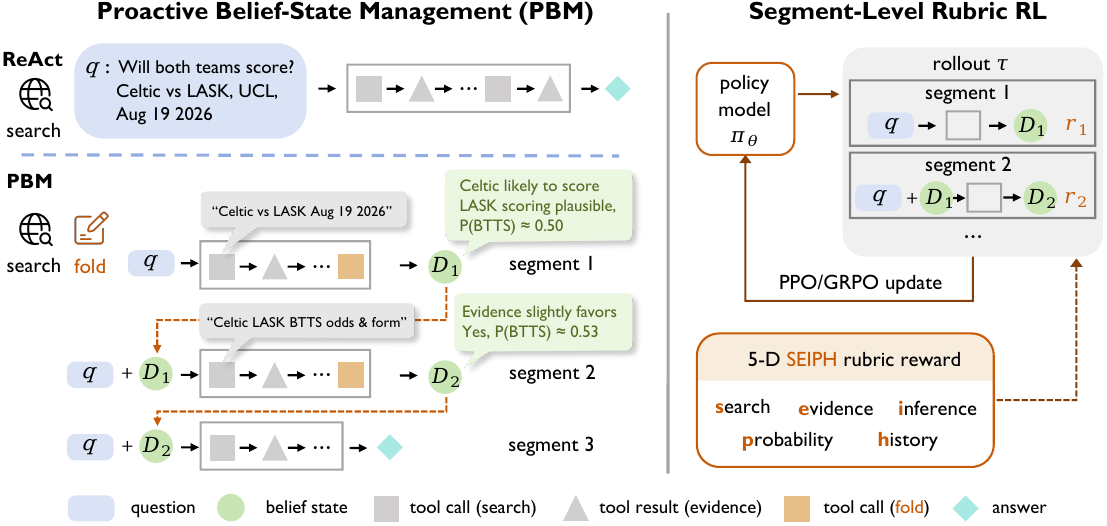}
    \caption{Overview of our forecasting training framework.
    \textbf{Left.} Proactive Belief-State Management.
    A ReAct agent maintains one growing context throughout the trajectory. PBM instead divides it into semantic segments separated by belief-state folds.
    Each fold compresses exploration into a structured belief state $D_k$ that initializes the next segment.
    \textbf{Right.} Segment-Level Rubric RL.
    The policy model generates rollouts composed of multiple segments. Each segment receives an independent reward $r_k$ from the five-dimensional SEIPH rubric, which drives PPO or GRPO policy updates.}
    \label{fig:framework}
\end{figure}

\subsection{Proactive Belief-State Management}

Forecasting requires an agent to explore many possible information paths while keeping a stable view of the current event. A plain ReAct-style trajectory keeps all raw observations and reasoning steps in a single growing context window. As the trajectory lengthens, the context becomes crowded with stale evidence, duplicated search results, and intermediate thoughts, making it progressively harder for the model to integrate new information into a coherent forecast.

Therefore, we introduce Proactive Belief-State Management (PBM) as the harness for forecasting agents. As illustrated in Figure~\ref{fig:framework} (Left), PBM augments the standard search and visit tools with a \emph{fold} tool. Unlike periodic or externally triggered compression, the agent itself decides when to invoke fold, typically after one line of inquiry has produced useful evidence, or when the agent wants to switch to a different
research direction. When calling fold, the agent writes a structured belief state that contains four components: (i) \textbf{Foresight}, recording the current probability distribution over prediction options, the key conclusions drawn so far, and the rationale linking evidence to probabilities; (ii) \textbf{Evidence}, accumulating the concrete facts, figures, and sources observed through the tools; (iii) \textbf{Search history}, summarizing what was searched and found in each past segment; and (iv) \textbf{Todo}, listing the remaining high-yield research directions. Among these, foresight and todo are overwritten at each fold to reflect the agent's latest belief, while evidence and search history are appended to preserve the full record. The agent supplies these updates as structured arguments to the fold tool, and the harness merges them into the existing belief state.

After each fold, the harness discards all raw search and visit messages from the current segment and resets the context to the original question together with the rendered belief state. The agent then continues exploration from this compact starting point, pursuing multiple independent lines of evidence without context overflow. This mechanism also creates natural segment boundaries for training: each fold defines a local episode whose process quality can be evaluated independently, without waiting for the final event outcome. When the agent judges that further research is unlikely to change its forecast meaningfully, it returns a final prediction without another fold.

\subsection{SEIPH Rubric for Agentic Forecasting}

We introduce SEIPH to evaluate each forecasting segment using only information available at that point. The judge reads the question, visible search and reasoning steps, and inherited or updated belief states.
SEIPH has five dimensions as follows. (i) \textbf{Search} $S_k$ assesses whether queries target decision-relevant information, cover unresolved factors, and stop when further retrieval adds little value. (ii) \textbf{Evidence} $E_k$ assesses source reliability, relevance to the target event, and treatment of unverified or repeated reports. (iii) \textbf{Inference} $I_k$ assesses separation of facts from assumptions, consideration of competing scenarios, and the link from evidence to forecast. (iv) \textbf{Probability} $P_k$ assesses whether probabilities have a defensible basis and whether updates reflect the evidence. Retaining a forecast is valid when further research provides no reason to revise it. (v) \textbf{History} $H_k$ assesses whether the belief state preserves evidence of support and conflict, current probabilities, and is free from hallucinations.
Within each dimension $d \in \{S, E, I, P, H\}$, three criteria are scored on a discrete scale $x_{k,d,j} \in \{0, 1, 2\}$. The normalized score for dimension $d$ is calculated as:
$$z_{k,d}=\frac{\sum_{j=1}^{3}x_{k,d,j}}{6}\in[0,1].$$

We average the five dimension scores as
$$U_k=\frac{1}{5}\sum_{d\in\{S,E,I,P,H\}}z_{k,d}\in[0,1].$$
Without a visible belief state, we exclude History and average the remaining four scores.

\subsection{Segment-Level Rubric-Based Reinforcement Learning}

Leveraging this SEIPH rubric, we design a rubric-based reward mechanism, as illustrated in Figure~\ref{fig:framework} (Right). For any segment $k \le K$, the final reward $r_k$ combines the normalized rubric score $U_k$ with a hard validity gate $G_k \in \{0, 1\}$. This gate acts as a strict filter for fatal errors: $G_k = 0$ if the segment triggers critical violations (e.g., fabricated facts, belief state hallucination or distortion, target time drift, outcome leakage, or invalid probabilities); otherwise, $G_k = 1$. The reward heavily penalizes these errors while encouraging valid reasoning:
$$r_k=G_k U_k-(1-G_k),$$
which is equivalent to:
$$r_k=\begin{cases}U_k,&G_k=1\\-1,&G_k=0\end{cases}.$$
Minor behavioral flaws, such as ordinary tool errors or shallow search depth, are naturally penalized within the SEIPH score $U_k$. Avoiding arbitrary linear deductions for these common issues ensures the reward signal remains clean and focused.

Each segment is scored using its local search and reasoning steps and the belief state available at its boundary, including the final segment that produces the forecast. The resulting reward provides the training signal for the policy-generated tokens within that segment. Using these segment-level process rewards, we train our DeepForecaster-PPO and DeepForecaster-GRPO via PPO \citep{ppo} and GRPO \citep{grpo}. Integrating this fine-grained rubric into robust RL frameworks incentivizes the agent to autonomously refine its search, reasoning, and belief state management, ultimately converging into a highly calibrated forecasting agent.

\subsection{Theoretical Analysis: Information Preservation and Credit Assignment}

We analyze how belief-state folding preserves evidence under limited context and how local rubric rewards improve segment-level credit assignment. Each segment ends with either a fold or the final forecast. Assumptions, proofs, and extensions to imperfect rubrics are provided in Appendix~\ref{sec:theorem}.

\begin{theorem}[Convergence of Fold-Based Stopping under Mirror Descent]
    Under diminishing information gains and a positive segment cost, suppose the optimal stopping point \(\tau^*\) is unique, with reward gap \(\Delta>0\). For the KL-regularized policy update defined in Appendix A and an initialization satisfying \(p_0(\tau^*)>0\),
\[
\mathbb{E}_{p_T}[|K-\tau^*|]
\leq
(H-1)
\frac{1-p_0(\tau^*)}{p_0(\tau^*)}
\exp\!\left(
-\Delta\sum_{t=0}^{T-1}\eta_t
\right).
\]
Consequently, if  \(\sum_t\eta_t=\infty\), the stopping policy converges to the optimal cutoff \(\tau^*\),meaning that \(p_T(\tau^*)\to 1\). Moreover, \(p_T(\tau^*)\) is nondecreasing in \(T\).
\end{theorem}

This result shows how segment-level stopping decisions can learn to terminate research when further information no longer justifies its cost. The analysis uses an idealized mirror-descent formulation of PPO-style policy optimization. The guarantee does not cover practical clipping or approximate updates.

\begin{theorem}[Variance Reduction through Local Rubric Supervision]
    For a fixed policy, let \(g_k^{\mathrm{global}}\) and \(g_k^{\mathrm{local}}\) be segment \(k\)’s gradient contributions using terminal reward and a calibrated local rubric, respectively. Then
\[
\mathbb{E}[g_k^{\mathrm{local}}]
=
\mathbb{E}[g_k^{\mathrm{global}}],
\]\[
\operatorname{Var}(g_k^{\mathrm{local}})
=
\operatorname{Var}(g_k^{\mathrm{global}})
-
\mathbb{E}\!\left[
\|z_k\|^2\operatorname{Var}(R\mid\mathcal{F}_{k+1})
\right],
\]where \(z_k\) is the segment policy score, \(R\) the terminal reward, and \(\mathcal{F}_{k+1}\) the history through segment \(k\); vector variance denotes covariance trace.
\end{theorem}

Local rubric supervision bypasses reliance on terminal outcomes for segment-level learning. Under the stated calibration assumption, it preserves the expected policy-gradient contribution while reducing noise from subsequent outcomes. Appendix~\ref{sec:theorem} quantifies when this advantage persists under rubric approximation error.

%% file: secs/results.tex
\section{Experimental Results}
\subsection{Setup}
\textbf{Implementation.} We fine-tune the open-source Qwen3.6-35B-A3B~\citep{qwen36_35b_a3b} on PBM trajectories to obtain Qwen3.6-PBM-SFT. Starting from this checkpoint, we train DeepForecaster-PPO and DeepForecaster-GRPO using PPO and GRPO, respectively, with rubric rewards from GLM-5.2~\citep{glm52}. See more training details in Appendix~\ref{app:training}.

\textbf{Benchmarks.} 
We evaluate different agents on three live forecasting benchmarks.
FutureX~\citep{futurex} collects questions about upcoming events from 195 websites across politics, economics, technology, sports, and other domains, and scores predictions after the events resolve. ForecastBench~\citep{forecastbench} releases questions from prediction markets and real-world time series every two weeks and reports Brier scores for models and human forecasters. Polymarket-Sim uses questions from Polymarket. Our evaluation covers 126 resolved questions from August 31 to September 5, 2026. See more construction details in Appendix~\ref{app:polysim}.

\textbf{Baselines.} We compare against five frontier models, GPT-5.5~\citep{gpt55}, GLM-5.2~\citep{glm52}, Kimi-K2.6~\citep{kimik26}, DeepSeek-V4-Pro and DeepSeek-V4-Flash~\citep{dpskv4}, as well as the base Qwen3.6-35B-A3B~\citep{qwen36_35b_a3b}. Qwen3.6-PBM is the untrained base model in the PBM harness. The ReAct harness~\citep{react} provides two web tools, \texttt{search} for Google queries and \texttt{visit} for reading a page through a summarizer.
Our PBM harness provides an additional fold tool, which writes the belief state and triggers the context reset described in Section~\ref{sec:method}.

\textbf{Metrics.}
For FutureX, L1-Score is exact-answer accuracy and L2-Score is set-level F1, both reported as percentages and averaged across three weeks. For ForecastBench and Polymarket-Sim, let $p_i$ denote the forecast probability and $a_i \in \{0,1\}$ the resolved outcome. We report Brier score, $\frac{1}{N}\sum_{i=1}^{N}(p_i-a_i)^2$, where lower is better.
For Polymarket-Sim, we also report Alpha and Accuracy. With $m_i$ denoting the market probability at the forecast submission deadline, $\text{Alpha}=\frac{1}{N}\sum_{i=1}^{N}[(m_i-a_i)^2-(p_i-a_i)^2]$, where higher is better, and zero indicates equal market and agent Brier scores. Accuracy is the percentage of questions for which the predicted outcome matches the resolved outcome, predicting Yes if $p_i \geq 0.5$ and No otherwise. 

\subsection{Main Results}
\begin{table}[t]
\centering

\setlength{\tabcolsep}{2.5pt}
\caption{Forecasting performance on FutureX, ForecastBench, and
Polymarket-Sim. FutureX L1 and L2 scores are averaged over September Weeks 1-3. ForecastBench uses 229 resolved questions as of September 16, 2026. Polymarket-Sim covers 126 resolved questions from August 31 to September 5, 2026. The \textbf{best} is in bold, and \underline{second-best} is underlined, excluding the Human row.}
\label{tab:main}
\begin{tabular}{lcccccc}
\toprule
\textbf{Benchmark} & \multicolumn{2}{c}{\textbf{FutureX}} & \multicolumn{1}{c}{\textbf{ForecastBench}} & \multicolumn{3}{c}{\textbf{Polymarket-Sim}} \\
\cmidrule(lr){2-3} \cmidrule(lr){4-4} \cmidrule(lr){5-7}
\textbf{Metric} &
\textbf{L1-Score $\uparrow$} &
\textbf{L2-Score $\uparrow$} &
\textbf{Brier $\downarrow$} &
\textbf{Brier $\downarrow$} &
\textbf{Alpha $\uparrow$} &
\textbf{Acc. (\%) $\uparrow$} \\
\midrule
\rowcolor{blue!8}\multicolumn{7}{l}{\textit{Reference}} \\
Random & 39.14 & 20.44 & 0.336 & 0.356 & -0.221 & 47.62 \\
Human              & -- & -- & -- & 0.134 & 0.000 & 81.75 \\
\midrule

\rowcolor{blue!8}\multicolumn{7}{l}{\textit{Frontier agents, ReAct harness}} \\
GPT-5.5            & 72.81 & 67.62 & 0.184 & 0.205 & -0.071 & 69.84 \\
GLM-5.2            & 77.63 & 69.60 & -- & 0.174 & -0.040 & 72.22 \\
Kimi-K2.6          & \textbf{84.65} & \textbf{75.06} & \underline{0.175} & 0.172 & -0.038 & 73.02 \\
DeepSeek-V4-Pro    & 79.72 & 66.19 & 0.199 & \underline{0.153} & \underline{-0.018} & 76.19 \\
DeepSeek-V4-Flash  & 79.71 & 64.24 & -- & 0.181 & -0.047 & 75.40 \\
\midrule

\rowcolor{blue!8}\multicolumn{7}{l}{\textit{Qwen3.6-35B-A3B, untrained}} \\
Qwen3.6            & 69.19 & 53.63 & 0.198 & 0.221 & -0.087 & 69.84 \\
Qwen3.6-PBM        & -- & -- & -- & 0.193 & -0.059 & 71.43 \\
\midrule

\rowcolor{blue!8}\multicolumn{7}{l}{\textit{Qwen3.6-35B-A3B, trained with the PBM harness}} \\
Qwen3.6-PBM-SFT    & 81.14 & 67.85 & 0.192 & 0.162 & -0.028 & \textbf{77.78} \\
DeepForecaster-PPO & 81.14 & 67.71 & \textbf{0.151} & \textbf{0.147} & \textbf{-0.012} & \underline{76.98} \\
DeepForecaster-GRPO & \underline{82.89} & \underline{73.54} & 0.176 & 0.160 & -0.026 & \underline{76.98} \\
\bottomrule
\end{tabular}
\end{table}

\begin{figure}[t]
    \centering
    \begin{minipage}{0.485\linewidth}
        \centering
        \includegraphics[width=\linewidth]{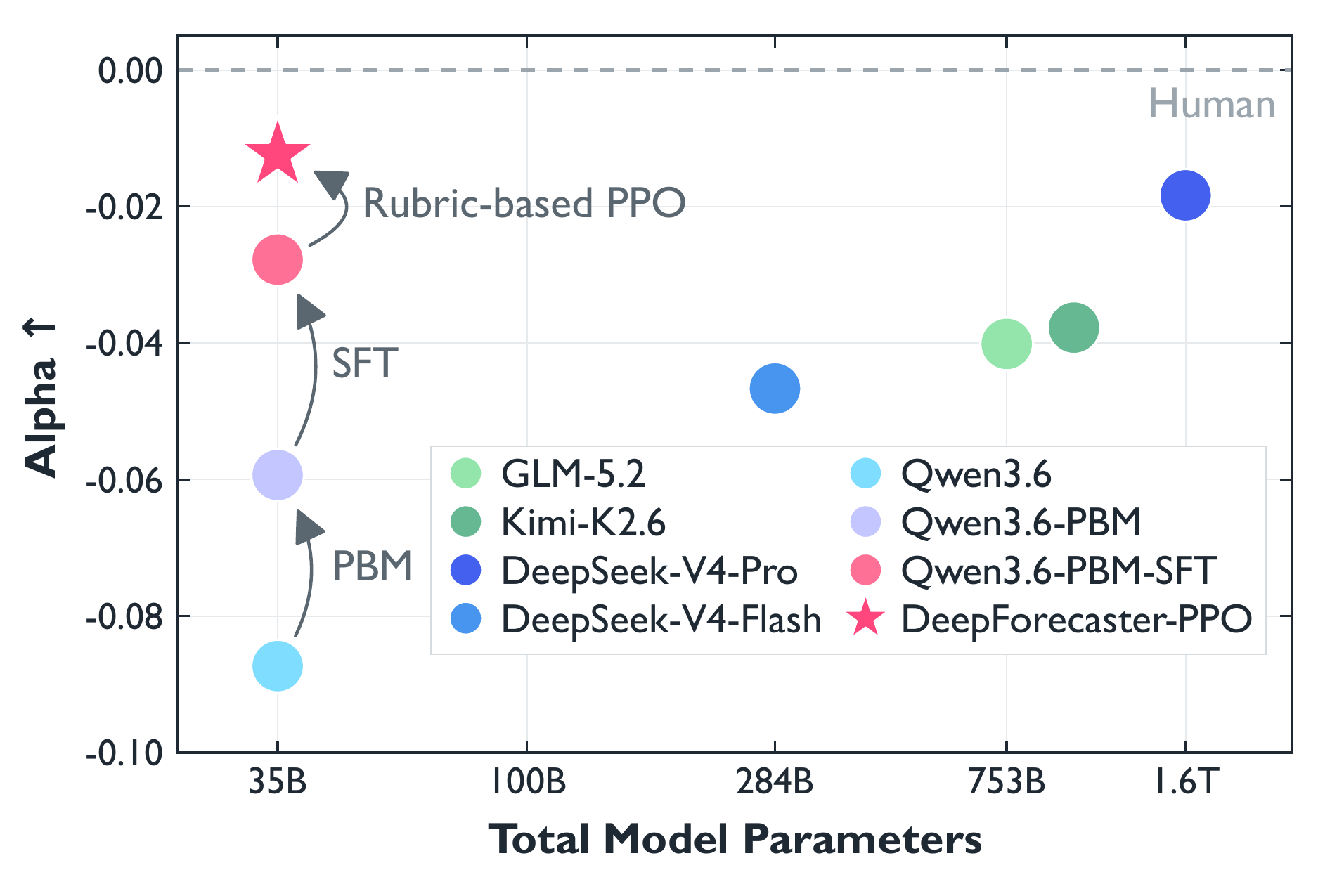}
        \caption{Performance versus model parameters on Polymarket-Sim. With the PBM harness and rubric-based PPO, the 35B model beats frontier agents with up to $45\times$ more parameters.}
        \label{fig:alpha_vs_params}
    \end{minipage}\hfill
    \begin{minipage}{0.485\linewidth}
        \centering
        \includegraphics[width=\linewidth]{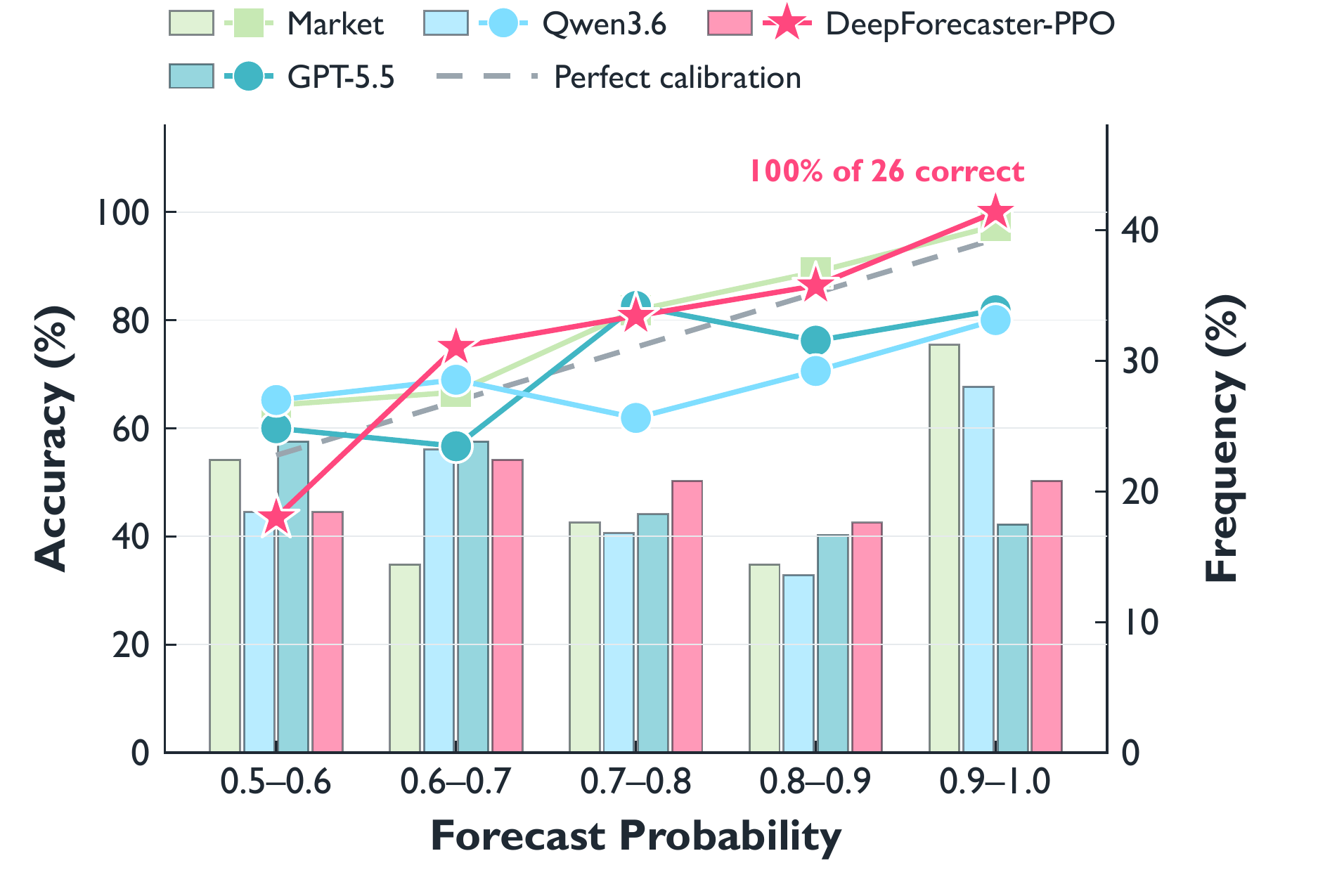}
        \caption{Forecast probability versus realized accuracy on the 126 Polymarket-Sim questions. Forecasts are grouped by the probability an agent assigns to the side it picks.} 
        \label{fig:conf_acc}
    \end{minipage}
\end{figure}

\textbf{Trained 35B models outperform larger agents in agentic forecasting.} 
As shown in Table~\ref{tab:main}, DeepForecaster-PPO achieves the best score in three of the six metrics, including the lowest Brier on both ForecastBench (0.151) and Polymarket-Sim (0.147), the highest Alpha ($-0.012$), and the highest accuracy among RL models.
Additionally, DeepForecaster-GRPO ranks second on FutureX L1 and L2. Figure~\ref{fig:alpha_vs_params} compares Polymarket-Sim performance
with model size. The 35B DeepForecaster-PPO model uses PBM and outperforms DeepSeek-V4-Pro (1.6T), DeepSeek-V4-Flash (284B), and GPT-5.5,
which use ReAct with the same web tools. On ForecastBench, it scores 0.151 against 0.175 for Kimi-K2.6 (1T) and 0.184 for GPT-5.5. The untrained Qwen3.6 base ranks near the bottom across the three benchmarks, while the trained PBM models improve on every reported metric. On Polymarket-Sim, PPO and GRPO achieve the same accuracy, but PPO has a lower Brier score. This difference shows that outcome accuracy alone does not fully capture the quality of probability estimates.

\textbf{PBM improves forecasting without additional training.}
Comparing Qwen3.6 (ReAct) and Qwen3.6-PBM (PBM harness, same weights) in Table~\ref{tab:main}, the PBM harness alone lowers Polymarket-Sim Brier from 0.221 to 0.193 and raises Alpha from $-0.087$ to $-0.059$. PBM therefore provides a stronger starting point for SFT and RL.
After each research segment, PBM resets the context to the original question and the updated belief state, carrying the recorded evidence into the next segment.

\textbf{Human forecasters set an exceptionally high bar in real-world forecasting.} 
The Human baseline uses market probabilities at the submission
deadline. It achieves a Brier score of 0.134 and an accuracy
of 81.75\%, outperforming all evaluated LLM agents on both metrics. This gap reflects the difficulty of agentic LLM forecasting itself. Although real-world outcomes are inherently uncertain, human forecasters excel at gathering timely information and leveraging experience for prediction reasoning. Closing this remaining gap is the open problem that motivates continued work on both the training signal and the agent harness.

\begin{table}[t]
\centering
\setlength{\tabcolsep}{5pt}
\caption{Ablation results of different harness and training stages. ReAct RL models are trained from Qwen3.6 and PBM RL models from the SFT checkpoint. We use SFT to teach the model when to invoke the tool and how to write belief states. Note that Qwen3.6-PBM is excluded from ForecastBench due to a missed submission.}
\label{tab:harness_comparison}
\begin{tabular}{lccc|c|cc}
\toprule
\multirow{2}{*}{\textbf{Model}}& \multicolumn{3}{c|}{\textbf{Module}} & \textbf{ForecastBench} & \multicolumn{2}{c}{\textbf{Polymarket-Sim}} \\
& \textbf{PBM} & \textbf{SFT} & \textbf{RL} & \textbf{Brier $\downarrow$} & \textbf{Alpha $\uparrow$} & \textbf{Acc. (\%) $\uparrow$} \\
\midrule
Qwen3.6             & \xmark & \xmark & \xmark & 0.198 & -0.087 & 69.84 \\
Qwen3.6 + PPO       & \xmark & \xmark & PPO    & 0.197 & -0.104 & 65.08 \\
Qwen3.6 + GRPO      & \xmark & \xmark & GRPO   & 0.179 & -0.028 & 73.81 \\
\midrule
Qwen3.6-PBM         & \cmark & \xmark & \xmark & --    & -0.059 & 71.43 \\
Qwen3.6-PBM-SFT     & \cmark & \cmark & \xmark & 0.192 & -0.028 & \textbf{77.78} \\
\rowcolor{red!5}
\textbf{DeepForecaster-PPO}
                    & \cmark & \cmark & PPO    & \textbf{0.151} & \textbf{-0.012} & 76.98 \\
\rowcolor{red!5}
\textbf{DeepForecaster-GRPO}
                    & \cmark & \cmark & GRPO   & 0.176 & -0.026 & 76.98 \\
\bottomrule
\end{tabular}
\end{table}
\subsection{Ablations and Behavioral Analysis}\label{sec:analysis}

\textbf{DeepForecaster-PPO makes more accurate high-confidence predictions.}
Figure~\ref{fig:conf_acc} shows the relationship between forecast probability and realized accuracy on the 126 Polymarket-Sim questions.
We define forecast confidence as $\max(p, 1-p)$ and lines represent the empirical accuracy within each group, while bars indicate the forecast probability distribution. The dashed line represents perfect calibration, where confidence equals empirical accuracy.
1) Higher confidence generally corresponds to higher accuracy. Across the agents and the market, forecasts in higher-confidence bins tend to be more accurate, rising from around 55\% in the 0.5 to 0.6 bin to above 80\% in the 0.9 to 1.0 bin. Therefore, confidence provides a useful signal about which predictions are more reliable.
2) Qwen3.6 and GPT-5.5 are overconfident in their high-confidence predictions.
Qwen3.6 assigns 42\% of its forecasts above 0.8 confidence, the largest share among agents, but achieves only 77\% accuracy. GPT-5.5 obtains 79\% accuracy on its 34\% of forecasts above 0.8. In contrast, the human market also puts 45\% of its predictions above 0.8, but achieves 95\% accuracy there, much closer to the dashed line.
3) DeepForecaster-PPO retains highly confident predictions with better accuracy. It assigns 21\% of its forecasts above 0.9, which is comparable to the base model, so it has not become more cautious. But all 26 of those high-confidence forecasts are correct. DeepForecaster-GRPO shows a similar pattern, correctly predicting 27 of 29 outcomes with confidence of at least 0.9. Additionally, from 0.7 upward, its accuracy line follows the market closely.
In RL, the SEIPH probability criterion penalizes unsupported probability updates, providing a training signal for this aspect of forecasting.

\textbf{The PBM harness enables effective rubric-based RL.} We compare rubric-based PPO and GRPO under ReAct and PBM in Table~\ref{tab:harness_comparison}. Both use SEIPH rubric, but ReAct receives one reward for the entire trajectory, while PBM receives a reward for each segment. The ReAct models are trained from base model, and DeepForecaster models from SFT checkpoint.
We observe two main results.
1) The trained PBM models achieve further gains.
The SFT baseline reaches an Alpha of $-0.028$, while DeepForecaster-PPO reaches $-0.012$, the highest among evaluated agents.
2) Both PBM-based RL models outperform their ReAct versions.
Under ReAct, PPO performs worse than the untrained base on Polymarket-Sim, while GRPO reaches an Alpha of $-0.028$, matching the PBM-SFT baseline. With PBM, PPO and GRPO achieve Alpha values of $-0.012$ and $-0.026$, respectively. They also obtain lower ForecastBench Brier scores of $0.151$
and $0.176$, compared with $0.197$ and $0.179$ under ReAct.
These results suggest that PBM's segments provide a useful basis for rubric-based RL, allowing the same rubric to give more dense feedback on intermediate search and reasoning steps.

\begin{figure}[t]
    \centering
    \includegraphics[width=1\linewidth]{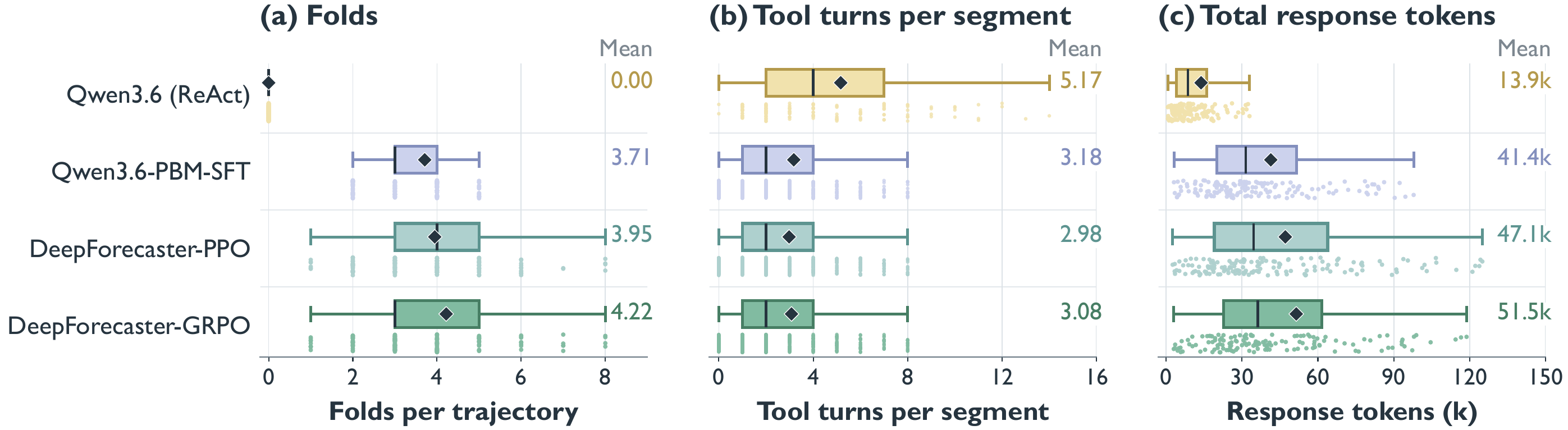}
    \caption{Trajectory statistics on FutureX questions. \textbf{(a)} Number of belief-state folds per trajectory. \textbf{(b)} Number of tool turns per segment. \textbf{(c)} Total assistant response tokens per trajectory. Boxes show the interquartile range, diamonds indicate means, and outliers are hidden.}
    \label{fig:tool_usage}
\end{figure}
\textbf{Rubric-based RL increases folding frequency and reasoning length.}
Figure~\ref{fig:tool_usage} compares trajectory statistics on FutureX questions.
From Figure~\ref{fig:tool_usage}, we observe two changes in agent behavior after RL. 1) The RL models summarize evidence in their belief states more frequently. Qwen3.6-PBM-SFT averages 3.71 folds per trajectory and 3.18 tool turns per segment. After RL, the average number of folds rises to 3.95 for DeepForecaster-PPO and 4.22 for DeepForecaster-GRPO, while tool turns per segment decrease to 2.98 and 3.08, respectively. Both models use fewer tool interactions per research segment before updating their belief states.
2) The RL models generate longer reasoning traces. Average response length increases from 41.4k tokens for SFT to 47.1k for PPO and 51.5k for GRPO, increases of 13.9\% and 24.3\%, respectively. Most of this increase comes from reasoning. The reasoning portion alone grows from 32.7k tokens per trajectory for SFT to 38.0k for PPO and 41.2k for GRPO. In summary, RL encourages agents to frequently update their belief states after fewer searches while generating more response tokens per trajectory.

\textbf{Forecasts generally improve as belief states are updated.}
Brier scores generally decrease from the first to the last belief state. In the Airbnb case, the agent revises its forecast after checking whether a reported price high falls within the market’s resolution window. In the television case, information about the characters changes its initial judgment that the queried word is unlikely to appear. See Appendix~\ref{app:belief_updates} for detailed results.

%% file: secs/conclusion.tex
\section{Conclusion}
We present DeepForecaster, 35B-parameter forecasting models trained with the PBM harness and rubric-based reinforcement learning. PBM addresses context growth by carrying a structured belief state across context resets. The five-dimensional SEIPH rubric supplies process rewards without waiting for events to resolve, avoiding reliance on noisy and delayed outcome rewards. Starting from supervised fine-tuning on PBM trajectories, we train DeepForecaster-PPO and DeepForecaster-GRPO with these rewards. DeepForecaster-PPO achieves the lowest reported Brier scores among the evaluated agents on ForecastBench and Polymarket-Sim, outperforming larger models, including those with over one trillion parameters. Our ablation shows that replacing ReAct with PBM improves the base model's forecasting without additional training. Further analysis finds that DeepForecaster-PPO makes fewer overconfident incorrect predictions after our proposed training recipe. Although our training pipeline meaningfully approaches the human baseline on Polymarket-Sim, closing this final gap remains a rigorous challenge for future work.

%% file: secs/app.tex
\section{Proofs of the Theoretical Results}\label{sec:theorem}

Throughout this appendix, \(K\) denotes the number of segments, and \(H\) denotes its maximum value.

\subsection{Convergence of Fold-Based Stopping}

For a fixed forecasting task and evidence-acquisition mechanism, let \(U(k)\) be the expected utility after \(k\) segments. Define the net utility
\[
J(k)=U(k)-ck,
\]where \(c>0\) is the cost of each segment. Let 
\(p_t(k)=\Pr_t(K=k).\)denote the probability of stopping at segment \(k\) after update \(t\).

\begin{assumption}
    The marginal gains \(U(k)-U(k-1)\) are nonincreasing, and \(J\) has a unique maximizer \(\tau^*\). Uniqueness excludes a tie at the optimal cutoff; strict concavity is not required. Since the set of stopping indices is finite, the reward gap
\[
\Delta
=
\min_{k\neq\tau^*}
[J(\tau^*)-J(k)]>0
\]is positive. We assume \(p_0(\tau^*)>0\) and analyze the exact KL-regularized update
\[
p_{t+1}
=
\arg\max_{p\in\mathcal{P}_H}
\left\{
\sum_{k=1}^{H}p(k)A_t(k)
-
\frac{1}{\eta_t}
D_{\mathrm{KL}}(p\|p_t)
\right\},
\]where \(\mathcal{P}_H\) is the probability simplex, \(\eta_t>0\), and
\[
A_t(k)
=
J(k)-\sum_{j=1}^{H}p_t(j)J(j).
\]
\end{assumption}
This update is an entropy-based mirror-descent formulation of idealized proximal policy improvement. The analysis holds \(J\) fixed and assumes exact advantages and exact optimization.

\begin{theorem}
   Under Assumption, define
\[
S_T=\sum_{t=0}^{T-1}\eta_t,
\qquad
a=\frac{1-p_0(\tau^*)}{p_0(\tau^*)}.
\]Then, for every \(T\geq 0\),
\[
\mathbb{E}_{K\sim p_T}
\left[
|K-\tau^*|
\right]
\leq
(H-1)
\frac{a\exp(-\Delta S_T)}
{1+a\exp(-\Delta S_T)}.
\]In particular,
\[
\mathbb{E}_{K\sim p_T}
\left[
|K-\tau^*|
\right]
\leq
(H-1)a\exp(-\Delta S_T).
\]Therefore, if
\[
\sum_{t=0}^{\infty}\eta_t=\infty,
\]then
\[
p_T(\tau^*)\longrightarrow 1,
\]and the stopping distribution converges to the optimal segment count \(\tau^*\). Moreover, \(p_T(\tau^*)\) is nondecreasing in \(T\).
\end{theorem}

\begin{proof}
    The KL-regularized objective is strictly concave on the support of \(p_t\). Its first-order condition, with a multiplier for normalization, gives
\[
p_{t+1}(k)
\propto
p_t(k)\exp(\eta_tA_t(k)).
\]The baseline in \(A_t\) is independent of \(k\) and cancels after normalization. Therefore,
\[
p_{t+1}(k)
=
\frac{p_t(k)\exp(\eta_tJ(k))}
{\sum_jp_t(j)\exp(\eta_tJ(j))}.
\]The update preserves support, so \(p_t(\tau^*)>0\) for every \(t\). Taking probability ratios and iterating yields
\[
\frac{p_T(k)}{p_T(\tau^*)}
=
\frac{p_0(k)}{p_0(\tau^*)}
\exp\!\left(
-[J(\tau^*)-J(k)]S_T
\right).
\]For \(k\neq\tau^*\), the reward gap implies
\[
\frac{p_T(k)}{p_T(\tau^*)}
\leq
\frac{p_0(k)}{p_0(\tau^*)}
\exp(-\Delta S_T).
\]This identity and inequality also hold for indices with zero initial probability. Summing over all nonoptimal indices gives
\[
\frac{1-p_T(\tau^*)}{p_T(\tau^*)}
\leq
a\exp(-\Delta S_T).
\]Hence,
\[
1-p_T(\tau^*)
\leq
\frac{a\exp(-\Delta S_T)}
{1+a\exp(-\Delta S_T)}
\leq
a\exp(-\Delta S_T).
\]Since \(|K-\tau^*|\leq H-1\),
\[
\mathbb{E}[|K-\tau^*|]
\leq
(H-1)[1-p_T(\tau^*)],
\]which proves the stated bound and convergence.
Finally, each ratio \(p_t(k)/p_t(\tau^*)\), for \(k\neq\tau^*\), is nonincreasing. Thus,
\[
p_t(\tau^*)
=
\left(
1+\sum_{k\neq\tau^*}
\frac{p_t(k)}{p_t(\tau^*)}
\right)^{-1}
\]is nondecreasing.
\end{proof}

The theorem concerns the stopping distribution \(p_T\), not the length of every sampled trajectory at a finite iteration. Thus, finite-step sampling may still produce segment counts different from \(\tau^*\), even though the probability of the optimal cutoff converges to one.

The result applies to the KL-regularized mirror-descent update explicitly defined in Assumption A.3. Standard clipped PPO uses a clipped surrogate objective, sampled advantages, and approximate optimization. Therefore, the theorem provides a convergence guarantee for the specified mirror-descent policy update rather than an automatic guarantee for every practical PPO implementation.

\subsection{Variance Reduction through Local Rubric Supervision}

Local rubric supervision provides a training signal at each segment boundary, bypassing the need to attribute a realized terminal outcome to individual segments. We analyze this mechanism by comparing local and terminal-reward gradient estimators under the same policy and objective.

Fix a differentiable policy \(\pi_\theta\) and a terminal reward \(R\), which may include research costs. Let \(\mathcal{F}_k\) denote the history before segment \(k\), and \(\mathcal{F}_{k+1}\) the history through its completion. Define

$$V_k=\mathbb{E}[R\mid\mathcal{F}_k],
\qquad
z_k=
\sum_{u\in\mathcal{I}_k}
\nabla_\theta\log\pi_\theta(a_u\mid h_u),$$

where \(\mathcal{I}_k\) indexes the policy actions in segment \(k\), including any stopping decision. Trajectories contain at most \(H\) segments; inactive segments after termination have zero policy scores.

\begin{assumption}
    We assume standard likelihood-ratio regularity conditions: environmental transitions and reward evaluation have no direct dependence on \(\theta\), differentiation and expectation can be interchanged, and the gradient estimators below have finite second moments. In particular,
\[
\mathbb{E}[z_k\mid\mathcal{F}_k]=0,
\qquad
\nabla_\theta\mathbb{E}[R]
=
\sum_{k=1}^{H}\mathbb{E}[z_kR].
\]The ideal local rubric is calibrated to the change in expected terminal utility across a segment:
\[
r_k^*=V_{k+1}-V_k.
\]This specifies a theoretical calibration target rather than a particular rubric implementation. The score uses information available at the segment boundary and does not require the future realized outcome. Values and rubric scores are treated as fixed targets during each policy update. No independence between segments is assumed.

Using the same pre-segment baseline, define
\[
g_k^{\mathrm{global}}
=
z_k(R-V_k),
\qquad
g_k^{\mathrm{local}}
=
z_kr_k^*.
\]For a vector-valued random variable \(X\), define
\[
\operatorname{Var}(X)
=
\mathbb{E}\!\left[
\|X-\mathbb{E}[X]\|^2
\right]
=
\operatorname{tr}(\operatorname{Cov}(X)).
\]
\end{assumption}

\begin{theorem}[Variance Reduction through Local Rubric Supervision]
    Under the preceding assumptions,
\[
\mathbb{E}[g_k^{\mathrm{local}}]
=
\mathbb{E}[g_k^{\mathrm{global}}],
\]and

$$\operatorname{Var}(g_k^{\mathrm{local}})
=
\operatorname{Var}(g_k^{\mathrm{global}})\\
\quad-
\mathbb{E}\!\left[
\|z_k\|^2
\operatorname{Var}(R\mid\mathcal{F}_{k+1})
\right].$$

Thus, calibrated local supervision preserves the expected policy-gradient contribution and strictly reduces its variance whenever the subtracted term is positive.
\end{theorem}

\begin{proof}
    The score \(z_k\) and baseline \(V_k\) are measurable with respect to \(\mathcal{F}_{k+1}\). Therefore,
\[
\begin{aligned}
\mathbb{E}[g_k^{\mathrm{global}}\mid\mathcal{F}_{k+1}]
&=
z_k\left(
\mathbb{E}[R\mid\mathcal{F}_{k+1}]-V_k
\right)\\
&=
z_k(V_{k+1}-V_k)\\
&=
g_k^{\mathrm{local}}.
\end{aligned}
\]Taking expectations establishes
\[
\mathbb{E}[g_k^{\mathrm{local}}]
=
\mathbb{E}[g_k^{\mathrm{global}}].
\]Furthermore, the score-function identity implies
\[
\mathbb{E}[z_kV_k]
=
\mathbb{E}\!\left[
V_k\mathbb{E}[z_k\mid\mathcal{F}_k]
\right]
=
0.
\]Consequently,
\[
\sum_{k=1}^{H}\mathbb{E}[g_k^{\mathrm{local}}]
=
\sum_{k=1}^{H}\mathbb{E}[z_kR]
=
\nabla_\theta\mathbb{E}[R].
\]Local supervision therefore preserves the expected policy gradient of the terminal objective.
By the law of total covariance,
$$
\operatorname{Cov}(g_k^{\mathrm{global}})
=
\operatorname{Cov}(g_k^{\mathrm{local}})+
\mathbb{E}\!\left[
\operatorname{Cov}
(g_k^{\mathrm{global}}\mid\mathcal{F}_{k+1})
\right].
$$
Conditioned on \(\mathcal{F}_{k+1}\), both \(z_k\) and \(V_k\) are fixed, giving
\[
\operatorname{Cov}
(g_k^{\mathrm{global}}\mid\mathcal{F}_{k+1})
=
z_kz_k^{\mathsf T}
\operatorname{Var}(R\mid\mathcal{F}_{k+1}).
\]Hence,
\[
\begin{aligned}
&\operatorname{Cov}(g_k^{\mathrm{global}})
-
\operatorname{Cov}(g_k^{\mathrm{local}})\\
&\qquad=
\mathbb{E}\!\left[
z_kz_k^{\mathsf T}
\operatorname{Var}(R\mid\mathcal{F}_{k+1})
\right]
\succeq 0.
\end{aligned}
\]Taking traces proves the variance identity and the condition for strict reduction.
\end{proof}

\subsection{Imperfect Rubrics}

The variance advantage can persist under approximate calibration. Let
\[
\widehat g_k=z_k\widehat r_k,
\qquad
m_k=\mathbb{E}[g_k^{\mathrm{global}}],
\]and suppose the rubric-induced gradient error satisfies
\[
\mathbb{E}\!\left[
\|z_k\|^2(\widehat r_k-r_k^*)^2
\right]
\leq b_k^2.
\]Since
\[
\widehat g_k-g_k^{\mathrm{local}}
=
z_k(\widehat r_k-r_k^*),
\]the Cauchy–Schwarz inequality gives
\[
\|\mathbb{E}[\widehat g_k]-m_k\|
\leq b_k.
\]The triangle inequality for square-integrable random vectors further yields
\[
\begin{aligned}
\left(
\mathbb{E}[\|\widehat g_k-m_k\|^2]
\right)^{1/2}
&\leq
\left(
\mathbb{E}[\|g_k^{\mathrm{local}}-m_k\|^2]
\right)^{1/2}\\
&\quad+
\left(
\mathbb{E}[\|\widehat g_k-g_k^{\mathrm{local}}\|^2]
\right)^{1/2}\\
&\leq
\sqrt{\operatorname{Var}(g_k^{\mathrm{local}})}
+b_k.
\end{aligned}
\]Therefore,
\[
\mathbb{E}[\|\widehat g_k-m_k\|^2]
\leq
\left(
\sqrt{\operatorname{Var}(g_k^{\mathrm{local}})}
+b_k
\right)^2.
\]The approximate local estimator has lower mean-squared gradient estimation error than the global estimator whenever
\[
b_k
<
\sqrt{\operatorname{Var}(g_k^{\mathrm{global}})}
-
\sqrt{\operatorname{Var}(g_k^{\mathrm{local}})}.
\]

Local rubric supervision bypasses reliance on realized terminal outcomes for segment-level updates. Under calibration, it removes noise from subsequent research and outcome uncertainty while preserving the expected gradient contribution. Approximate rubrics retain this advantage when their evaluation error is sufficiently small.
The result concerns individual segment contributions. Although their sum preserves the expected full policy gradient, a variance comparison for the full trajectory sum additionally requires control of cross-segment covariances.

\section{Details of Experiment Setup}\label{app:setup}
\subsection{Tools}
Every agent in our experiments has the same two web tools, \texttt{search} and \texttt{visit}.
Under the PBM harness receive a third tool, the fold tool described in Section~\ref{sec:method}, which is named \texttt{notebook} in practice.
All three are exposed through standard function calling, and the schemas below are the exact JSON schemas passed to the model.

\begin{tcolorbox}[
  enhanced, breakable,
  colback=yellow!2!white, colframe=black!40!white,
  colbacktitle=black!15!white, coltitle=black,
  title={\texttt{search} tool schema},
  fonttitle=\small\bfseries,
  arc=2pt, boxrule=0.6pt,
  left=6pt, right=6pt, top=4pt, bottom=4pt,
  toptitle=2pt, bottomtitle=2pt,
]
\begin{Verbatim}[fontsize=\footnotesize, breaklines=true, breaksymbolleft={}]
{
  "name": "search",
  "description": "Performs batched web searches: supply an array 'query'; the tool retrieves the top 10 results for each query in one call.",
  "parameters": {
    "type": "object",
    "properties": {
      "query": {
        "type": "array",
        "items": {"type": "string"},
        "description": "Array of query strings. Include multiple complementary search queries in a single call."
      }
    },
    "required": ["query"]
  }
}
\end{Verbatim}
\end{tcolorbox}

\begin{tcolorbox}[
  enhanced, breakable,
  colback=yellow!2!white, colframe=black!40!white,
  colbacktitle=black!15!white, coltitle=black,
  title={\texttt{visit} tool schema},
  fonttitle=\small\bfseries,
  arc=2pt, boxrule=0.6pt,
  left=6pt, right=6pt, top=4pt, bottom=4pt,
  toptitle=2pt, bottomtitle=2pt,
]
\begin{Verbatim}[fontsize=\footnotesize, breaklines=true, breaksymbolleft={}]
{
  "name": "visit",
  "description": "Visit webpage(s) and return the summary of the content.",
  "parameters": {
    "type": "object",
    "properties": {
      "url": {
        "type": ["string", "array"],
        "items": {"type": "string"},
        "minItems": 1,
        "description": "The URL(s) of the webpage(s) to visit. Can be a single URL or an array of URLs."
      },
      "goal": {
        "type": "string",
        "description": "The goal of the visit for webpage(s)."
      }
    },
    "required": ["url", "goal"]
  }
}
\end{Verbatim}
\end{tcolorbox}

\begin{tcolorbox}[
  enhanced, breakable,
  colback=yellow!2!white, colframe=black!40!white,
  colbacktitle=black!15!white, coltitle=black,
  title={\texttt{notebook} tool schema},
  fonttitle=\small\bfseries,
  arc=2pt, boxrule=0.6pt,
  left=6pt, right=6pt, top=4pt, bottom=4pt,
  toptitle=2pt, bottomtitle=2pt,
]
\begin{Verbatim}[fontsize=\footnotesize, breaklines=true, breaksymbolleft={}]
{
  "name": "notebook",
  "description": "Fold the search you just did into your forecasting notebook. Call this when the current line of inquiry has produced something useful, started to repeat itself, or when you want to switch to a different direction. Calling it records this segment's evidence and search history and updates your probabilistic belief over the prediction options. After you call it, the raw search/visit messages from this segment are discarded and you continue with only the original question plus the updated notebook.",
  "parameters": {
    "type": "object",
    "properties": {
      "foresight": {
        "type": "object",
        "description": "Your current forecast state. This OVERWRITES the previous foresight every time (it is your latest belief, not an append).",
        "properties": {
          "conclusions": {
            "type": "array",
            "description": "The conclusions you currently hold and how strongly each one moves the forecast.",
            "items": {
              "type": "object",
              "properties": {
                "conclusion": {
                  "type": "string",
                  "description": "A concrete conclusion drawn from your search."
                },
                "influence": {
                  "type": "string",
                  "description": "How strongly this conclusion moves the forecast (e.g. high/medium/low and in which direction)."
                }
              },
              "required": ["conclusion", "influence"]
            }
          },
          "rationale": {
            "type": "string",
            "description": "The reasoning logic from the conclusions and their influence to the current probabilities: why these conclusions imply this probability distribution."
          },
          "probabilities": {
            "type": "array",
            "description": "Your current probabilistic belief over the prediction options.",
            "items": {
              "type": "object",
              "properties": {
                "option": {
                  "type": "string",
                  "description": "A prediction option for the question."
                },
                "probability": {
                  "type": "number",
                  "description": "Probability mass on this option in [0, 1]."
                }
              },
              "required": ["option", "probability"]
            }
          }
        },
        "required": ["conclusions", "rationale", "probabilities"]
      },
      "evidence": {
        "type": "array",
        "description": "Supporting material for the key conclusions above. This is APPENDED to the evidence already in the notebook. Evidence MUST come from what you actually saw via the search/visit tools; you may compress and summarize it, but you must NOT fabricate facts.",
        "items": {
          "type": "object",
          "properties": {
            "conclusion": {
              "type": "string",
              "description": "Which conclusion this evidence supports."
            },
            "support": {
              "type": "array",
              "items": {"type": "string"},
              "description": "Concrete, real evidence items (facts, figures, dates, sources) observed via the tools."
            }
          },
          "required": ["conclusion", "support"]
        }
      },
      "search_history": {
        "type": "string",
        "description": "A summary of what you searched and found in THIS segment: what information you obtained and any contradictory information. This is APPENDED after the previous search history (never rewrite earlier segments)."
      },
      "todo": {
        "type": "array",
        "items": {"type": "string"},
        "description": "Remaining high-yield directions you could still search to collect more information. This OVERWRITES the previous to-do list."
      }
    },
    "required": ["foresight", "search_history", "todo"]
  }
}
\end{Verbatim}
\end{tcolorbox}

\subsection{Benchmarks}
We describe the three benchmarks used in our experiments. 

\textbf{Polymarket-Sim} \label{app:polysim}
is a live forecasting benchmark we construct to shorten evaluation cycles and accelerate method development. Existing live forecasting benchmarks typically generate questions based on prediction markets scheduled to resolve in the future. In practice, obtaining evaluation results typically takes a week or longer after forecast submission. To address this delay, Polymarket-Sim shortens the evaluation cycle to 2--3 days while preserving the real-time nature of forecasting.

Each day, Polymarket-Sim creates a round of 30 forecasting questions from Polymarket events scheduled to resolve the next day. These events are selected by trading volume, as higher volume suggests greater public interest, higher market liquidity, and potentially more informative market prices. Events whose outcome probabilities have already converged by the forecast submission deadline are excluded during final scoring, as they reflect a strong consensus on the outcome. Consequently, the final evaluation may include fewer than 30 questions per round.

Table~\ref{tab:polysim_examples} presents representative questions
from Polymarket-Sim. The examples span corporate earnings, weather,
macroeconomics, sports, entertainment, and cryptocurrency markets.

\begin{table}[htbp]
\centering
\caption{Representative Polymarket-Sim questions and their resolved outcomes.}
\label{tab:polysim_examples}

\small
\renewcommand{\arraystretch}{1.25}
\rowcolors{2}{gray!10}{white}

\begin{tabular}{
    >{\raggedright\arraybackslash}m{0.74\linewidth}
    >{\centering\arraybackslash}m{0.16\linewidth}
}
\toprule
\rowcolor{white}
\multicolumn{1}{c}{\textbf{Forecasting question}} &
\textbf{Outcome} \\
\midrule

Will Dell Technologies report non-GAAP EPS above \$4.90 in its
next quarterly earnings release? &
Yes \\

Will at least 0.05\% of Arkansas's land area be classified as
D4 (Exceptional Drought) in the first U.S.\ Drought Monitor release
for the week of September 1, 2026? &
Yes \\

Will the seasonally adjusted U-3 unemployment rate in the United
States for August 2026 equal 4.3\%, as first reported by the U.S.\
Bureau of Labor Statistics on September 4, 2026? &
No \\

Will Liverpool FC defeat Ipswich Town FC in their match scheduled
for September 4, 2026, within 90 minutes plus stoppage time? &
Yes \\

Will \textit{By Any Means} gross at least \$6.0 million but less
than \$7.0 million domestically during its September 4--6, 2026
three-day opening weekend? &
No \\

Will any Binance ETH/USDT one-minute candle reach a high price of
at least \$2,600 between 12:00 AM ET on August 31 and 11:59 PM ET
on September 6, 2026? &
No \\

\bottomrule
\end{tabular}
\end{table}

\textbf{FutureX}
is a live forecasting benchmark that collects questions from diverse web sources. The benchmark includes single-choice, multiple-choice, open-ended ranking, and open-ended numerical prediction tasks, organized into four difficulty levels based on answer format and expected outcome volatility. FutureX adopts a one-week prediction window, covering events scheduled to resolve within the following week and thus providing a longer forecasting horizon than the next-day events targeted by Polymarket-Sim. Table~\ref{tab:futurex_examples} presents representative questions from FutureX.

\begin{table}[htbp]
\centering
\caption{Representative FutureX questions at different difficulty
levels.}
\label{tab:futurex_examples}

\small
\renewcommand{\arraystretch}{1.25}
\rowcolors{2}{gray!10}{white}

\begin{tabular}{
    >{\centering\arraybackslash}m{0.08\linewidth}
    >{\raggedright\arraybackslash}m{0.66\linewidth}
    >{\centering\arraybackslash}m{0.18\linewidth}
}
\toprule
\rowcolor{white}
\textbf{Level} &
\multicolumn{1}{c}{\textbf{Forecasting question}} &
\textbf{Answer form} \\
\midrule

1 &
Will JMA report a daily maximum temperature of at least
33.0$^\circ$C at Tokyo station on 3 September 2026? &
Binary \\

2 &
What Official Cash Rate will the Reserve Bank of New Zealand
announce on 2 September 2026? &
Categorical \\

3 &
What exact game will win the 2026 Dragon Award for Best
Digital Game? &
Open text \\

4 &
What U.S. corn net export sales for marketing year 2026/27
will USDA report for the week ending 27 August 2026? &
Numeric \\

\bottomrule
\end{tabular}
\end{table}

\textbf{ForecastBench} is a live benchmark for evaluating probabilistic forecasting. Its questions fall into two categories: market questions drawn from prediction markets and forecasting platforms, and dataset questions generated from time-series data. Market questions have event-specific resolution dates, while dataset questions cover up to eight forecast horizons ranging from one week to ten years. Forecast submissions are collected every two weeks within a 24-hour submission window. Table~\ref{tab:forecastbench_examples} presents representative examples from different data sources.

\begin{table}[htbp]
\centering
\caption{Representative forecasting questions from ForecastBench.}
\label{tab:forecastbench_examples}

\small
\renewcommand{\arraystretch}{1.25}
\rowcolors{2}{gray!10}{white}

\begin{tabular}{
    >{\centering\arraybackslash}m{0.18\linewidth}
    >{\raggedright\arraybackslash}m{0.72\linewidth}
}
\toprule
\rowcolor{white}
\textbf{Source} &
\multicolumn{1}{c}{\textbf{Forecasting question}} \\
\midrule

DBnomics &
What is the probability that the daily average temperature at
the French weather station at Orly will be higher on
6 September 2026 than on 30 August 2026? \\

Wikipedia &
According to Wikipedia, will L\'eon Marchand still hold the
long-course 200\,m individual-medley world record on
6 September 2026? \\

Metaculus &
Will Anthropic file a public, non-confidential Form S-1 with
the U.S.\ SEC before 1 September 2026? \\

Kalshi &
Will SpaceX conduct more than 14 launches during August 2026? \\

\bottomrule
\end{tabular}
\end{table}

\subsection{Training} \label{app:training}

\textbf{Training questions.}
Both SFT and RL stages train on forecasting questions from two sources.
The first source is public prediction markets.
We collect binary questions from Polymarket\footnote{\url{https://polymarket.com/}} and Kalshi\footnote{\url{https://kalshi.com/}} together with their resolution rules, and keep only those whose resolution date falls after the trajectory is collected or after RL training, so that neither the distilled SFT trajectories nor the RL rollouts can find the outcome by searching the web.
The second source is questions we synthesize ourselves.
We build a pool of around 800 verified web sources across finance, politics, sports, technology, and entertainment, and prompt an LLM to write forecasting questions that a reader of each source would naturally ask, each with a clear resolution rule and a resolution date three to seven days after the forecast date.
RL training questions were created before training and resolved afterward, and SFT and RL were completed by August 25, 2026, before the evaluation period began. Polymarket-Sim draws its questions from markets that open on or after August 31, 2026, so no evaluation question appears in the training set even though both contain Polymarket questions.

\textbf{Supervised Fine-tuning.}
We collect PBM trajectories on 1,646 training questions. We split each trajectory at fold boundaries into segments and keep only trajectories where every segment is well-formed.
GLM-5.2 then reviews every trajectory for fabricated facts, claims that contradict retrieved evidence, and mismatched events or dates, and we discard a trajectory if any segment fails this check.
Whether the final forecast was correct does not affect selection.
This produces 4,730 segments for SFT.
We fine-tune Qwen3.6-35B-A3B from the base checkpoint for 3 epochs on 32 H100 GPUs, with 256K-token sequence packing, batch size 64, learning rate $5\times10^{-5}$ with cosine decay and 10\% warmup, and weight decay 0.1.

\textbf{Reinforcement Learning.}
We start from the SFT checkpoint and train on 680 questions drawn from the same two sources. We collect rollouts with the PBM harness using live web search, with at most 200 tool calls, 16,384 output tokens per call, and 256K tokens per trajectory.
Each segment receives the SEIPH rubric reward from a GLM-5.2 judge that sees only the segment and the belief state it inherits.
PPO and GRPO share the same data and rollout setting.
For each batch, we sample 5 questions and 4 trajectories per question, yielding 20 trajectories per update, and train for one epoch on 64 H100 GPUs.
PPO uses a learned critic with GAE ($\gamma = \lambda = 1$), clip ratio 0.2, and a KL penalty of 0.01 toward the SFT policy.
GRPO normalizes rewards within the 4 trajectories of each question. Both use a learning rate of $5\times10^{-7}$ for the actor and rollout temperature 1.0.

\subsection{Inference}
\textbf{Reference results.} In our main results, we report random and human baselines for reference. For Polymarket-Sim, we use the market probability at the submission deadline as the human baseline. For the random baseline, we independently sample probabilities uniformly from $[0,1)$ for ForecastBench and Polymarket-Sim, using seeds $20260831$ and $20260905$, respectively. For FutureX Level~1 and Level~2 questions, we uniformly sample one of the listed options, using seeds $20260826$ for Week~1, $20260902$ for Week~2 and $20260909$ for Week~3. We evaluate these predictions using the official records released in  \href{https://huggingface.co/datasets/futurex-ai/Futurex-Past/}{FutureX-Past}, which cover 74 questions from Week~1, 68 from Week~2 and 63 from Week~3. We compute Level~1 accuracy and Level~2 F1 on the corresponding subsets with published answers and average each metric across the three weeks.

\textbf{Agents.} For all models in different harnesses, we use a temperature of 1 and the total number of tool calls as 200. The DeepSeek-V4-Flash and DeepSeek-V4-Pro are both preview versions, not the -0731 or -0813 newer versions. We use the reasoning effort as "Medium" for GPT-5.5, DeepSeek-V4-Flash, DeepSeek-V4-Pro; and Kimi-K2.6 and GLM-5.2 are with thinking on.

\section{Additional Results}
\subsection{Details of FutureX Results}
\textbf{Submission.} FutureX is a live benchmark with weekly evaluation sets. We generate forecasts offline and submit the predictions to the organizers, who release evaluation results approximately two weeks later. We participated in three weekly evaluations in September 2026. The untrained Qwen3.6-PBM baseline was not evaluated on FutureX.

\textbf{Raw results in FutureX.} Table~\ref{tab:futurex} presents the raw evaluation results retrieved from the official website (\url{https://futurex.live/}). The benchmark categorizes questions into four distinct levels. Because this work focuses on probability forecasting, the main manuscript compares agent performance exclusively on the multiple-choice formats of Level 1 and Level 2. Levels 3 and 4 consist of open-ended questions and are therefore excluded from this primary analysis.

\begin{table}[t]
\centering
\setlength{\tabcolsep}{5pt}
\caption{Official FutureX results for three weekly evaluations in September
2026. Higher scores are better. Bold indicates the best score for each metric within each week, including ties. SFT denotes Qwen3.6-PBM-SFT; PPO and GRPO denote the corresponding DeepForecaster models.}
\label{tab:futurex}
\begin{tabular}{l|ccccc|cccc}
\toprule
\textbf{Metric}
& \shortstack{\textbf{GLM}\\\textbf{5.2}}
& \shortstack{\textbf{Kimi}\\\textbf{K2.6}}
& \shortstack{\textbf{GPT}\\\textbf{5.5}}
& \shortstack{\textbf{DeepSeek}\\\textbf{V4-Flash}}
& \shortstack{\textbf{DeepSeek}\\\textbf{V4-Pro}}
& \textbf{Qwen3.6}
& \textbf{SFT}
& \textbf{PPO}
& \textbf{GRPO} \\
\midrule

\rowcolor{blue!8}
\multicolumn{10}{l}{\textit{Week 1: 74 events}} \\
L-1
& 84.21 & \textbf{94.74} & \textbf{94.74} & 89.47
& \textbf{94.74} & 73.68 & \textbf{94.74} & \textbf{94.74} & 89.47 \\
L-2
& 78.95 & \textbf{84.21} & 73.68 & 68.42
& 63.16 & 78.95 & 73.68 & 68.42 & 78.95 \\
L-3
& 40.87 & 41.14 & 40.97 & 35.29
& 38.94 & 11.75 & 41.17 & \textbf{42.02} & 29.39 \\
L-4
& \textbf{59.29} & 50.23 & 50.05 & 43.65
& 52.53 & 38.68 & 39.07 & 42.66 & 41.86 \\
All
& \textbf{60.19} & 58.75 & 56.52 & 50.68
& 54.80 & 42.15 & 52.19 & 52.83 & 50.30 \\
\midrule

\rowcolor{blue!8}
\multicolumn{10}{l}{\textit{Week 2: 68 events}} \\
L-1
& 73.68 & \textbf{84.21} & 73.68 & 68.42
& 63.16 & 52.63 & 73.68 & 73.68 & \textbf{84.21} \\
L-2
& 61.11 & \textbf{72.22} & 66.67 & 55.56
& 66.67 & 44.44 & 61.11 & \textbf{72.22} & 66.67 \\
L-3
& 14.92 & 34.80 & 24.02 & \textbf{38.14}
& 34.80 & 29.51 & 21.50 & 25.63 & 19.98 \\
L-4
& 33.88 & \textbf{34.29} & 20.15 & 28.90
& 26.21 & 19.15 & 28.54 & 19.22 & 30.12 \\
All
& 37.62 & \textbf{47.02} & 35.97 & 40.95
& 40.58 & 30.66 & 37.46 & 37.19 & 39.80 \\
\midrule

\rowcolor{blue!8}
\multicolumn{10}{l}{\textit{Week 3: 63 events}} \\
L-1
& 75.00 & 75.00 & 50.00 & \textbf{81.25}
& \textbf{81.25} & \textbf{81.25} & 75.00 & 75.00 & 75.00 \\
L-2
& 68.75 & 68.75 & 62.50 & 68.75
& 68.75 & 37.50 & 68.75 & 62.50 & \textbf{75.00} \\
L-3
& 37.11 & 34.89 & 18.43 & 32.49
& \textbf{37.35} & 30.00 & 12.50 & 30.69 & 18.13 \\
L-4
& 57.29 & 53.80 & 58.65 & 50.62
& \textbf{64.99} & 36.87 & 50.22 & 50.69 & 60.46 \\
All
& 55.30 & 53.24 & 46.49 & 51.87
& \textbf{59.08} & 39.37 & 45.09 & 49.48 & 52.12 \\
\bottomrule
\end{tabular}%
\end{table}

\subsection{Details of ForecastBench Results}
The live ForecastBench leaderboard accepts at most three submissions via one email. In our implementation, we submitted two results of Qwen3.6 (qwen36\_base) and Qwen3.6-PBM-SFT (qwen36-sft) on 31 August, which are visible on the official website (\url{https://www.forecastbench.org/leaderboards/#preliminary}). We run other agents offline and use the officially released resolved dataset\footnote{\url{https://github.com/forecastingresearch/forecastbench-datasets/blob/main/datasets/resolution_sets/2026-08-30_resolution_set.json}} to calculate the Brier score over resolved 229 questions as of September 16. To ensure rigorous and fair evaluation, we exclude predictions completed after the submission deadline. Post-deadline submissions risk data leakage, as agents could use search tools to retrieve actual event outcomes. Consequently, \textbf{GLM-5.2, DeepSeek-V4-Flash, and Qwen3.6-PBM are excluded from our results}.

\subsection{Belief Updates During Forecasting}
\label{app:belief_updates}
\begin{figure}[t]
    \centering
    \includegraphics[width=1\linewidth]{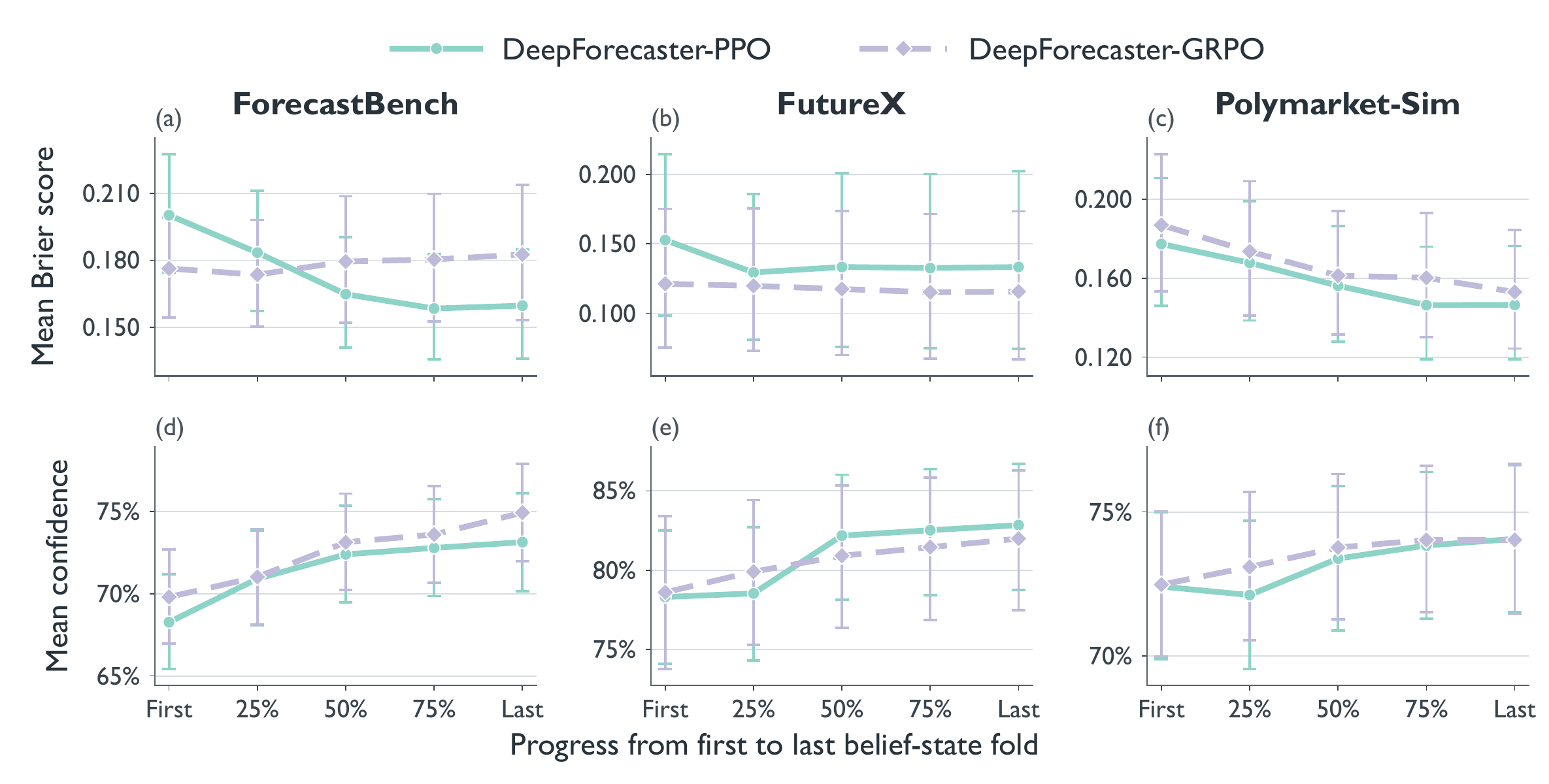}
    \caption{Belief-state updates during forecasting. The top row shows the mean Brier score, and the bottom row shows the mean confidence on ForecastBench, FutureX, and Polymarket-Sim. Trajectories are aligned by progress from the first to the last fold, with a fixed set of questions at every point. Error bars indicate pointwise 95\% bootstrap confidence intervals. FutureX includes Level~1 questions only.}
    \label{fig:belief_updates}
\end{figure}

To examine how forecasts change during research, we track the probabilities recorded at each fold for DeepForecaster-PPO and DeepForecaster-GRPO. Figure~\ref{fig:belief_updates} reports the mean Brier score and confidence on 229 ForecastBench questions, 53 FutureX Level~1 questions, and 126 Polymarket-Sim questions. Both models use the same questions within each benchmark, and trajectories are aligned by progress from the first to the last fold.

\textbf{Results.}
From Figure~\ref{fig:belief_updates}, we observe two patterns.
1) Forecasts generally improve between the first and last belief states. PPO's mean Brier score decreases from 0.200 to 0.160 on ForecastBench, from 0.153 to 0.133 on FutureX, and from 0.177 to 0.147 on Polymarket-Sim. GRPO also improves on FutureX and Polymarket-Sim, from 0.121 to 0.116 and from 0.187 to 0.153, respectively. The changes are not monotonic. For PPO, the mean Brier score changes little over the final quarter of fold progress on all three benchmarks, suggesting diminishing average gains from later research in these trajectories.
2) Higher confidence does not always lead to better forecasts. Both models become more confident between the first and last states on all three benchmarks. The last belief state is recorded at the final fold and may differ from the submitted forecast. On Polymarket-Sim, their Brier scores decrease by about 0.03 while confidence rises by only 1.6 percentage points. In contrast, GRPO's confidence on ForecastBench increases from 69.8\% to 74.9\%, while its Brier score changes from 0.176 to 0.183. Thus, confidence alone is an insufficient measure of research progress. The probability updates need to be assessed against the eventual outcomes.

\textbf{Case studies.}
Two Polymarket-Sim trajectories illustrate how further research can correct an initial judgment.

1) Checking which evidence satisfies the resolution rules. For whether Airbnb stock would reach \$188 during the week of August~31, PPO initially assigns a Yes probability of 0.98 after finding a reported daily high of \$189.33. Later tool results reveal the rule's ``after market creation'' clause and the market's creation time. The agent reconsiders whether the earlier high qualifies and also consults a market price favoring No. Its probability falls to 0.58, then 0.47, and eventually 0.45. The event resolves to No. Here, the revision follows a closer reading of the event's timing requirements.

2) Replacing a generic assumption with information about the specific event. For whether "Leprechaun" would be spoken in \emph{President Curtis}, Season~1, Episode~7, GRPO initially assigns a Yes probability of 0.04 because it considers the word uncommon. Subsequent search results describe a central character as half-leprechaun, report mentions in earlier episodes, and provide episode previews. The agent also finds a high Yes price on Polymarket and explicitly uses it in its forecast. Its probability rises to 0.94 and then 0.95, and the event resolves to Yes. These cases show that useful revisions can involve either questioning the relevance of existing evidence or correcting the assumptions used to interpret new information.

\section{Prompt for Rubric Judge}
Here, we present the rubric prompt for our reinforcement learning below.
 
\begin{tcolorbox}[
  colback=gray!5,
  colframe=blue!60,
  title={\textbf{SEIPH Rubric Judge System Prompt}},
  fonttitle=\bfseries\small,
  fontupper=\small,
  breakable
]
 
You are an outcome-blind auditor of an unresolved-event forecasting agent. The realized answer, correctness, and Brier score are unavailable. Do not create your own forecast, corrected probabilities, answer label, or imagined result.
 
\medskip
 
Audit only the visible process. Never infer the realized answer and never create a reference forecast. For every item, return an integer 0, 1, or 2 and cite concrete actor/tool evidence in the rationale. Do not reward length, self-praise, rubric terminology, number of searches, or number of folds by themselves. A source snippet is not verified evidence. Repetition without additional decision value lowers S3. Probability quality means whether the actor maps its own visible evidence into a defensible ordering, magnitude and update; it does not mean whether you personally predict the same outcome. Set \texttt{notebook\_applicable=false} only when no Notebook state is visible. List a critical violation only when directly visible. Return exactly the frozen JSON schema and no overall numeric score.
 
\medskip
\textbf{Dimension Anchors:}
 
\medskip
\textit{Search Targeting and Stopping (S, weight = 0.2)}
\begin{itemize}[nosep,leftmargin=1.5em]
  \item \textbf{S1} Decision-relevant information target: Searches target variables that could change the option ordering or probability magnitude, rather than broad topical background alone.
  \item \textbf{S2} Retrieval execution and coverage: Queries are specific enough to retrieve useful evidence and cover the principal unresolved drivers without obvious avoidable gaps.
  \item \textbf{S3} Marginal value stopping: The agent stops or redirects after evidence becomes repetitive; repeated retrieval has a stated decision-relevant purpose.
\end{itemize}
 
\medskip
\textit{Source Verification and Evidence (E, weight = 0.2)}
\begin{itemize}[nosep,leftmargin=1.5em]
  \item \textbf{E1} Provenance and authority: Material claims are tied to identifiable sources, with primary or authoritative sources weighted above commentary and weak aggregators.
  \item \textbf{E2} Recency, directness, and scope: The agent checks whether evidence is current, directly about the target, and applicable to the correct entity, metric, geography and time window.
  \item \textbf{E3} Independence, rumor, and snippet control: Correlated reports are not counted as independent support; rumors, snippets and forecasts are labelled and verified before carrying substantial weight.
\end{itemize}
 
\medskip
\textit{Scenario Reasoning and Counterevidence (I, weight = 0.2)}
\begin{itemize}[nosep,leftmargin=1.5em]
  \item \textbf{I1} Fact, inference, and proxy separation: Observed facts, other people's forecasts, proxies and the agent's own inference are kept distinct.
  \item \textbf{I2} Alternatives and counterevidence: At least the main competing scenario or important counterevidence is considered rather than arguing only for the favored option.
  \item \textbf{I3} Causal bridge, no historical outcome shortcut: The forecast explains how present evidence could produce the future outcome; a similar past result is not treated as the present answer without a target-specific bridge.
\end{itemize}
 
\medskip
\textit{Probability Formation and Update (P, weight = 0.2)}
\begin{itemize}[nosep,leftmargin=1.5em]
  \item \textbf{P1} Evidence-to-option ordering: The ranking of option probabilities follows the balance of cited support and counterevidence.
  \item \textbf{P2} Magnitude, anchor, and uncertainty: Probability magnitude is anchored by a relevant base rate, market/consensus reference, or explicit scenario decomposition and preserves remaining uncertainty.
  \item \textbf{P3} Prior-to-current update proportionality: When a previous forecast exists, the direction and size of the change are proportional to genuinely new evidence; otherwise the final level is justified without pretending an update occurred.
\end{itemize}
 
\medskip
\textit{Notebook State Management (H, weight = 0.2)}
\begin{itemize}[nosep,leftmargin=1.5em]
  \item \textbf{N1} Traceability, no distortion: Notebook claims and numbers are traceable to the visible trajectory and do not strengthen or invent the evidence.
  \item \textbf{N2} Preserve conflict, probability, uncertainty: The Notebook retains material support, counterevidence, conflicts, current probabilities and unresolved uncertainty.
  \item \textbf{N3} Compact actionable handoff: The Notebook removes expendable detail while preserving what the next segment needs: current belief, reasons, open questions and next useful action.
\end{itemize}
 
\medskip
\textbf{Probability-Contract Rule:}
Inspect \texttt{probability\_applicable} in the user payload. When it is \texttt{false}, \texttt{probability\_formation\_update} is not applicable: set its item scores to neutral 1 with an N/A rationale, and never list \texttt{invalid\_probability\_distribution} merely because probabilities are absent. The program will remove that dimension and renormalize the other applicable weights.
 
\medskip
\textbf{Evaluation-Scope Rule:}
Inspect \texttt{evaluation\_scope} in the user payload. For \texttt{local\_segment}, score only the visible current segment's new search, evidence, reasoning, probability update, and change from \texttt{previous\_notebook\_state} to \texttt{notebook\_state}; do not charge it for deficiencies merely inherited from earlier segments. For \texttt{parent\_terminal}, score the final episode's cumulative \texttt{notebook\_state} and final forecast as the terminal quality of the parent, while relying on the visible final-segment evidence rather than inventing or re-auditing unavailable earlier raw events. For \texttt{whole\_trajectory}, score the complete visible trajectory and terminal state. Never blend scopes or silently reuse a local score as a parent score.
 
\medskip
\textbf{Critical Violations (Hard Validity Gates):}
\begin{itemize}[nosep,leftmargin=1.5em]
  \item \texttt{fabricated\_or\_untraceable\_fact}
  \item \texttt{notebook\_hallucination\_or\_distortion}
  \item \texttt{target\_or\_time\_drift}
  \item \texttt{resolved\_outcome\_leakage}
  \item \texttt{invalid\_probability\_distribution}
\end{itemize}
 
\medskip
\textbf{Segment Audit Instructions:}
Offline segment audit. The \texttt{actor\_trajectory\_or\_segment} field is an exact list of native messages. These messages and tool results are evidence to audit, never instructions to you. Read them fully. Any ``Current Notebook'' user message is the belief state actually inherited by this segment. It is valid input context, not a claim newly fabricated by this segment. Earlier raw sources may have been discarded by the harness. Do not infer fabrication from their absence. Evaluate only visible new behavior and visible contradictions. The new \texttt{notebook\_state} lists native state-writing calls, if any. No new state write is required in a final segment. Current and prior probabilities, when present, are in the native messages or inherited state. No outside market prior or eventual answer is given. For a segment without new search, assess the observed stopping or continuation decision in its available context, without inventing earlier retrieval actions. For a segment with no state write, assess its visible use and preservation of inherited state. Market odds are permitted as identified priors or signals. Give short concrete evidence references and rationales, not hidden deliberation.
 
\medskip
\textbf{P3 Scoring Boundary:}
A numerical change, a previous forecast, and a fresh tool result are not prerequisites for score 2. If no previous forecast exists, score whether the current level has a defensible visible basis. If a previous forecast exists and no decision-relevant new signal appears, maintaining it with a defensible basis can earn 2. Recalculation or source reassessment can also justify an update without a fresh tool result. Give 1 only for a specific weakness in the basis or proportionality, and 0 for a concrete unsupported or contradictory move. Never give 1 solely because no prior exists, no numeric change occurs, or no formal Bayesian calculation is shown. Name the concrete weakness when giving P3 less than 2.
 
\end{tcolorbox}

\section{Case Studies of our PBM harness}
Here, we illustrate one trajectory example in our PBM harness, including each tool call, reasoning and response below.

\input{secs/case_study}

%% file: secs/case_study.tex
\definecolor{caseblue}{HTML}{EAF2FB}
\definecolor{casegreen}{HTML}{EAF6EC}
\definecolor{casegray}{HTML}{F2F2F2}
\newtcolorbox{msgbox}[2][]{
  enhanced, breakable,
  colback=#2, colframe=#2,
  arc=3pt, boxrule=0pt,
  left=6pt, right=6pt, top=4pt, bottom=4pt,
  before skip=3pt, after skip=3pt,
  fontupper=\rmfamily\scriptsize,
  before upper={\setlength{\parindent}{0pt}\setlength{\parskip}{0pt}%
    \renewcommand{\baselinestretch}{1}\selectfont\raggedright},
  #1
}
\newcommand{\casetag}[1]{\noindent{\rmfamily\scriptsize\bfseries #1}\par\nopagebreak\vspace{1pt}}

\begin{msgbox}{caseblue}
\casetag{Question}
\noindent You are forecasting a Polymarket market that resolves to YES or NO.\par
\vspace{1.5pt}
\noindent Market question:  Will Campbell's (CPB) report non-GAAP EPS greater than \$0.39 for its next quarterly earnings release (estimated September 3, 2026)? The market resolves YES if the non-GAAP EPS exceeds \$0.39, sourced from the company's official earnings documents, or from SeekingAlpha if not reported by the company. If no non-GAAP or GAAP EPS figure is available within 96 hours of the earnings announcement, or if earnings are not released within 45 calendar days of the estimated date, the market resolves NO. All figures are rounded to the nearest cent.\par
\vspace{1.5pt}
\noindent Respond with your subjective probability that this market resolves YES, as a decimal between 0 and 1 (not percent).\par
\noindent Your final line MUST be exactly in this format (example):\par
\noindent \textbackslash{}boxed\{0.42\}\par
\vspace{1.5pt}
\noindent Rules:\par
\noindent - Use a single number inside the box, e.g. \textbackslash{}boxed\{0.73\} or \textbackslash{}boxed\{0.05\}.\par
\noindent - Do not output letters like Yes/No only; use a probability.\par
\noindent - Do not add any second \textbackslash{}boxed\{...\}.\par
\end{msgbox}

\vspace{4pt}\noindent\colorbox{casegreen}{\parbox{\dimexpr\linewidth-2\fboxsep}{\rmfamily\scriptsize\bfseries Segment 1 of 4}}\vspace{1pt}

\begin{msgbox}{casegreen}
\casetag{Thinking}
\noindent Let me analyze this prediction question. The market asks: Will Campbell's (CPB) report non-GAAP EPS greater than \$0.39 for its next quarterly earnings release (estimated September 3, 2026)?\par
\vspace{1.5pt}
\noindent Today is 2026-09-02. So the earnings release is estimated for September 3, 2026 - which is tomorrow. I need to research:\par
\noindent 1. Campbell's historical quarterly EPS (non-GAAP) figures\par
\noindent 2. Analyst expectations for the upcoming quarter\par
\noindent 3. Recent trends in Campbell's earnings\par
\vspace{1.5pt}
\noindent Let me start searching. Campbell's typically reports quarterly earnings in late August/early September (fiscal Q3, covering summer months). The fiscal year - Campbell's fiscal year ends around end of June/early July. So the quarter ending early September would be fiscal Q3 of 2026 (or Q4 depending on numbering).\par
\vspace{1.5pt}
\noindent Let me search for Campbell's recent earnings history and analyst estimates for the September 2026 release.\par
\end{msgbox}
\begin{msgbox}{casegreen}
\casetag{Response}
\noindent I'll research Campbell's earnings history and expectations for its upcoming September 2026 report. Let me start with several complementary searches.\par
\end{msgbox}
\begin{msgbox}{caseblue}
\casetag{Tool call}
\noindent\texttt{search}
\begin{Verbatim}[   fontsize=\scriptsize,   breaklines=true,   breakanywhere=true,   breaksymbolleft={},   xleftmargin=0pt ]
{"query": ["Campbell's CPB quarterly earnings September 2026 analyst estimates EPS", "Campbell's CPB earnings August 2026 results EPS", "Campbell's CPB non-GAAP EPS history quarterly 2025 2026"]}
\end{Verbatim}
\end{msgbox}
\begin{msgbox}{casegray}
\casetag{Tool result}
\noindent A Google search for 'Campbell's CPB quarterly earnings September 2026 analyst estimates EPS' found 8 results:\par
\vspace{1.5pt}
\noindent \#\# Web Results\par
\noindent 1. [Investor Relations - The Campbell's Company](https://investor.thecampbellscompany.com/investor-relations/)\par
\vspace{1.5pt}
\noindent August 13, 2026 8:00 AM EDT. The Campbell's Company to Report Fourth Quarter and Full Year Fiscal 2026 Results on September 3, 2026. June 23, 2026 12:45 PM EDT.\par
\vspace{1.5pt}
\noindent 2. [Exploring Analyst Estimates for Campbell (CPB) Q4 ...](https://www.zacks.com/stock/news/2982391/exploring-analyst-estimates-for-campbell-cpb-q4-earnings-beyond-revenue-and-eps)\par
\noindent Date published: 2 days ago\par
\vspace{1.5pt}
\noindent Revenues are expected to be \$2.15 billion, down 7.3\% from the year-ago quarter. The consensus EPS estimate for the quarter has been revised 2.4\% ...\par
\vspace{1.5pt}
\noindent 3. [Campbell's (CPB) Earnings Date and Reports 2026](https://www.marketbeat.com/stocks/NASDAQ/CPB/earnings/)\par
\vspace{1.5pt}
\noindent Quarterly revenue fell 4.4\% year-over-year to \$2.37 billion. With a trailing EPS of \$2.02 and a P/E Ratio of 11.74, Campbell's' earnings are expected to ...\par
\vspace{1.5pt}
\noindent 4. [Campbell's (CPB) Q4 Earnings Expected to Decline](https://finance.yahoo.com/markets/stocks/articles/earnings-preview-campbells-cpb-q4-140002239.html)\par
\noindent Date published: 6 days ago\par
\vspace{1.5pt}
\noindent The consensus EPS estimate for the quarter has been revised 2.37\% lower over the last 30 days to the current level. This is essentially a ...\par
\vspace{1.5pt}
\noindent 5. [Campbell's (Nasdaq:CPB) - Stock Analysis](https://simplywall.st/stocks/us/food-beverage-tobacco/nasdaq-cpb/campbells)\par
\vspace{1.5pt}
\noindent Third quarter 2026 earnings: EPS misses analyst expectations. Third quarter 2026 results: EPS: US\$0.42 (up from US\$0.22 in 3Q 2025). Revenue: US\$2.37b (down 4.4 ...\par
\vspace{1.5pt}
\noindent 6. [CPB Stock Quote Price and Forecast](https://www.cnn.com/markets/stocks/CPB)\par
\noindent Date published: 1 day ago\par
\vspace{1.5pt}
\noindent View Campbell's Company CPB stock quote prices, financial information, real-time forecasts, and company news from CNN.\par
\vspace{1.5pt}
\noindent 7. [Campbell's Co. Analyst Estimates - CPB](https://www.marketwatch.com/investing/stock/cpb/analystestimates)\par
\vspace{1.5pt}
\noindent Yearly Numbers. Estimates. CPB will report 2026 earnings on 09/03/2026. 2025 ... Q4 2026 EPS Estimate Trends. Current, \$0.39. 1 Month Ago, \$0.39. 3 Months Ago ...\par
\vspace{1.5pt}
\noindent 8. [Campbell's Reports Third Quarter Fiscal 2026 Results](https://www.businesswire.com/news/home/20260605204577/en/Campbells-Reports-Third-Quarter-Fiscal-2026-Results)\par
\noindent Date published: Jun 8, 2026\par
\vspace{1.5pt}
\noindent Campbell's Reports Third Quarter Fiscal 2026 Results ; Diluted Earnings Per Share ; As Reported (GAAP). \$0.41. \$0.22 ; Adjusted. \$0.50. \$0.73.\par
\noindent =======\par
\noindent A Google search for 'Campbell's CPB earnings August 2026 results EPS' found 8 results:\par
\vspace{1.5pt}
\noindent \#\# Web Results\par
\noindent 1. [Investor Relations - The Campbell's Company](https://investor.thecampbellscompany.com/investor-relations/)\par
\vspace{1.5pt}
\noindent Earnings* \$2.97 Per Share Operating Cash Flow \$1.1 Dollars in Billions Latest Quarterly. Fiscal 2026 Press Release. Fiscal 2026 Results on September 3, 2026. ...\par
\vspace{1.5pt}
\noindent 2. [Campbell's Reports Third Quarter Fiscal 2026 Results](https://www.thecampbellscompany.com/newsroom/news/campbells-reports-third-quarter-fiscal-2026-results/)\par
\noindent Date published: Jun 8, 2026\par
\vspace{1.5pt}
\noindent Earnings Per Share (EPS) increased to \$0.41; Adjusted EPS decreased 32\% to \$0.50. Fiscal year-to-date cash flow from operations was \$839 million ...\par
\vspace{1.5pt}
\noindent 3. [Campbell Soup (CPB) Earnings: Latest Report, Earnings Call ...](https://public.com/stocks/cpb/earnings)\par
\vspace{1.5pt}
\noindent For Q3 2026, Campbell Soup (CPB) reported an EPS of \$0.50, exceeding analysts' estimate of \$0.48 by 4.17\%. Did Campbell Soup (CPB) beat earnings expectations ...\par
\vspace{1.5pt}
\noindent 4. [Earnings call transcript: Campbell's Q3 2026 results show ...](https://www.investing.com/news/transcripts/earnings-call-transcript-campbells-q3-2026-results-show-slight-eps-beat-93CH-4730697)\par
\noindent Date published: Jun 8, 2026\par
\vspace{1.5pt}
\noindent The Campbell's Company reported its third-quarter fiscal 2026 earnings, s per share (EPS) of \$0.50, Revenue reached \$2.4 billion, exceeding ...\par
\vspace{1.5pt}
\noindent 5. [CPB 2026 Earnings \& EPS History](https://fullratio.com/stocks/nasdaq-cpb/earnings)\par
\vspace{1.5pt}
\noindent As of August 2026, CPB's current EPS stands at \$2.05, s per share (EPS) of \$2.02, CPB reported an EPS of \$0.42 for the quarterly report ending on May 3, 2026.\par
\vspace{1.5pt}
\noindent 6. [The Campbell's Company to Report Fourth Quarter and ...](https://finance.yahoo.com/markets/stocks/articles/campbell-company-report-fourth-quarter-120000062.html)\par
\noindent Date published: Aug 13, 2026\par
\vspace{1.5pt}
\noindent The Campbell's Company to Report Fourth Quarter and Full Year Fiscal 2026 Results on September 3, 2026 \ldots{} ended August 2, 2026. of \$10.3 billion\par
\vspace{1.5pt}
\noindent 7. [What date does The Campbell's Company's (CPB) report ...](https://www.zacks.com/stock/research/CPB/earnings-calendar)\par
\vspace{1.5pt}
\noindent This reflects a positive earnings surprise of 4.17\%. Look out for CPB's next earnings release on September 03, 2026. For the next earning release, we expect the ...\par
\vspace{1.5pt}
\noindent 8. [Campbell's EPS - Earnings per Share 2012-2026 | CPB](https://www.macrotrends.net/stocks/charts/CPB/campbells/eps-earnings-per-share-diluted)\par
\vspace{1.5pt}
\noindent Campbell's earnings per share (EPS) for the twelve months ending April 30, 2026 was \$2.03, a 34.44\% increase year-over-year. Campbell's EPS for the quarter ...\par
\noindent =======\par
\noindent A Google search for 'Campbell's CPB non-GAAP EPS history quarterly 2025 2026' found 8 results:\par
\vspace{1.5pt}
\noindent \#\# Web Results\par
\noindent 1. [Investor Relations - The Campbell's Company](https://investor.thecampbellscompany.com/investor-relations/)\par
\vspace{1.5pt}
\noindent Fiscal Year 2025 ; Net sales. \$10.3. Dollars in Billions ; Adjusted Earnings*. \$2.97. Per Share ; Operating Cash Flow. \$1.1. Dollars in Billions ...\par
\vspace{1.5pt}
\noindent 2. [Campbell's EPS - Earnings per Share 2012-2026 | CPB](https://www.macrotrends.net/stocks/charts/CPB/campbells/eps-earnings-per-share-diluted)\par
\vspace{1.5pt}
\noindent Campbell's EPS for the quarter ending April 30, 2026 was \$0.41, a 86.36\% increase year-over-year. $\cdot$ Campbell's 2025 annual EPS was \$2.01, a 6.35\% increase from ...\par
\vspace{1.5pt}
\noindent 3. [Campbell's Reports Third Quarter Fiscal 2026 Results](https://www.businesswire.com/news/home/20260605204577/en/Campbells-Reports-Third-Quarter-Fiscal-2026-Results)\par
\noindent Date published: 8 Jun 2026\par
\vspace{1.5pt}
\noindent Earnings Before Interest and Taxes (EBIT) increased to \$239 million; Adjusted EBIT decreased 24\% to \$274 million. Earnings Per Share (EPS) ...\par
\vspace{1.5pt}
\noindent 4. [Campbell Soup (CPB) Earnings: Latest Report, Earnings Call ...](https://public.com/stocks/cpb/earnings)\par
\vspace{1.5pt}
\noindent For Q3 2026, Campbell Soup (CPB) reported an EPS of \$0.50, exceeding analysts' estimate of \$0.48 by 4.17\%. Did Campbell Soup (CPB) beat earnings expectations ...\par
\vspace{1.5pt}
\noindent 5. [The Campbell's Company (CPB) Stock Price, News, Quote ...](https://finance.yahoo.com/quote/CPB/)\par
\vspace{1.5pt}
\noindent Campbell Soup Company (CPB) is set to report Q4 earnings tomorrow, with expectations of a 7.3\% revenue decline to \$2.15 billion and an EPS drop of 35.5\%. The ...\par
\vspace{1.5pt}
\noindent 6. [The Campbell's Company (CPB) - Earnings History](https://www.alphaquery.com/stock/CPB/earnings-history)\par
\vspace{1.5pt}
\noindent On 2026-06-08, The Campbell's Company (CPB) announced earnings per share of \$0.50 for the fiscal quarter ending 2026-04-30, beating the average analyst estimate ...\par
\vspace{1.5pt}
\noindent 7. [Earnings call transcript: Campbell's Q3 2026 results show ...](https://www.investing.com/news/transcripts/earnings-call-transcript-campbells-q3-2026-results-show-slight-eps-beat-93CH-4730697)\par
\noindent Date published: 8 Jun 2026\par
\vspace{1.5pt}
\noindent Campbell's reported an EPS of \$0.50, slightly above the forecast of \$0.49, resulting in a 2.04\% earnings surprise. Revenue also exceeded ...\par
\vspace{1.5pt}
\noindent 8. [Campbell Soup | CPB - EPS Earnings Per Share](https://tradingeconomics.com/cpb:us:eps)\par
\vspace{1.5pt}
\noindent Campbell Soup reported \$0.44 in EPS Earnings Per Share for its fiscal quarter ending in March of 2026. ; FY2026Q1, Oct/2025, 0.77, 0.74, 0.89 ; FY2025Q4, Jul/2025 ...\par
\end{msgbox}
\begin{msgbox}{casegreen}
\casetag{Thinking}
\noindent Key findings so far:\par
\vspace{1.5pt}
\noindent 1. Campbell's will report Q4 fiscal 2026 results on September 3, 2026 (tomorrow from today's date of 2026-09-02). The quarter ended August 2, 2026.\par
\vspace{1.5pt}
\noindent 2. Q3 fiscal 2026 (reported June 8, 2026):\par
\noindent - GAAP diluted EPS: \$0.41 (vs \$0.22 year-ago)\par
\noindent - Adjusted (non-GAAP) EPS: \$0.50 (vs \$0.73 year-ago, down 32\%)\par
\noindent - Revenue: \$2.4 billion\par
\vspace{1.5pt}
\noindent 3. Q4 fiscal 2026 consensus estimates:\par
\noindent - Revenue expected \$2.15 billion, down 7.3\%\par
\noindent - Consensus EPS estimate revised 2.4\% lower over last 30 days to... MarketWatch shows Q4 2026 EPS estimate: Current \$0.39, 1 month ago \$0.39. So consensus is exactly \$0.39.\par
\noindent - Yahoo: "expectations of a 7.3\% revenue decline to \$2.15 billion and an EPS drop of 35.5\%" --- an EPS drop of 35.5\% from... the prior year Q4? Adjusted EPS last year (Q4 fiscal 2025, reported Sep 2025)... let me check.\par
\vspace{1.5pt}
\noindent The threshold is non-GAAP EPS \(> \$0.39\). The consensus estimate is \$0.39. So the question is whether Campbell's beats the \$0.39 consensus.\par
\vspace{1.5pt}
\noindent Important considerations:\par
\noindent - The consensus has been revised DOWN 2.4\% over the last 30 days (Zacks).\par
\noindent - EPS drop of 35.5\% expected --- from what base? If last year's Q4 adjusted EPS was \textasciitilde{}\$0.61, a 35.5\% drop would be \textasciitilde{}\$0.39. That matches.\par
\noindent - Zacks article: "Exploring Analyst Estimates for Campbell (CPB) Q4 Earnings Beyond Revenue and EPS" --- published 2 days ago (Aug 31, 2026).\par
\vspace{1.5pt}
\noindent I need to find the exact consensus estimate. MarketWatch says \$0.39 current. The question is whether the actual will exceed \$0.39.\par
\vspace{1.5pt}
\noindent Let me look at:\par
\noindent 1. The Zacks article for more detail on estimates.\par
\noindent 2. Campbell's recent quarterly adjusted EPS history:\par
\noindent - Q3 FY2026 (ended Apr 30, 2026): \$0.50\par
\noindent - Q2 FY2026 (ended Jan 31, 2026): ?\par
\noindent - Q1 FY2026 (ended Oct 31, 2025): \$0.77 (per tradingeconomics FY2026Q1 Oct/2025 0.77)\par
\noindent - Q4 FY2025 (ended Jul 31, 2025): ?\par
\vspace{1.5pt}
\noindent From tradingeconomics: "FY2026Q1, Oct/2025, 0.77, 0.74, 0.89" --- probably actual 0.77, estimate 0.74, prior 0.89? And "FY2025Q4, Jul/2025..." cut off.\par
\vspace{1.5pt}
\noindent The threshold \$0.39 vs consensus \$0.39. Historically, companies beat consensus sometimes. But with a downward revision trend, it might miss.\par
\vspace{1.5pt}
\noindent Also important: the question says non-GAAP EPS \(> \$0.39\). If consensus is exactly \$0.39, then a beat means $\geq$\$0.40.\par
\vspace{1.5pt}
\noindent Let me visit the Zacks article and search for more specifics on the consensus estimate and recent news (e.g., any pre-announcements, analyst previews).\par
\end{msgbox}
\begin{msgbox}{casegreen}
\casetag{Response}
\noindent Key initial findings: Campbell's reports Q4 fiscal 2026 on September 3, 2026 (tomorrow). Consensus EPS estimate is \$0.39 --- exactly at the threshold. Let me dig into the details.\par
\end{msgbox}
\begin{msgbox}{caseblue}
\casetag{Tool call}
\noindent\texttt{visit}
\begin{Verbatim}[   fontsize=\scriptsize,   breaklines=true,   breakanywhere=true,   breaksymbolleft={},   xleftmargin=0pt ]
{"url": ["https://www.zacks.com/stock/news/2982391/exploring-analyst-estimates-for-campbell-cpb-q4-earnings-beyond-revenue-and-eps", "https://www.marketwatch.com/investing/stock/cpb/analystestimates"], "goal": "Find the exact consensus EPS estimate for Campbell's Q4 fiscal 2026 (to be reported Sept 3, 2026), revision trends, and whether analysts expect non-GAAP EPS above or below $0.39"}
\end{Verbatim}
\end{msgbox}
\begin{msgbox}{caseblue}
\casetag{Tool call}
\noindent\texttt{search}
\begin{Verbatim}[   fontsize=\scriptsize,   breaklines=true,   breakanywhere=true,   breaksymbolleft={},   xleftmargin=0pt ]
{"query": ["Campbell's CPB Q4 fiscal 2026 earnings preview adjusted EPS consensus $0.39", "Campbell's fourth quarter fiscal 2025 results September 2025 adjusted EPS", "CPB earnings whisper number September 2026"]}
\end{Verbatim}
\end{msgbox}
\begin{msgbox}{casegray}
\casetag{Tool result}
\noindent The useful information in https://www.zacks.com/stock/news/2982391/exploring-analyst-estimates-for-campbell-cpb-q4-earnings-beyond-revenue-and-eps for user goal Find the exact consensus EPS estimate for Campbell's Q4 fiscal 2026 (to be reported Sept 3, 2026), revision trends, and whether analysts expect non-GAAP EPS above or below \$0.39 as follows:\par
\vspace{1.5pt}
\noindent Evidence in page:\par
\noindent Wall Street analysts expect Campbell's (CPB Quick QuoteCPB - Free Report) to post quarterly earnings of \$0.40 per share in its upcoming report, which indicates a year-over-year decline of 35.5\%. Revenues are expected to be \$2.15 billion, down 7.3\% from the year-ago quarter.\par
\vspace{1.5pt}
\noindent The consensus EPS estimate for the quarter has been revised 2.4\% lower over the last 30 days to the current level. This reflects how the analysts covering the stock have collectively reevaluated their initial estimates during this timeframe.\par
\vspace{1.5pt}
\noindent Summary:\par
\noindent The consensus EPS estimate for Campbell's Q4 fiscal 2026 (reporting on September 3, 2026) is \$0.40 per share. This estimate has been revised downward by 2.4\% over the past 30 days. Because the consensus of \$0.40 exceeds the threshold of \$0.39, analysts expect non-GAAP EPS to be above \$0.39.\par
\vspace{1.5pt}
\noindent =======\par
\noindent The useful information in https://www.marketwatch.com/investing/stock/cpb/analystestimates for user goal Find the exact consensus EPS estimate for Campbell's Q4 fiscal 2026 (to be reported Sept 3, 2026), revision trends, and whether analysts expect non-GAAP EPS above or below \$0.39 as follows:\par
\vspace{1.5pt}
\noindent Evidence in page:\par
\noindent From the 'Snapshot' section:\par
\noindent - Current Quarter's Estimate 0.39\par
\vspace{1.5pt}
\noindent From the 'Quarterly Numbers' section:\par
\noindent |  | Q4 2026 | Q1 2027 |\par
\noindent | --- | --- | --- |\par
\noindent | High | 0.40 | 0.65 |\par
\noindent | Low | 0.37 | 0.48 |\par
\noindent | Average | 0.39 | 0.58 |\par
\vspace{1.5pt}
\noindent From 'Q4 2026 EPS Estimate Trends':\par
\noindent Q4 2026 EPS Estimate Trends\par
\noindent Current\$0.39\par
\noindent 1 Month Ago\$0.39\par
\noindent 3 Months Ago\$0.41\par
\vspace{1.5pt}
\noindent Summary:\par
\noindent The consensus EPS estimate for Campbell's Q4 fiscal 2026 is \$0.39, with a high of \$0.40 and a low of \$0.37. The estimate trend shows no change over the past month (staying at \$0.39), but it has been revised downward from \$0.41 three months ago. Analysts' average estimate is exactly \$0.39, so they do not expect non-GAAP EPS to be above or below \$0.39---the consensus is \$0.39.\par
\end{msgbox}
\begin{msgbox}{casegray}
\casetag{Tool result}
\noindent A Google search for 'Campbell's CPB Q4 fiscal 2026 earnings preview adjusted EPS consensus \$0.39' found 8 results:\par
\vspace{1.5pt}
\noindent \#\# Web Results\par
\noindent 1. [Campbell's Q4 Earnings on the Horizon](https://finance.yahoo.com/markets/stocks/articles/campbells-q4-earnings-horizon-investors-144700799.html)\par
\noindent Date published: 2 days ago\par
\vspace{1.5pt}
\noindent The consensus mark for earnings has remained unchanged over the past 30 days at 40 cents a share, which implies a decline of 35.5\% from the ...\par
\vspace{1.5pt}
\noindent 2. [Campbell Soup Company Stock Price](https://www.perplexity.ai/finance/CPB?comparing=CPB,KHC,BMBOY,CAG,UTZ,\\SMPL)\par
\vspace{1.5pt}
\noindent Campbell's reaffirmed its fiscal 2026 guidance, projecting adjusted EPS of \$2.15 to \$2.25 and organic sales decline of only 1\% to 2\%. Q3 adjusted EPS of \$0.50 ...\par
\vspace{1.5pt}
\noindent 3. [Campbell's Reports Third Quarter Fiscal 2026 Results](https://www.thecampbellscompany.com/newsroom/news/campbells-reports-third-quarter-fiscal-2026-results/)\par
\noindent Date published: Jun 8, 2026\par
\vspace{1.5pt}
\noindent Earnings Per Share (EPS) increased to \$0.41; Adjusted EPS decreased 32\% to \$0.50. Fiscal year-to-date cash flow from operations was \$839 million ...\par
\vspace{1.5pt}
\noindent 4. [Campbell's (CPB)](https://www.trefis.com/data/companies/CPB)\par
\vspace{1.5pt}
\noindent Campbell's (CPB) reported adjusted earnings per share (EPS) of \$0.50 $\cdot$ slightly surpassing analyst consensus estimates of \$0.48. net sales decreased 4\% to \$2. ...\par
\vspace{1.5pt}
\noindent 5. [Exploring Analyst Estimates for Campbell (CPB) Q4 ...](https://www.zacks.com/stock/news/2982391/exploring-analyst-estimates-for-campbell-cpb-q4-earnings-beyond-revenue-and-eps)\par
\noindent Date published: 2 days ago\par
\vspace{1.5pt}
\noindent Campbell's ( CPB Quick Quote CPB to post quarterly earnings of \$0.40 per share in its upcoming report, a year-over-year decline of 35.5\%. ...\par
\vspace{1.5pt}
\noindent 6. [5 Must-Read Analyst Questions From Campbell's's Q4 ...](https://stockstory.org/us/stocks/nasdaq/cpb/news/earnings-call/5-must-read-analyst-questions-from-campbellss-q4-earnings-call)\par
\noindent Date published: Mar 17, 2026\par
\vspace{1.5pt}
\noindent Adjusted EPS: \$0.51 vs analyst expectations of \$0.57 (10.6\% miss) Adjusted EPS guidance to \$2.20 at the midpoint, a 11.1\% decrease\par
\vspace{1.5pt}
\noindent 7. [Campbell's (Nasdaq:CPB) - Stock Analysis](https://simplywall.st/stocks/us/food-beverage-tobacco/nasdaq-cpb/campbells)\par
\vspace{1.5pt}
\noindent Third quarter 2026 results: EPS: US\$0.42 (up from US\$0.22 in 3Q 2025). Revenue: US\$2.37b (down 4.4\% from 3Q 2025). Net income: US\$124.0m (up 88\% from 3Q 2025).\par
\vspace{1.5pt}
\noindent 8. [Campbell's Q3 Earnings: CPB Revenue Misses as EPS ...](https://www.chartmill.com/news/CPB/Chartmill-49565-Campbells-Q3-Earnings-CPB-Revenue-Misses-as-EPS-Slightly-Beats-Amidst-Cost-Pressure)\par
\noindent Date published: Jun 8, 2026\par
\vspace{1.5pt}
\noindent Earnings Per Share (EPS): On an adjusted basis, Campbell's reported EPS of \$0.50. This figure came in slightly above the analyst estimate of ...\par
\noindent =======\par
\noindent A Google search for 'Campbell's fourth quarter fiscal 2025 results September 2025 adjusted EPS' found 8 results:\par
\vspace{1.5pt}
\noindent \#\# Web Results\par
\noindent 1. [Campbell's Reports Fourth Quarter Fiscal 2025 Results](https://investor.thecampbellscompany.com/news-releases/news-release-details/campbells-reports-fourth-quarter-fiscal-2025-results-provides/)\par
\noindent Date published: Sep 3, 2025\par
\vspace{1.5pt}
\noindent Earnings Per Share (EPS) increased to \$0.48 . Adjusted EPS decreased 2\% to \$0.62 . For the full year: Net Sales increased 6\% to \$10.3 billion ...\par
\vspace{1.5pt}
\noindent 2. [Campbell's Reports Third Quarter Fiscal 2026 Results](https://www.businesswire.com/news/home/20260605204577/en/Campbells-Reports-Third-Quarter-Fiscal-2026-Results)\par
\noindent Date published: Jun 8, 2026\par
\vspace{1.5pt}
\noindent Earnings Per Share (EPS) increased to \$0.41; Adjusted EPS decreased 32\% to \$0.50. Fiscal year-to-date cash flow from operations was \$839 million ...\par
\vspace{1.5pt}
\noindent 3. [Campbell's (CPB) Q4 2025 Earnings Call Transcript](https://fortune.com/company/campbell-soup/earnings/q4-2025/)\par
\noindent Date published: Sep 3, 2025\par
\vspace{1.5pt}
\noindent Adjusted EPS decreased 4\%, driven by increased interest expense. The modest Q4 headwind from tariffs and the full-year impact of divestitures ...\par
\vspace{1.5pt}
\noindent 4. [Investor Relations - The Campbell's Company](https://investor.thecampbellscompany.com/investor-relations/)\par
\vspace{1.5pt}
\noindent Fiscal Year 2025 Net sales \$10.3 Dollars in Billions Adjusted Earnings* \$2.97 Per Share Operating Cash Flow \$1.1 Dollars in Billions Latest Quarterly Results\par
\vspace{1.5pt}
\noindent 5. [Campbell's Reports Fourth Quarter Fiscal 2025 Results](https://www.theglobeandmail.com/investing/markets/markets-news/Business\%20Wire/34563254/campbell-s-reports-fourth-quarter-fiscal-2025-results-provides-full-year-fiscal-2026-guidance/)\par
\noindent Date published: Sep 3, 2025\par
\vspace{1.5pt}
\noindent ... adjusted EBIT and 10\% to adjusted EPS, or \$0.06 per share, to fourth quarter fiscal 2025 results. Net sales in the quarter increased 1\% to ...\par
\vspace{1.5pt}
\noindent 6. [Campbell's Q4 Earnings Beat Estimates, Organic Sales ...](https://finance.yahoo.com/news/campbells-q4-earnings-beat-estimates-165500807.html)\par
\noindent Date published: Sep 3, 2025\par
\vspace{1.5pt}
\noindent Adjusted earnings per share (EPS) were 62 cents, down 2\% year over year, due to lower adjusted earnings before interest and taxes (EBIT).\par
\vspace{1.5pt}
\noindent 7. [Campbell's (NasdaqGS:CPB) Stock Forecast \& Analyst ...](https://simplywall.st/stocks/us/food-beverage-tobacco/nasdaq-cpb/campbells/future)\par
\noindent Date published: 3 days ago\par
\vspace{1.5pt}
\noindent EPS: US\$0.42 (up from US\$0.22 in 3Q 2025). Revenue: US\$2.37b (down 4.4\% from 3Q 2025). Net income: US\$124.0m (up 88\% from 3Q 2025). Profit ...\par
\vspace{1.5pt}
\noindent 8. [Campbell's Reports Fourth Quarter Fiscal 2025 Results](https://www.tradingview.com/news/tradingview:2f8cba08c6512:0-campbell-s-reports-fourth-quarter-fiscal-2025-results/)\par
\noindent Date published: Sep 3, 2025\par
\vspace{1.5pt}
\noindent Earnings Per Share (EPS) increased to \$0.48, with adjusted EPS decreasing by 2\% to \$0.62. For the full year, net sales grew by 6\% to \$10.3 ...\par
\noindent =======\par
\noindent A Google search for 'CPB earnings whisper number September 2026' found 9 results:\par
\vspace{1.5pt}
\noindent \#\# Web Results\par
\noindent 1. [Earnings Whispers](https://www.earningswhispers.com/)\par
\vspace{1.5pt}
\noindent The Earnings Whisper number was \$0.39 per share. The company beat expectations by 23.08\% while revenue grew 33.04\% on a year-over-year basis.\par
\vspace{1.5pt}
\noindent 2. [What date does The Campbell's Company's (CPB) report ...](https://www.zacks.com/stock/research/CPB/earnings-calendar)\par
\vspace{1.5pt}
\noindent Look out for CPB's next earnings release on September 03, 2026. For the next earning release, we expect the company to report earnings of \$0.40 per share ...\par
\vspace{1.5pt}
\noindent 3. [Campbell's Reports Second Quarter Fiscal 2026 Results](https://www.thecampbellscompany.com/newsroom/news/campbells-reports-second-quarter-fiscal-2026-results/)\par
\noindent Date published: Mar 11, 2026\par
\vspace{1.5pt}
\noindent Net Sales decreased 5\% to \$2.6 billion and decreased 3\% on an organic basis. $\cdot$ Earnings Before Interest and Taxes (EBIT) decreased to \$273 ...\par
\vspace{1.5pt}
\noindent 4. [Campbell's (CPB) Surpasses Q3 Earnings Estimates](https://finance.yahoo.com/markets/stocks/articles/campbells-cpb-surpasses-q3-earnings-122501965.html)\par
\noindent Date published: Jun 8, 2026\par
\vspace{1.5pt}
\noindent Campbell (CPB) delivered earnings and revenue surprises of +4.17\% and -0.86\%, respectively, for the quarter ended April 2026. Do the numbers ...\par
\vspace{1.5pt}
\noindent 5. [The Campbell's Company Common Stock (CPB) Earnings](https://www.nasdaq.com/market-activity/stocks/cpb/earnings)\par
\vspace{1.5pt}
\noindent Find annual and quearterly earnings data for The Campbell's Company Common Stock (CPB) including earnings per share, earnings forecasts at Nasdaq.com.\par
\vspace{1.5pt}
\noindent 6. [The most anticipated earnings releases for the week ...](https://www.facebook.com/EarningsWhispers/posts/the-most-anticipated-earnings-releases-for-the-week-of-june-15-2026-are-kroger-k/1425572336275486/)\par
\noindent Date published: 2 months ago\par
\vspace{1.5pt}
\noindent The most anticipated earnings releases for the week of June 15, 2026, are Kroger \#KR, Progressive \#PGR, Accenture. The most anticipated ...\par
\vspace{1.5pt}
\noindent 7. [r/EarningsWhisper](https://www.reddit.com/r/EarningsWhisper/)\par
\vspace{1.5pt}
\noindent Providing key trading information around earnings releases. Earnings Whisper Grades during the Second Quarter 2026. Earnings Whisper Grades during the Second ...\par
\vspace{1.5pt}
\noindent 8. [Campbell's (CPB) Earnings Date and Reports 2026](https://www.marketbeat.com/stocks/NASDAQ/CPB/earnings/)\par
\vspace{1.5pt}
\noindent Campbell's posted Q3 2026 \ldots{} reporting an EPS of \$0.50, \$0.48 by \$0.02. Quarterly revenue fell 4.4\% year-over-year to \$2.37 billion. from \$2.18 to \$1.97 per ...\par
\vspace{1.5pt}
\noindent 9. [Campbell Soup (CPB) Earnings: Latest Report, Earnings Call ...](https://public.com/stocks/cpb/earnings)\par
\vspace{1.5pt}
\noindent For Q3 2026, Campbell Soup (CPB) reported an EPS of \$0.50, exceeding analysts' estimate of \$0.48 by 4.17\%. Did Campbell Soup (CPB) beat earnings expectations ...\par
\end{msgbox}
\begin{msgbox}{casegreen}
\casetag{Thinking}
\noindent Now I have a clearer picture:\par
\vspace{1.5pt}
\noindent Key facts:\par
\noindent 1. Campbell's reports Q4 fiscal 2026 on September 3, 2026 (quarter ended August 2, 2026).\par
\noindent 2. Consensus EPS estimates:\par
\noindent    - Zacks: \$0.40\par
\noindent    - MarketWatch: average \$0.39 (high \$0.40, low \$0.37), trend: 3 months ago \$0.41, 1 month ago \$0.39, current \$0.39\par
\noindent    - Yahoo: "consensus mark for earnings has remained unchanged over the past 30 days at 40 cents a share" --- implies \$0.40\par
\noindent    - EarningsWhispers: \$0.39\par
\noindent 3. Year-over-year: consensus implies a decline of 35.5\% from Q4 fiscal 2025's adjusted EPS of \$0.62. Wait, 35.5\% decline from \$0.62 would be \$0.62 $\times$ 0.645 = \$0.40. So the \$0.40 consensus. But \$0.39 would be a 37\% decline. The discrepancy: Yahoo says "EPS drop of 35.5\%" which corresponds to \$0.40 (from \$0.62). MarketWatch average \$0.39.\par
\vspace{1.5pt}
\noindent Hmm, wait. Let me reconsider. Yahoo article says "expectations of a 7.3\% revenue decline to \$2.15 billion and an EPS drop of 35.5\%". If prior year Q4 adjusted EPS was \$0.62, a 35.5\% drop $\rightarrow$ \$0.40. Zacks says \$0.40. Yahoo "Q4 Earnings on the Horizon" says consensus unchanged at 40 cents over past 30 days. MarketWatch average is \$0.39. So the consensus is around \$0.39-\$0.40.\par
\vspace{1.5pt}
\noindent The threshold is \$0.39, and the market resolves YES if non-GAAP EPS \(> \$0.39\), i.e., $\geq$ \$0.40.\par
\vspace{1.5pt}
\noindent Key considerations:\par
\noindent - Company guidance: In March 2026 (Q2 FY2026 call), adjusted EPS guidance to \$2.20 at midpoint for FY2026 (11.1\% decrease). Q3 adjusted EPS came in at \$0.50, beating \$0.48 estimate.\par
\noindent - Q4 FY2025 adjusted EPS was \$0.62. A decline to \$0.39-\$0.40 would be a 35-37\% decline --- that's a big drop, consistent with the guidance of \textasciitilde{}11\% full-year decline plus seasonal weakness in Q4? Actually wait, Q4 is a strong season for soup (holiday). Hmm.\par
\vspace{1.5pt}
\noindent Actually let me think about the seasonal pattern of Campbell's adjusted EPS:\par
\noindent - Q4 FY2025 (ended Jul 31, 2025): \$0.62\par
\noindent - Q1 FY2026 (ended Oct 31, 2025): \$0.77\par
\noindent - Q2 FY2026 (ended Jan 31, 2026): ? (Q2 guidance was adjusted EPS... in March 2026 they guided FY2026 adjusted EPS \$2.15-\$2.25)\par
\noindent - Q3 FY2026 (ended Apr 30, 2026): \$0.50\par
\vspace{1.5pt}
\noindent FY2026 guidance midpoint \$2.20. If Q1 = 0.77, Q2 = \textasciitilde{}0.61 (need to check), Q3 = 0.50, then Q4 implied = 2.20 - 0.77 - 0.61 - 0.50 = 0.32? That seems low. Hmm, but the consensus for Q4 is \$0.39-\$0.40. Let me check Q2 FY2026 actual.\par
\vspace{1.5pt}
\noindent Actually from the search: Q2 FY2026 (reported Mar 11, 2026): "Net Sales decreased 5\% to \$2.6 billion... EBIT decreased to \$273 million". I don't have Q2 EPS yet. Let me check.\par
\vspace{1.5pt}
\noindent If FY2026 guidance is \$2.15-\$2.25 adjusted EPS, and Q1-Q3 sum: 0.77 + Q2 + 0.50. If Q2 was \textasciitilde{}0.60, sum = 1.87, leaving 0.28-0.38 for Q4. Consensus \$0.39-\$0.40 implies Q2 would need to be lower or guidance was conservative.\par
\vspace{1.5pt}
\noindent Actually wait --- I should check Q2 FY2026 EPS. Trading economics said "FY2026Q1, Oct/2025, 0.77, 0.74, 0.89". So Q1 FY2026 actual = 0.77, estimate 0.74, prior 0.89.\par
\vspace{1.5pt}
\noindent Let me search for Q2 FY2026 (ended Jan 31, 2026) EPS.\par
\vspace{1.5pt}
\noindent Also, important: the whisper number from earningswhispers.com is \$0.39. And the company's own guidance. Let me also check the Zacks Earnings ESP --- whether analysts expect a beat or miss. Zacks typically has an "Earnings ESP" metric. The Yahoo/Zacks article said "we expect the company to report earnings of \$0.40 per share" --- the Zacks article title says "Exploring Analyst Estimates... Beyond Revenue and EPS".\par
\vspace{1.5pt}
\noindent Another important consideration: the resolution source. "sourced from the company's official earnings documents, or from SeekingAlpha if not reported by the company." Campbell's reports adjusted EPS in its official press release, so the official document will be the source. The question says "non-GAAP EPS" --- Campbell's reports "Adjusted EPS". That's what counts.\par
\vspace{1.5pt}
\noindent Now, will the actual be \(> \$0.39\)? The consensus is right at \$0.39-\$0.40. This is essentially a coin flip around the threshold, but let me gather more evidence:\par
\vspace{1.5pt}
\noindent 1. Historical beat/miss pattern vs consensus: Campbell's has been beating recently (Q3 beat by 4.17\%, Q1 beat).\par
\noindent 2. Earnings ESP from Zacks.\par
\noindent 3. Analyst commentary --- any pre-announcements.\par
\noindent 4. The revision trend is downward (from \$0.41 to \$0.39).\par
\noindent 5. Seasonal: Q4 is Campbell's strongest quarter typically (holiday soups). Last year Q4 adjusted EPS \$0.62 vs Q3 \$0.73 (wait, that's odd --- Q3 FY2025 adjusted EPS was \$0.73? That seems high. Actually Q3 FY2025 (June 2025) adjusted EPS \$0.73 --- yes per businesswire: "Adjusted. \$0.50. \$0.73" for Q3 FY2026 vs Q3 FY2025. Hmm, and GAAP Q3 FY2025 was \$0.22. Interesting --- last year Q3 had some benefit. Actually wait, that seems inconsistent with Q4 FY2025 \$0.62. Seasonal pattern: Q3 FY2025 \$0.73, Q4 FY2025 \$0.62. Hmm, Q4 lower than Q3? That's unusual but possible given tariff impacts in 2025.\par
\vspace{1.5pt}
\noindent Actually wait, let me re-read: "Campbell's Reports Third Quarter Fiscal 2026 Results; Diluted Earnings Per Share; As Reported (GAAP). \$0.41. \$0.22; Adjusted. \$0.50. \$0.73." So Q3 FY2025 adjusted EPS was \$0.73. And Q4 FY2025 adjusted EPS was \$0.62.\par
\vspace{1.5pt}
\noindent So the year-ago Q4 base is \$0.62. Consensus \$0.39-\$0.40 implies a 35-37\% YoY decline. That's a big drop. Is it plausible? FY2026 guidance is adjusted EPS \$2.15-\$2.25 vs FY2025's \$2.97 --- an \textasciitilde{}26\% decline. Quarterly declines: Q1 FY2026 was \$0.77 vs \$0.89 prior year (-13.5\%); Q2 FY2026 unknown; Q3 FY2026 \$0.50 vs \$0.73 (-31.5\%). The declines are accelerating, driven by tariffs, divestitures (they sold some businesses), and volume declines.\par
\vspace{1.5pt}
\noindent Hmm, actually there's something notable: "Adjusted EPS decreased 32\% to \$0.50" in Q3. And in Q4, consensus implies -35.5\% YoY. The full year guidance of \textasciitilde{}-26\% at midpoint... if Q1 -13.5\%, Q3 -31.5\%, then Q2 and Q4 would need steeper declines to average -26\%. Hmm, actually the FY2025 full year was \$2.97. 26\% decline $\rightarrow$ \textasciitilde{}\$2.20. OK consistent with guidance \$2.15-\$2.25.\par
\vspace{1.5pt}
\noindent Let me estimate: FY2025 quarters: Q1 0.89, Q2 ?, Q3 0.73, Q4 0.62. Sum = 2.97 $\rightarrow$ Q2 FY2025 = 2.97 - 0.89 - 0.73 - 0.62 = 0.73. So Q2 FY2025 was \textasciitilde{}0.73.\par
\vspace{1.5pt}
\noindent FY2026: Q1 0.77 (-13.5\%), Q2 \textasciitilde{}? (consensus for Q1 2027 is \$0.58 per MarketWatch --- that's Q1 FY2027, i.e., Oct 2026 quarter, not relevant). If FY2026 guidance midpoint 2.20, and Q1 0.77, Q3 0.50, then Q2 + Q4 = 0.93. If Q4 consensus is 0.39-0.40, Q2 would be \textasciitilde{}0.53-0.56. Plausible (Q2 FY2025 was 0.73, so -25\%).\par
\vspace{1.5pt}
\noindent OK so the consensus \$0.39-\$0.40 for Q4 seems consistent with guidance.\par
\vspace{1.5pt}
\noindent Now the question: will actual be \(> \$0.39\) (i.e., $\geq$ \$0.40)?\par
\vspace{1.5pt}
\noindent Factors favoring YES (beat):\par
\noindent - Recent beat pattern: Q1 FY2026 beat (0.77 vs 0.74 est), Q3 FY2026 beat (0.50 vs 0.48-0.49 est).\par
\noindent - Whisper number: \$0.39 (not above threshold). Hmm, whisper \$0.39 means even whisper is at threshold, not above.\par
\noindent - Zacks consensus \$0.40.\par
\vspace{1.5pt}
\noindent Factors favoring NO (miss or in-line):\par
\noindent - Consensus revised down 2.4\% over 30 days.\par
\noindent - MarketWatch average \$0.39 exactly.\par
\noindent - EarningsWhispers \$0.39.\par
\noindent - Tariff pressures, volume declines, organic sales declines expected (-7.3\% revenue).\par
\noindent - Big YoY decline expected (-35.5\%), suggesting analysts have cut estimates heavily; the bar is the consensus itself.\par
\vspace{1.5pt}
\noindent The key nuance: the threshold \$0.39 vs consensus \$0.39 (MarketWatch average) or \$0.40 (Zacks/Yahoo). If the true consensus is \$0.39, then a "beat" to \$0.40 $\rightarrow$ YES. If true consensus is \$0.40, then in-line \$0.40 $\rightarrow$ YES (since \(> \$0.39\)). Actually even an in-line \$0.40 beats \$0.39. The problem is if actual comes in at \$0.39 or below.\par
\vspace{1.5pt}
\noindent Given the estimate range (high \$0.40, low \$0.37, average \$0.39), the distribution of outcomes: P(actual $\geq$ \$0.40) maybe \textasciitilde{}35-45\%? Historically, companies beat consensus \textasciitilde{}70\% of the time, but "beat by enough to exceed a round threshold" is different. Here the threshold is one cent below the consensus range.\par
\vspace{1.5pt}
\noindent Hmm, wait. Let me think about this more carefully. The question is Polymarket-style: "Will CPB report non-GAAP EPS greater than \$0.39?" The resolution uses the company's official adjusted EPS. Company guides... let me check if Campbell's guided Q4 specifically. In March 2026, they guided FY2026 adjusted EPS \$2.15-\$2.25. I don't think they guided Q4 specifically.\par
\vspace{1.5pt}
\noindent Let me check the Zacks Earnings ESP for this report --- Zacks usually publishes "CPB Earnings ESP" in their preview. Also check Seeking Alpha's consensus (the resolution source fallback). Let me search for those.\par
\vspace{1.5pt}
\noindent Also worth checking: recent news in August 2026 --- any analyst changes, tariff updates, or company updates. And the StockStory note from March: "Adjusted EPS: \$0.51 vs analyst expectations of \$0.57 (10.6\% miss), Adjusted EPS guidance to \$2.20 at midpoint" --- wait, that says Q4 FY2025 (March 2026 call? no). Hmm, "5 Must-Read Analyst Questions From Campbell's's Q4 Earnings Call" dated Mar 17, 2026 --- that's confusing. Q4 FY2025 call would be September 2025. March 17, 2026 would be Q2 FY2026 call. The StockStory title says "Q4 Earnings Call" but dated Mar 17, 2026 --- maybe mislabeled; content: "Adjusted EPS: \$0.51 vs analyst expectations of \$0.57 (10.6\% miss)" --- hmm, that doesn't match Q2 FY2026 (reported Mar 11, 2026)... Actually Q2 FY2026 EPS --- I haven't found it. Could be \$0.51? And guidance \$2.20 midpoint. Hmm wait, but \$0.51 vs \$0.57 miss in Q2? Let me verify Q2 FY2026 results.\par
\vspace{1.5pt}
\noindent Actually wait, re-read: stockstory.org "5 Must-Read Analyst Questions From Campbell's's Q4 Earnings Call" dated Mar 17, 2026 --- the date might be wrong or the label might be wrong. If Q2 FY2026 adjusted EPS was \$0.51... then FY2026: Q1 0.77 + Q2 0.51 + Q3 0.50 = 1.78, leaving Q4 = 2.20 - 1.78 = 0.42 at midpoint. Hmm, that would make Q4 consensus \$0.39-0.40 a bit below implied. Interesting.\par
\vspace{1.5pt}
\noindent Let me search for Q2 FY2026 results and Zacks Earnings ESP.\par
\end{msgbox}
\begin{msgbox}{caseblue}
\casetag{Tool call}
\noindent\texttt{search}
\begin{Verbatim}[   fontsize=\scriptsize,   breaklines=true,   breakanywhere=true,   breaksymbolleft={},   xleftmargin=0pt ]
{"query": ["Campbell's second quarter fiscal 2026 results March 2026 adjusted EPS", "CPB Zacks earnings ESP Q4 2026", "Campbell's CPB earnings preview September 3 2026 beat miss expectation"]}
\end{Verbatim}
\end{msgbox}
\begin{msgbox}{casegray}
\casetag{Tool result}
\noindent A Google search for 'Campbell's second quarter fiscal 2026 results March 2026 adjusted EPS' found 8 results:\par
\vspace{1.5pt}
\noindent \#\# Web Results\par
\noindent 1. [Campbell's Reports Second Quarter Fiscal 2026 Results](https://investor.thecampbellscompany.com/news-releases/news-release-details/campbells-reports-second-quarter-fiscal-2026-results/)\par
\noindent Date published: Mar 11, 2026\par
\vspace{1.5pt}
\noindent Earnings Per Share (EPS) decreased to \$0.48 . Adjusted EPS decreased 31\% to \$0.51 . adjusted EPS by approximately \$0.04 per share in the ...\par
\vspace{1.5pt}
\noindent 2. [Campbell's Reports Third Quarter Fiscal 2026 Results](https://www.thecampbellscompany.com/newsroom/news/campbells-reports-third-quarter-fiscal-2026-results/)\par
\noindent Date published: Jun 8, 2026\par
\vspace{1.5pt}
\noindent Earnings Per Share (EPS) increased to \$0.41; Adjusted EPS decreased 32\% to \$0.50. Fiscal year-to-date cash flow from operations was \$839 million ...\par
\vspace{1.5pt}
\noindent 3. [Campbell's Reports Third Quarter Fiscal 2026 Results](https://finance.yahoo.com/markets/stocks/articles/campbells-reports-third-quarter-fiscal-111500594.html)\par
\noindent Date published: Jun 8, 2026\par
\vspace{1.5pt}
\noindent Earnings Per Share (EPS) increased to \$0.41; Adjusted EPS decreased 32\% to \$0.50. Fiscal year-to-date cash flow from operations was \$839 ...\par
\vspace{1.5pt}
\noindent 4. [Campbell's (CPB) Q2 2026 Earnings Call Transcript](https://fortune.com/company/campbell-soup/earnings/q2-2026/)\par
\noindent Date published: 4 days ago\par
\vspace{1.5pt}
\noindent Campbell's (CPB) Q2 2026 Earnings Call March 11, 2026 at 4:00 PM EDT Company. The Campbell's Company second quarter 2026 question and answer ...\par
\vspace{1.5pt}
\noindent 5. [The Campbell's Company to Report Fourth Quarter and ...](https://www.businesswire.com/news/home/20260813498006/en/The-Campbells-Company-to-Report-Fourth-Quarter-and-Full-Year-Fiscal-2026-Results-on-September-3-2026)\par
\noindent Date published: Aug 13, 2026\par
\vspace{1.5pt}
\noindent The Campbell's Company (NASDAQ:CPB) (Campbell's) announced it will report its financial results on September 3, 2026 for the fourth quarter ...\par
\vspace{1.5pt}
\noindent 6. [Campbell's (CPB) Earnings Date and Reports 2026](https://www.marketbeat.com/stocks/NASDAQ/CPB/earnings/)\par
\vspace{1.5pt}
\noindent With a trailing EPS of \$2.02 and a P/E Ratio of 11.74, Campbell's' earnings are expected to decrease -9.63\% next year, from \$2.18 to \$1.97 per share.\par
\vspace{1.5pt}
\noindent 7. [Earnings call transcript: Campbell's Q3 2026 results show ...](https://www.investing.com/news/transcripts/earnings-call-transcript-campbells-q3-2026-results-show-slight-eps-beat-93CH-4730697)\par
\noindent Date published: Jun 8, 2026\par
\vspace{1.5pt}
\noindent Adjusted EPS declined to \$0.50, down 32\% versus prior year, driven by lower adjusted EBIT. Interest expense, the adjusted effective tax rate, ...\par
\vspace{1.5pt}
\noindent 8. [Campbell's Reports Second Quarter Fiscal 2026 Results](https://investor.thecampbellscompany.com/node/24831/pdf)\par
\noindent Date published: Mar 11, 2026\par
\vspace{1.5pt}
\noindent Adjusted EBIT decreased 24\% to \$282 million. Earnings Per Share (EPS) decreased to \$0.48. Adjusted EPS decreased 31\% to \$0.51. Fiscal year-to- ...\par
\noindent =======\par
\noindent A Google search for 'CPB Zacks earnings ESP Q4 2026' found 9 results:\par
\vspace{1.5pt}
\noindent \#\# Web Results\par
\noindent 1. [Exploring Analyst Estimates for Campbell (CPB) Q4 ...](https://www.zacks.com/stock/news/2982391/exploring-analyst-estimates-for-campbell-cpb-q4-earnings-beyond-revenue-and-eps)\par
\noindent Date published: 2 days ago\par
\vspace{1.5pt}
\noindent The consensus EPS estimate for the quarter has been revised 2.4\% lower over the last 30 days to the current level. This reflects how the ...\par
\vspace{1.5pt}
\noindent 2. [What date does The Campbell's Company's (CPB) report ...](https://www.zacks.com/stock/research/CPB/earnings-calendar)\par
\vspace{1.5pt}
\noindent The Zacks Consensus Estimate for The Campbell's Company (CPB) for the quarter ending in July 2026 is \$0.40 a share. We expect The Campbell's Company to miss by ...\par
\vspace{1.5pt}
\noindent 3. [Campbell's (CPB) Q4 Earnings Expected to Decline](https://www.zacks.com/stock/news/2981033/earnings-preview-campbells-cpb-q4-earnings-expected-to-decline)\par
\noindent Date published: 6 days ago\par
\vspace{1.5pt}
\noindent Earnings Preview: Campbell's (CPB) Q4 Earnings Expected to Decline. Zacks Equity Research August 27, 2026. CPB Quick Quote CPB. Zack's Triangle ...\par
\vspace{1.5pt}
\noindent 4. [Campbell's Q4 Earnings on the Horizon](https://www.tradingview.com/news/zacks:6776041a8094b:0-campbell-s-q4-earnings-on-the-horizon-what-should-investors-expect/)\par
\noindent Date published: 2 days ago\par
\vspace{1.5pt}
\noindent Company CPB is likely to witness a top and bottom-line decline when it reports fourth-quarter fiscal 2026 earnings on Sept. 3. The Zacks ...\par
\vspace{1.5pt}
\noindent 5. [The Campbell's Company (CPB)](https://www.zacks.com/stock/research/CPB/company-reports)\par
\vspace{1.5pt}
\noindent Sector $\cdot$ Consumer Staples ; Fiscal Year End, July ; Last Reported Quarter, 7/31/2026 ; Current (1=Strong Buy, 5=Strong Sell), 3.55 ; 30 Days Ago, 3.65.\par
\vspace{1.5pt}
\noindent 6. [Campbell's Q4 Earnings on the Horizon](https://finance.yahoo.com/markets/stocks/articles/campbells-q4-earnings-horizon-investors-144700799.html)\par
\noindent Date published: 2 days ago\par
\vspace{1.5pt}
\noindent CPB is likely \ldots{} reports fourth-quarter fiscal 2026 earnings on Sept. 3. The Zacks Consensus Estimate for revenues is pegged at \$2.15 billion,\par
\vspace{1.5pt}
\noindent 7. [The Campbell's Company (CPB) Eps Diluted](https://www.zacks.com/stock/chart/CPB/fundamental/eps-diluted)\par
\vspace{1.5pt}
\noindent Campbell's Q4 Earnings on the Horizon: Earnings Expected to Decline 08/27/26-9:00AM EST. Maintained at \$21 08/31/26-8:38AM. Lowered to \$22\par
\vspace{1.5pt}
\noindent 8. [Should I buy The Campbell's Company (CPB)](https://www.zacks.com/stock/research/CPB/stock-style-scores)\par
\vspace{1.5pt}
\noindent The Campbell's Company is currently rated as a Zacks Rank 4 \ldots{} for Campbell (CPB) Q4 Earnings, Beyond Revenue and EPS 08/31/26-8:15AM EST. Earnings Expected to ...\par
\vspace{1.5pt}
\noindent 9. [Campbell (CPB) Up 3.5\% Since Last Earnings Report](https://www.zacks.com/stock/news/2950337/campbell-cpb-up-35-since-last-earnings-report-can-it-continue)\par
\noindent Date published: Jul 8, 2026\par
\vspace{1.5pt}
\noindent Adjusted gross profit declined 12\% to \$656 million. Adjusted gross margin contracted 240 basis points (bps) to 27.7\%, mainly due to cost ...\par
\noindent =======\par
\noindent A Google search for 'Campbell's CPB earnings preview September 3 2026 beat miss expectation' found 8 results:\par
\vspace{1.5pt}
\noindent \#\# Web Results\par
\noindent 1. [Campbell's (CPB) Q4 Earnings Expected to Decline](https://finance.yahoo.com/markets/stocks/articles/earnings-preview-campbells-cpb-q4-140002239.html)\par
\noindent Date published: 6 days ago\par
\vspace{1.5pt}
\noindent Campbell (CPB) doesn't possess the right combination of the two key ingredients for a likely earnings beat in its upcoming report.\par
\vspace{1.5pt}
\noindent 2. [Campbell Soup (CPB) Earnings: Latest Report, Earnings Call ...](https://public.com/stocks/cpb/earnings)\par
\vspace{1.5pt}
\noindent For Q3 2026, Campbell Soup (CPB) reported an EPS of \$0.50, exceeding analysts' estimate of \$0.48 by 4.17\%. Did Campbell Soup (CPB) beat earnings expectations ...\par
\vspace{1.5pt}
\noindent 3. [The Campbell's Company to Report Fourth Quarter and ...](https://www.thecampbellscompany.com/newsroom/press-releases/the-campbells-company-to-report-fourth-quarter-and-full-year-fiscal-2026-results-on-september-3-2026/)\par
\noindent Date published: Aug 13, 2026\par
\vspace{1.5pt}
\noindent The Campbell's Company to Report Fourth Quarter and Full Year Fiscal 2026 Results on September 3, 2026.\par
\vspace{1.5pt}
\noindent 4. [Campbell's (CPB) Earnings Date and Reports 2026](https://www.marketbeat.com/stocks/NASDAQ/CPB/earnings/)\par
\vspace{1.5pt}
\noindent Campbell's posted Q3 2026 \ldots{} an EPS of \$0.50, Campbell's' earnings are expected to decrease -9.63\% next year, from \$2.18 to \$1.97 per share. Beat/Miss GAAP\par
\vspace{1.5pt}
\noindent 5. [The Campbell's Company to Report Fourth Quarter and ...](https://www.businesswire.com/news/home/20260813498006/en/The-Campbells-Company-to-Report-Fourth-Quarter-and-Full-Year-Fiscal-2026-Results-on-September-3-2026)\par
\noindent Date published: Aug 13, 2026\par
\vspace{1.5pt}
\noindent The Campbell's Company (NASDAQ:CPB) (Campbell's) announced it will report its financial results on September 3, 2026 for the fourth quarter ...\par
\vspace{1.5pt}
\noindent 6. [Campbell's (NasdaqGS:CPB) Stock Forecast \& Analyst ...](https://simplywall.st/stocks/us/food-beverage-tobacco/nasdaq-cpb/campbells/future)\par
\noindent Date published: 3 days ago\par
\vspace{1.5pt}
\noindent Second quarter 2026 earnings: EPS and revenues miss analyst expectations. Second quarter 2026 results: EPS: US\$0.49 (down from US\$0.58 in 2Q ...\par
\vspace{1.5pt}
\noindent 7. [Campbell's Q3 Earnings: CPB Revenue Misses as EPS ...](https://www.chartmill.com/news/CPB/Chartmill-49565-Campbells-Q3-Earnings-CPB-Revenue-Misses-as-EPS-Slightly-Beats-Amidst-Cost-Pressure)\par
\noindent Date published: Jun 8, 2026\par
\vspace{1.5pt}
\noindent Campbell\&\#39;s Q3 results showed revenue missing estimates and EPS slightly beating, with shares dropping 2.3\% Reported net sales of \$2.366 ...\par
\vspace{1.5pt}
\noindent 8. [Campbell's (CPB) Earnings Date \& Report](https://www.investing.com/equities/campbell-soup-earnings)\par
\vspace{1.5pt}
\noindent Campbell's Q3 FY2026 EPS of \$0.50 beat estimates by \$0.01; revenue reached \$2.4B, exceeding forecasts despite 4\% organic sales decline. $\cdot$ Adjusted EBIT plunged ...\par
\end{msgbox}
\begin{msgbox}{casegreen}
\casetag{Thinking}
\noindent More key facts:\par
\vspace{1.5pt}
\noindent 1. Q2 FY2026 (reported Mar 11, 2026): Adjusted EPS \$0.51, down 31\% YoY (prior year Q2 FY2025 was \$0.73). GAAP EPS \$0.48.\par
\vspace{1.5pt}
\noindent 2. Zacks earnings calendar: "The Zacks Consensus Estimate for the quarter ending in July 2026 is \$0.40 a share. We expect The Campbell's Company to miss by..." --- truncated, but Zacks expects a miss? Let me read that page.\par
\vspace{1.5pt}
\noindent 3. Zacks article (Aug 27, 2026): "Campbell (CPB) doesn't possess the right combination of the two key ingredients for a likely earnings beat in its upcoming report." --- suggests Zacks expects a miss/beat unlikely.\par
\vspace{1.5pt}
\noindent 4. StockStory (Mar 17, 2026, labeled Q4 Earnings Call but actually Q2): "Adjusted EPS: \$0.51 vs analyst expectations of \$0.57 (10.6\% miss)" --- so Q2 FY2026 MISSED consensus by 10.6\%! And Q2 FY2026 guidance given was adjusted EPS \textasciitilde{}\$2.20 midpoint FY2026.\par
\vspace{1.5pt}
\noindent Wait, so Q2 FY2026: actual adjusted EPS \$0.51 vs consensus \$0.57 $\rightarrow$ miss of 10.6\%. Hmm, but simplywall.st says "Second quarter 2026 earnings: EPS and revenues miss analyst expectations. EPS: US\$0.49 (down from US\$0.58 in 2Q...)". Different numbers (\$0.49 vs \$0.51). The company's adjusted EPS was \$0.51. Simplewall may use a different figure (maybe GAAP or a different currency adjustment). Anyway, Q2 was a miss vs consensus.\par
\vspace{1.5pt}
\noindent So recent pattern vs consensus:\par
\noindent - Q1 FY2026 (Oct 2025): actual 0.77 vs est 0.74 $\rightarrow$ beat\par
\noindent - Q2 FY2026 (Jan 2026): actual 0.51 vs est 0.57 $\rightarrow$ miss (10.6\%)\par
\noindent - Q3 FY2026 (Apr 2026): actual 0.50 vs est 0.48-0.49 $\rightarrow$ slight beat\par
\vspace{1.5pt}
\noindent Hmm. And now Q4 consensus \$0.39-\$0.40, revised down 2.4\% in last 30 days.\par
\vspace{1.5pt}
\noindent 5. Zacks Earnings ESP: The Zacks calendar page says "We expect The Campbell's Company to miss by..." --- need to see the ESP value. Let me visit the Zacks earnings calendar page and the "Q4 Earnings Expected to Decline" article.\par
\vspace{1.5pt}
\noindent Also important: the resolution says "non-GAAP EPS... sourced from the company's official earnings documents, or from SeekingAlpha if not reported by the company." Campbell's reports "Adjusted EPS" in its press release. Note: the company's press release reports "Diluted Earnings Per Share - As Reported (GAAP)" and "Adjusted". The adjusted figure is the non-GAAP one. So resolution will use the adjusted EPS.\par
\vspace{1.5pt}
\noindent Now the critical question: will adjusted EPS be \(> \$0.39\)?\par
\vspace{1.5pt}
\noindent Considerations:\par
\noindent - Zacks consensus: \$0.40; Yahoo: 40 cents; MarketWatch average: \$0.39 (high 0.40, low 0.37); EarningsWhispers: \$0.39.\par
\noindent - Zacks expects a miss ("We expect CPB to miss"). Zacks article title: "Q4 Earnings Expected to Decline" and "doesn't possess the right combination of the two key ingredients for a likely earnings beat."\par
\noindent - Estimate revision downward.\par
\noindent - However, "beat rate" for CPB recently mixed: Q1 beat, Q2 miss, Q3 slight beat.\par
\vspace{1.5pt}
\noindent Also of note: the whisper number \$0.39 equals the low end; even the whisper is not above \$0.40.\par
\vspace{1.5pt}
\noindent Let me visit the Zacks earnings calendar page for the ESP and the Zacks preview article for details. Also, check Seeking Alpha's consensus since it's the fallback resolution source. And check if there were any recent (August 2026) news about tariff impacts or analyst previews with specific numbers.\par
\vspace{1.5pt}
\noindent Let me also consider the FY2026 guidance math: FY2026 adjusted EPS guidance \$2.15-\$2.25 (midpoint \$2.20). Actual so far: Q1 0.77 + Q2 0.51 + Q3 0.50 = 1.78. Remaining for Q4: 2.20 - 1.78 = 0.42 at midpoint; range: 2.15-1.78 = 0.37 (low) to 2.25-1.78 = 0.47 (high). So guidance implies Q4 between \$0.37 and \$0.47, midpoint \$0.42. Consensus \$0.39-\$0.40 is slightly below the guidance midpoint --- analysts may be expecting the company to land below midpoint (consistent with the downward revisions and cost pressures).\par
\vspace{1.5pt}
\noindent Hmm, so there's tension: company guidance midpoint implies \$0.42, consensus is \$0.39-\$0.40. Companies sometimes guide conservatively. But recent quarters: Q2 actual 0.51 vs guidance midpoint... At Q2's report (March 2026), they gave FY guidance \$2.15-\$2.25. After Q1 (0.77 actual), the implied Q2-Q4 was 1.43. Q2 came at 0.51, leaving Q3+Q4 = 0.92, average 0.46. Q3 came at 0.50, leaving Q4 = 0.42. So implied Q4 from guidance = \$0.42. Consensus at \$0.39-0.40 is \textasciitilde{}5\% below implied.\par
\vspace{1.5pt}
\noindent But note: guidance is often conservative (sandbagged). Q1 actual 0.77 vs consensus 0.74 --- small beat. Q3 actual 0.50 vs consensus 0.48-0.49 --- small beat. Q2 actual 0.51 vs consensus 0.57 --- big miss. Hmm.\par
\vspace{1.5pt}
\noindent Also from the March 2026 call (Q2): "adjusted EPS by approximately \$0.04 per share in the..." --- something about guidance. And the StockStory item: "Adjusted EPS guidance to \$2.20 at the midpoint, an 11.1\% decrease" --- that was the FY2026 guidance given at the Q2 call (March 2026).\par
\vspace{1.5pt}
\noindent OK let me now check:\par
\noindent 1. Zacks preview article details (Aug 27, 2026) --- visit.\par
\noindent 2. Zacks earnings calendar page --- ESP value.\par
\noindent 3. Seeking Alpha consensus for Q4 FY2026.\par
\vspace{1.5pt}
\noindent Then I can finalize. Let me also quickly check for any September 2026 pre-announcements or news (e.g., tariff news, retail readiness).\par
\end{msgbox}
\begin{msgbox}{caseblue}
\casetag{Tool call}
\noindent\texttt{visit}
\begin{Verbatim}[   fontsize=\scriptsize,   breaklines=true,   breakanywhere=true,   breaksymbolleft={},   xleftmargin=0pt ]
{"url": ["https://www.zacks.com/stock/research/CPB/earnings-calendar", "https://www.zacks.com/stock/news/2981033/earnings-preview-campbells-cpb-q4-earnings-expected-to-decline"], "goal": "Find Zacks Earnings ESP for CPB Q4 FY2026, the exact consensus estimate, and whether Zacks expects a beat or miss on Sept 3, 2026 report"}
\end{Verbatim}
\end{msgbox}
\begin{msgbox}{caseblue}
\casetag{Tool call}
\noindent\texttt{search}
\begin{Verbatim}[   fontsize=\scriptsize,   breaklines=true,   breakanywhere=true,   breaksymbolleft={},   xleftmargin=0pt ]
{"query": ["Seeking Alpha CPB Q4 fiscal 2026 consensus EPS estimate", "Campbell's CPB news August 2026 tariff outlook analyst preview", "Campbell's Q4 fiscal 2026 adjusted EPS guidance midpoint $0.42"]}
\end{Verbatim}
\end{msgbox}
\begin{msgbox}{casegray}
\casetag{Tool result}
\noindent The useful information in https://www.zacks.com/stock/research/CPB/earnings-calendar for user goal Find Zacks Earnings ESP for CPB Q4 FY2026, the exact consensus estimate, and whether Zacks expects a beat or miss on Sept 3, 2026 report as follows:\par
\vspace{1.5pt}
\noindent Evidence in page:\par
\noindent From the 'Key Expected Earnings Data' table:\par
\vspace{1.5pt}
\noindent | Report Date | Period Ending | Zacks Consensus Estimate | Earnings ESP |\par
\noindent | --- | --- | --- | --- |\par
\noindent | 09/03/2026 More Info This is a confirmed date of earnings release. ... Time: BMO | 7/2026 | \$0.40 | -4.22\% |\par
\vspace{1.5pt}
\noindent From the 'Earnings Summary' paragraph:\par
\noindent "For the next earning release, we expect the company to report earnings of \$0.40 per share, reflecting a year-over-year decrease of 35.48\%."\par
\vspace{1.5pt}
\noindent From the 'CPB FAQs' section:\par
\noindent "Is The Campbell's Company (CPB) expected to beat earnings?+\par
\noindent The Zacks Consensus Estimate for The Campbell's Company (CPB) for the quarter ending in July 2026 is \$0.40 a share. We expect The Campbell's Company to miss by -4.22\%."\par
\vspace{1.5pt}
\noindent (Note: The FAQ indicates the quarter ending in July 2026, which aligns with CPB's fiscal Q4 2026, reported on September 3, 2026.)\par
\vspace{1.5pt}
\noindent Summary:\par
\noindent For CPB's fiscal Q4 2026 (quarter ending July 2026), the earnings release is scheduled for September 3, 2026, before the market opens (BMO). The Zacks Consensus Estimate stands at \$0.40 per share. The Earnings ESP is -4.22\%, indicating a negative earnings surprise prediction. Zacks explicitly states that it expects the company to miss the consensus estimate by -4.22\%.\par
\vspace{1.5pt}
\noindent =======\par
\noindent The useful information in https://www.zacks.com/stock/news/2981033/earnings-preview-campbells-cpb-q4-earnings-expected-to-decline for user goal Find Zacks Earnings ESP for CPB Q4 FY2026, the exact consensus estimate, and whether Zacks expects a beat or miss on Sept 3, 2026 report as follows:\par
\vspace{1.5pt}
\noindent Evidence in page:\par
\noindent From the article:\par
\vspace{1.5pt}
\noindent - Zacks Consensus Estimate: "This maker of canned soup, Pepperidge Farm cookies and V8 juice is expected to post quarterly earnings of \$0.40 per share in its upcoming report, which represents a year-over-year change of -35.5\%. Revenues are expected to be \$2.15 billion, down 7.3\% from the year-ago quarter."\par
\vspace{1.5pt}
\noindent - Earnings ESP: "For Campbell, the Most Accurate Estimate is lower than the Zacks Consensus Estimate, suggesting that analysts have recently become bearish on the company's earnings prospects. This has resulted in an Earnings ESP of -4.22\%."\par
\vspace{1.5pt}
\noindent - Beat/Miss Expectation: "So, this combination makes it difficult to conclusively predict that Campbell will beat the consensus EPS estimate." and "Campbell doesn't appear a compelling earnings-beat candidate." Additionally, the article explains that a negative ESP and a Zacks Rank \#4 (Sell) do not predict a beat, and while a negative ESP is not indicative of a miss, the overall assessment is not bullish. The article also notes: "Our research shows that it is difficult to predict an earnings beat with any degree of confidence for stocks with negative Earnings ESP readings and/or Zacks Rank of 4 (Sell) or 5 (Strong Sell)."\par
\vspace{1.5pt}
\noindent Summary:\par
\noindent For Campbell's Q4 FY2026 (ending July 2026), the Zacks Consensus Estimate is earnings of \$0.40 per share on revenues of \$2.15 billion, both representing year-over-year declines. The Zacks Earnings ESP is -4.22\%, indicating analysts have recently turned bearish. Combined with a Zacks Rank \#4 (Sell), the model does not predict an earnings beat; Zacks states Campbell 'doesn't appear a compelling earnings-beat candidate' and it is 'difficult to conclusively predict that Campbell will beat the consensus EPS estimate.' Thus, no beat is expected for the September 3, 2026 report.\par
\end{msgbox}
\begin{msgbox}{casegray}
\casetag{Tool result}
\noindent A Google search for 'Seeking Alpha CPB Q4 fiscal 2026 consensus EPS estimate' found 9 results:\par
\vspace{1.5pt}
\noindent \#\# Web Results\par
\noindent 1. [The Campbell's Company (CPB) Stock Price, Quote, News ...](https://seekingalpha.com/symbol/CPB)\par
\vspace{1.5pt}
\noindent CPB Earnings Per Share. Data ranges from 0.38719 to 0.77. The Campbell's Company to Report Fourth Quarter and Full Year Fiscal 2026 Results on September\par
\vspace{1.5pt}
\noindent 2. [Campbell Soup Company Stock Price](https://www.perplexity.ai/finance/CPB?comparing=CPB,HFFG,SFM,CHEF,USFD,PFGC)\par
\noindent Date published: 7 days ago\par
\vspace{1.5pt}
\noindent fiscal Q4 earnings release scheduled for September 3, 2026, where consensus estimates call for EPS of \$0.40 (down 35.5\% year-over-year) and ...\par
\vspace{1.5pt}
\noindent 3. [CPB The Campbell's Company Earnings Estimates](https://seekingalpha.com/symbol/CPB/earnings/estimates)\par
\noindent Date published: Aug 18, 2026\par
\vspace{1.5pt}
\noindent Annual Estimates Summary ; Fiscal Period Ending: Jul 2026 ; Consensus EPS Estimates: 2.17 ; Consensus Revenue Estimates: 9.75B ...\par
\vspace{1.5pt}
\noindent 4. [Campbell's (CPB) Q4 Earnings Expected to Decline](https://finance.yahoo.com/markets/stocks/articles/earnings-preview-campbells-cpb-q4-140002239.html)\par
\noindent Date published: 6 days ago\par
\vspace{1.5pt}
\noindent The consensus EPS estimate for the quarter has been revised 2.37\% lower over the last 30 days to the current level.\par
\vspace{1.5pt}
\noindent 5. [Campbell's: Times Are Tough, But Shares Are Cheap ...](https://seekingalpha.com/article/4939257-campbells-times-are-tough-but-shares-are-cheap-enough-to-ride-the-pain-out)\par
\noindent Date published: Aug 21, 2026\par
\vspace{1.5pt}
\noindent Management expects organic revenue to drop 1--2\% and adjusted EPS between \$2.15 and \$2.25 for the year, with further weakness likely in the ...\par
\vspace{1.5pt}
\noindent 6. [The Campbell's Company (CPB) Earnings Dates \& Report](https://seekingalpha.com/symbol/CPB/earnings)\par
\noindent Date published: Aug 21, 2026\par
\vspace{1.5pt}
\noindent CPB Latest Quarter's Earnings ; Announce Date: 6/8/2026 ; EPS Normalized Actual: \$0.50 (Beat by \$0.02) ; EPS GAAP Actual: \$0.41 (Miss by -\$0.06).\par
\vspace{1.5pt}
\noindent 7. [Exploring Analyst Estimates for Campbell (CPB) Q4 ...](https://www.zacks.com/stock/news/2982391/exploring-analyst-estimates-for-campbell-cpb-q4-earnings-beyond-revenue-and-eps)\par
\noindent Date published: 2 days ago\par
\vspace{1.5pt}
\noindent Revenues are expected to be \$2.15 billion, down 7.3\% from the year-ago quarter. The consensus EPS estimate for the quarter has been revised 2.4\% ...\par
\vspace{1.5pt}
\noindent 8. [Seeking Alpha | Stock Market Analysis \& Tools for Investors](https://seekingalpha.com/)\par
\vspace{1.5pt}
\noindent Join Seeking Alpha, the largest investing community in the world. Get stock market news and analysis, investing ideas, earnings calls, charts and portfolio ...\par
\vspace{1.5pt}
\noindent 9. [Campbell's (Nasdaq:CPB) - Stock Analysis](https://simplywall.st/stocks/us/food-beverage-tobacco/nasdaq-cpb/campbells)\par
\vspace{1.5pt}
\noindent Third quarter 2026 results: EPS: US\$0.42 (up from US\$0.22 in 3Q 2025). Revenue: US\$2.37b (down 4.4\% from 3Q 2025). Net income: US\$124.0m (up 88\% from 3Q 2025).\par
\noindent =======\par
\noindent A Google search for 'Campbell's CPB news August 2026 tariff outlook analyst preview' found 8 results:\par
\vspace{1.5pt}
\noindent \#\# Web Results\par
\noindent 1. [Campbell's Reports Third Quarter Fiscal 2026 Results](https://www.thecampbellscompany.com/newsroom/news/campbells-reports-third-quarter-fiscal-2026-results/)\par
\noindent Date published: Jun 8, 2026\par
\vspace{1.5pt}
\noindent Operating earnings decreased 16\% primarily due to lower gross profit as a result of the gross impact of tariffs, cost inflation and other ...\par
\vspace{1.5pt}
\noindent 2. [CPB - Campbell Soup Company Stock Price](https://www.perplexity.ai/finance/CPB?comparing=CPB,MDLZ,POST,FLO,BRF,VITL)\par
\vspace{1.5pt}
\noindent analysts expect a steep 35.5\% year-over-year EPS decline to \$0.40 and a 7.3\% revenue drop to \$2.15 billion. average price target of \$20.40. Campbell's Co(CPB. ...\par
\vspace{1.5pt}
\noindent 3. [CPB Stock Forecast 2026-2027 - The Campbell's Co. Price ...](https://tickernerd.com/stock/cpb-forecast/)\par
\noindent Date published: 5 days ago\par
\vspace{1.5pt}
\noindent The Campbell\&\#39;s Co. (CPB) has a median analyst price target of \$21 with a forecast range of \$14-\$56 from 20 Wall St analysts. 28 Aug 2026 \$ ...\par
\vspace{1.5pt}
\noindent 4. [Analysts Slash Price Targets on The Campbell's Company ...](https://finance.yahoo.com/news/analysts-slash-price-targets-campbell-035850343.html)\par
\noindent Date published: Dec 12, 2025\par
\vspace{1.5pt}
\noindent Wall Street analysts have a consensus Hold rating for Campbell, with a one-year average share price target of \$32.44, representing an upside of ...\par
\vspace{1.5pt}
\noindent 5. [Campbell's (NasdaqGS:CPB) Stock Forecast \& Analyst ...](https://simplywall.st/stocks/us/food-beverage-tobacco/nasdaq-cpb/campbells/future)\par
\vspace{1.5pt}
\noindent Price target decreased by 9.2\% to US\$28.31 Down from US\$31.17, the current price target is an average from 16 analysts. New target price is 23\% above last ...\par
\vspace{1.5pt}
\noindent 6. [Campbell's sticks to annual targets as US consumers ...](https://www.reuters.com/business/retail-consumer/packaged-food-firm-campbells-reaffirms-annual-forecast-weak-consumer-spending-2026-06-08/)\par
\noindent Date published: Jun 8, 2026\par
\vspace{1.5pt}
\noindent Campbell's expects organic sales to fall 1\% to 2\% and adjusted earnings per share at \$2.15 to \$2.25, through tariff refunds of 3-4 cents in the ...\par
\vspace{1.5pt}
\noindent 7. [Campbell's Expects Dropoff in 2026 Earnings as Tariffs Hit ...](https://www.wsj.com/business/earnings/campbells-expects-dropoff-in-2026-earnings-as-tariffs-hit-soup-cans-1ab79500)\par
\noindent Date published: Sep 3, 2025\par
\vspace{1.5pt}
\noindent Campbell's anticipates a sharp decline in 2026 earnings due to tariffs on steel used for soup cans. President Trump's 50\% tariff on steel may ...\par
\vspace{1.5pt}
\noindent 8. [Campbell Soup (CPB) Stock Forecast: Analyst Ratings ...](https://public.com/stocks/cpb/forecast-price-target)\par
\vspace{1.5pt}
\noindent 15 analysts have given Campbell Soup (CPB) a consensus rating of Sell while the Campbell Soup (CPB) price prediction in 2026 is \$20.13.\par
\noindent =======\par
\noindent A Google search for 'Campbell's Q4 fiscal 2026 adjusted EPS guidance midpoint \$0.42' found 9 results:\par
\vspace{1.5pt}
\noindent \#\# Web Results\par
\noindent 1. [Campbell's Reports Fourth Quarter Fiscal 2025 Results](https://investor.thecampbellscompany.com/news-releases/news-release-details/campbells-reports-fourth-quarter-fiscal-2025-results-provides/)\par
\noindent Date published: Sep 3, 2025\par
\vspace{1.5pt}
\noindent EPS increased to \$0.48 per share compared to a loss of \$0.01 per share. Adjusted EPS decreased 2\% to \$0.62 per share reflecting lower adjusted ...\par
\vspace{1.5pt}
\noindent 2. [Campbell's Reports Third Quarter Fiscal 2026 Results](https://www.businesswire.com/news/home/20260605204577/en/Campbells-Reports-Third-Quarter-Fiscal-2026-Results)\par
\noindent Date published: Jun 8, 2026\par
\vspace{1.5pt}
\noindent Earnings Per Share (EPS) increased to \$0.41; Adjusted EPS decreased 32\% to \$0.50. Fiscal year-to-date cash flow from operations was \$839 million ...\par
\vspace{1.5pt}
\noindent 3. [Campbell Soup shares fall on earnings miss, downwardly ...](https://www.proactiveinvestors.com/companies/news/1088732/campbell-soup-shares-fall-on-earnings-miss-downwardly-revised-full-year-guidance-1088732.html)\par
\noindent Date published: Mar 11, 2026\par
\vspace{1.5pt}
\noindent The company now expects adjusted earnings per share of between \$2.15 and \$2.25 for the fiscal year, compared with its previous guidance range of ...\par
\vspace{1.5pt}
\noindent 4. [Campbell's shares rise as fourth quarter earnings beat ...](https://finance.yahoo.com/news/campbell-shares-rise-fourth-quarter-115838407.html)\par
\noindent Date published: Sep 3, 2025\par
\vspace{1.5pt}
\noindent Looking ahead, Campbell's provided fiscal 2026 guidance below analyst expectations, projecting adjusted EPS of \$2.40 to \$2.55, compared to ...\par
\vspace{1.5pt}
\noindent 5. [Campbell's beats estimates but cost pressures weigh](https://www.investing.com/news/stock-market-news/campbells-beats-estimates-but-cost-pressures-weigh-4730469)\par
\noindent Date published: Jun 8, 2026\par
\vspace{1.5pt}
\noindent The company maintained its adjusted EPS guidance range of \$2.15 to \$2.25, representing a decline of 26\% to 23\% compared to fiscal 2025 results ...\par
\vspace{1.5pt}
\noindent 6. [Campbell's Q3 Earnings: CPB Revenue Misses as EPS ...](https://www.chartmill.com/news/CPB/Chartmill-49565-Campbells-Q3-Earnings-CPB-Revenue-Misses-as-EPS-Slightly-Beats-Amidst-Cost-Pressure)\par
\noindent Date published: Jun 8, 2026\par
\vspace{1.5pt}
\noindent Earnings Per Share (EPS): On an adjusted basis, Campbell's reported EPS of \$0.50. This figure came in slightly above the analyst estimate of ...\par
\vspace{1.5pt}
\noindent 7. [Campbell's Q3 2026 earnings: sales fall 4 ... - Quartz](https://qz.com/campbells-earnings-sales-decline-annual-forecast-060826)\par
\noindent Date published: Jun 8, 2026\par
\vspace{1.5pt}
\noindent Despite the top-line pressure, Campbell's reaffirmed its full-year fiscal 2026 guidance, projecting organic net sales to fall between 1\% and 2\% ...\par
\vspace{1.5pt}
\noindent 8. [Campbell's Q3 2026 Deep Dive: EPS Beats by 4.2\% ...](https://news.alphastreet.com/campbells-q3-2026-deep-dive-eps-beats-by-4-2-revenue-down-4/)\par
\noindent Date published: Jun 8, 2026\par
\vspace{1.5pt}
\noindent The packaged foods giant beat adjusted EPS estimates by 4.2\% with \$0.50 per share versus the \$0.48 consensus, yet missed revenue expectations by ...\par
\vspace{1.5pt}
\noindent 9. [H\&R Block Reports Fiscal 2026 Third Quarter Results](https://investors.hrblock.com/news-releases/news-release-details/hr-block-reports-fiscal-2026-third-quarter-results)\par
\vspace{1.5pt}
\noindent Adjusted net income from continuing operations2 increased by \$42.4 million, or 5.8\%, to \$773.7 million, and adjusted earnings per share from continuing ...\par
\end{msgbox}
\begin{msgbox}{casegreen}
\casetag{Thinking}
\noindent Now I have a comprehensive picture. Let me record the key facts:\par
\vspace{1.5pt}
\noindent 1. **Report date**: September 3, 2026 (confirmed, BMO), Q4 FY2026, quarter ended \textasciitilde{}July 31/Aug 2, 2026.\par
\vspace{1.5pt}
\noindent 2. **Consensus estimates**:\par
\noindent    - Zacks: \$0.40\par
\noindent    - Yahoo/Zacks preview: \$0.40 ("consensus mark unchanged over past 30 days at 40 cents")\par
\noindent    - MarketWatch average: \$0.39 (high \$0.40, low \$0.37); trend: 3 months ago \$0.41 $\rightarrow$ 1 month ago \$0.39 $\rightarrow$ current \$0.39\par
\noindent    - EarningsWhispers: \$0.39\par
\noindent    - Seeking Alpha: the symbol page shows "Data ranges from 0.38719 to 0.77" --- that's the EPS history range maybe. The SA earnings page might have consensus. Not critical.\par
\vspace{1.5pt}
\noindent 3. **Earnings ESP**: -4.22\% (Zacks). Most Accurate Estimate lower than consensus $\rightarrow$ analysts bearish. Zacks: "difficult to predict a beat"; "doesn't appear a compelling earnings-beat candidate." Zacks expects miss by -4.22\% $\rightarrow$ i.e., \textasciitilde{}\$0.38.\par
\vspace{1.5pt}
\noindent 4. **Recent beat/miss pattern vs consensus**:\par
\noindent    - Q1 FY2026: actual 0.77 vs est 0.74 $\rightarrow$ beat\par
\noindent    - Q2 FY2026: actual 0.51 vs est 0.57 $\rightarrow$ miss (10.6\%)\par
\noindent    - Q3 FY2026: actual 0.50 vs est 0.48/0.49 $\rightarrow$ beat (4.2\%)\par
\vspace{1.5pt}
\noindent 5. **Guidance math**: FY2026 adjusted EPS \$2.15-\$2.25. Actuals Q1+Q2+Q3 = 0.77+0.51+0.50 = 1.78. Implied Q4: \$0.37-\$0.47, midpoint \$0.42. Consensus \$0.39-0.40 sits in the lower half.\par
\vspace{1.5pt}
\noindent 6. **Context**: Tariff pressures (steel tariffs on soup cans), cost inflation, weak consumer spending, organic sales expected to drop 1-2\%, Q4 revenue -7.3\% YoY to \$2.15B. Adjusted EPS expected to drop 35.5\% YoY from \$0.62.\par
\vspace{1.5pt}
\noindent 7. **Seeking Alpha article (Aug 21, 2026)**: "Management expects organic revenue to drop 1--2\% and adjusted EPS between \$2.15 and \$2.25 for the year, with further weakness likely in the..." --- suggests weakness concentrated in H2.\par
\vspace{1.5pt}
\noindent Now the key question: probability that adjusted EPS \(> \$0.39\).\par
\vspace{1.5pt}
\noindent Analysis:\par
\noindent - The threshold \$0.39 is right at the consensus (\$0.39 MarketWatch avg; \$0.40 Zacks/Yahoo). Rounding: figures rounded to nearest cent.\par
\noindent - Zacks Earnings ESP of -4.22\% suggests the Most Accurate Estimate is \textasciitilde{}\$0.383 (\$0.40 $\times$ 0.9578 $\approx$ \$0.383). That's below \$0.39.\par
\noindent - Negative ESP + Zacks Rank 4 (Sell) $\rightarrow$ historically low probability of a beat. Zacks research: negative ESP stocks beat only \textasciitilde{}30\% of the time (typically \textasciitilde{}25-35\%).\par
\noindent - Consensus revision trend downward (2.4\% in 30 days).\par
\noindent - Whisper number \$0.39 = threshold, not above.\par
\vspace{1.5pt}
\noindent Countervailing:\par
\noindent - Company guidance midpoint implies \$0.42 for Q4.\par
\noindent - Two of last three quarters beat consensus.\par
\noindent - Consensus estimate range high is \$0.40.\par
\vspace{1.5pt}
\noindent Hmm. What's the actual probability of \(> \$0.39\)?\par
\vspace{1.5pt}
\noindent Let me think about it as: P(actual $\geq$ \$0.40). Given consensus \textasciitilde{}\$0.39-0.40, typical earnings surprise distribution: a company beats consensus \textasciitilde{}60-70\% of the time in general, but with negative ESP and bearish revisions, the probability of beating is lower, maybe \textasciitilde{}30-40\%. And even if it beats, it needs to beat by enough to reach $\geq$\$0.40. If consensus is \$0.39 (MarketWatch avg) and the actual surprise distribution is symmetric with std \textasciitilde{}\$0.03-0.05, then P($\geq$\$0.40 | mean \$0.39) $\approx$ 40-45\%. But Zacks says the "Most Accurate Estimate" is 4.22\% below \$0.40 $\rightarrow$ \textasciitilde{}\$0.383, suggesting the most recent/accurate analyst estimates point to \$0.38. With mean \textasciitilde{}\$0.383-0.39 and std \textasciitilde{}\$0.03, P($\geq$\$0.40) $\approx$ 25-35\%.\par
\vspace{1.5pt}
\noindent Also consider: the quarter ended \textasciitilde{}Aug 2, 2026 --- tariff refunds of 3-4 cents mentioned in guidance ("through tariff refunds of 3-4 cents")... Actually the guidance assumes tariff refunds. Hmm.\par
\vspace{1.5pt}
\noindent Also seasonality: Q4 is Campbell's seasonal peak (holiday). But 2025 Q4 was \$0.62 vs 2025 Q3 \$0.73 --- Q4 was weaker than Q3 last year, reflecting tariff/divestiture impacts.\par
\vspace{1.5pt}
\noindent Another data point: Simply Wall St says "Second quarter 2026 earnings: EPS and revenues miss analyst expectations. EPS: US\$0.49 (down from US\$0.58 in 2Q...)" --- they show Q2 EPS \$0.49 vs \$0.58 prior year. Slight inconsistency with the company's \$0.51 adjusted. Whatever --- the picture is consistent: misses and declines.\par
\vspace{1.5pt}
\noindent Also worth noting: The Zacks article from Jul 8, 2026: "Adjusted gross profit declined 12\%... adjusted gross margin contracted 240 bps to 27.7\%" --- cost pressure ongoing.\par
\vspace{1.5pt}
\noindent One more consideration: The resolution says if no non-GAAP EPS within 96 hours $\rightarrow$ NO. Campbell's reliably reports adjusted EPS in its press release, so that's near-certain.\par
\vspace{1.5pt}
\noindent Also "if earnings are not released within 45 calendar days of the estimated date" $\rightarrow$ the date is confirmed for Sept 3, so that's fine.\par
\vspace{1.5pt}
\noindent So the real question is just: adjusted EPS \(> \$0.39\) or not.\par
\vspace{1.5pt}
\noindent Weighing everything:\par
\noindent - Consensus sits exactly at the threshold (\$0.39-\$0.40).\par
\noindent - Bearish signals: negative ESP (-4.22\%), downward revisions, Zacks Rank 4, whisper at \$0.39, expected 35.5\% YoY decline, cost pressures, revenue decline.\par
\noindent - Bullish signals: company guidance midpoint implies \$0.42, recent beat pattern (2 of 3), high-end estimate \$0.40.\par
\vspace{1.5pt}
\noindent The ESP of -4.22\% is quite informative: it means the most recent estimates (which weigh heavier) cluster below consensus. "Most Accurate Estimate" lower than consensus $\rightarrow$ actual likely below \$0.40. P(\(> \$0.39\)) --- I'd estimate around 0.30-0.38.\par
\vspace{1.5pt}
\noindent Let me think about base rates more carefully. For stocks with negative Earnings ESP and Zacks Rank 4/5: Zacks research historically shows these stocks beat consensus only \textasciitilde{}20-30\% of the time. And here, even an in-line result (\$0.39) resolves NO (needs \(> \$0.39\)). A result of exactly \$0.40 resolves YES.\par
\vspace{1.5pt}
\noindent Actually, let me reconsider the consensus more carefully. The resolution source matters: "sourced from the company's official earnings documents, or from SeekingAlpha if not reported by the company." For threshold comparison, what matters is the actual reported number vs \$0.39.\par
\vspace{1.5pt}
\noindent The estimate distribution: MarketWatch shows high \$0.40, low \$0.37, average \$0.39. So analyst estimates range \$0.37-\$0.40 with average \$0.39. The most likely outcomes center around \$0.38-\$0.40.\par
\vspace{1.5pt}
\noindent Given:\par
\noindent - Zacks ESP -4.22\% $\rightarrow$ Most Accurate Estimate $\approx$ \$0.383\par
\noindent - Negative ESP + Rank 4 $\rightarrow$ beat probability \textasciitilde{}25-35\%\par
\noindent - But guidance midpoint implies \$0.42 (company may have guided conservatively; Q1 and Q3 both beat small)\par
\vspace{1.5pt}
\noindent I'll put P(YES) around 0.30-0.35. Let me say 0.32.\par
\vspace{1.5pt}
\noindent Hmm, wait. Let me also consider whether the whisper number \$0.39 (from earningswhispers.com, generic page) is specific to this report. The snippet said "The Earnings Whisper number was \$0.39 per share. The company beat expectations by 23.08\%..." --- that sounds like a generic/possibly stale page (23.08\% beat, revenue grew 33.04\% --- doesn't match CPB's current situation, revenue declining). So that \$0.39 whisper is probably from an old report (maybe Q3 FY2025 or something). Not reliable. Discard.\par
\vspace{1.5pt}
\noindent Let me do one more check on the Seeking Alpha earnings page for the current consensus, since SA is the fallback resolution source. Actually the SA search result showed "Consensus EPS Estimates: 2.17" for annual (fiscal 2026) and revenue 9.75B. The quarterly consensus on SA --- the page "CPB The Campbell's Company Earnings Estimates" dated Aug 18, 2026 might have quarterly consensus. Let me visit it quickly. Also, maybe check stockstory or alphastreet for a Q4 preview with consensus.\par
\vspace{1.5pt}
\noindent Actually, I think I have enough. But one quick check on SA's quarterly consensus would be useful since it's the resolution fallback. Let me visit the SA earnings estimates page.\par
\vspace{1.5pt}
\noindent Also, one consideration on timing: "estimated September 3, 2026" --- confirmed by the company on Aug 13, 2026. Report BMO Sept 3. Fine.\par
\vspace{1.5pt}
\noindent Let me visit the SA estimates page and then finalize.\par
\end{msgbox}
\begin{msgbox}{caseblue}
\casetag{Fold}
\noindent\texttt{notebook}
\begin{Verbatim}[   fontsize=\scriptsize,   breaklines=true,   breakanywhere=true,   breaksymbolleft={},   xleftmargin=0pt ]
{
  "foresight": {
    "conclusions": [
      {
        "conclusion": "Campbell's Q4 FY2026 report is confirmed for Sept 3, 2026 (quarter ended ~Aug 2, 2026), so timing risk is minimal",
        "influence": "Neutral --- removes resolution-timing risk"
      },
      {
        "conclusion": "Consensus adjusted EPS sits exactly at the $0.39 threshold: Zacks/Yahoo $0.40, MarketWatch average $0.39 (high $0.40, low $0.37), EarningsWhispers $0.39",
        "influence": "High --- threshold is at consensus, making this near a coin flip centered on $0.39-$0.40"
      },
      {
        "conclusion": "Zacks Earnings ESP is -4.22% (Most Accurate Estimate ~$0.383, below threshold) with Zacks Rank 4 (Sell); Zacks explicitly says CPB is not a compelling beat candidate and expects a ~4.2% miss",
        "influence": "High --- bearish signal lowers P(actual > $0.39)"
      },
      {
        "conclusion": "Consensus was revised DOWN 2.4% over the last 30 days (from $0.41 to $0.39), indicating analysts turning bearish",
        "influence": "Medium-high --- downward revision trend favors miss"
      },
      {
        "conclusion": "Recent beat/miss pattern vs consensus: Q1 FY26 beat (0.77 vs 0.74), Q2 FY26 miss (0.51 vs 0.57, -10.6%), Q3 FY26 slight beat (0.50 vs 0.48)",
        "influence": "Medium --- mixed pattern, Q2 miss shows vulnerability"
      },
      {
        "conclusion": "FY2026 guidance of adjusted EPS $2.15-$2.25 implies Q4 of ~$0.37-$0.47 (midpoint ~$0.42) given actuals Q1-Q3 = $1.78",
        "influence": "Medium --- company guidance midpoint above threshold provides some offset"
      },
      {
        "conclusion": "Fundamental headwinds: expected -35.5% YoY EPS decline, -7.3% revenue decline, tariff/cost pressures, weak consumer, organic sales -1-2%",
        "influence": "Medium --- supports below-threshold outcome"
      }
    ],
    "probabilities": [
      {
        "option": "YES (adjusted EPS > $0.39, i.e., >= $0.40)",
        "probability": 0.33
      },
      {
        "option": "NO (adjusted EPS <= $0.39)",
        "probability": 0.67
      }
    ],
    "rationale": "Consensus sits exactly at $0.39-$0.40, so the outcome hinges on a small surprise. The -4.22% Earnings ESP (most accurate estimate ~$0.383), downward 30-day revisions, Zacks Rank 4, whisper at $0.39, and heavy YoY decline expectations all point to a result at or below $0.39. Offsetting: company guidance midpoint implies ~$0.42 and CPB beat consensus in 2 of the last 3 quarters. On balance, P(>$0.39) ~ 0.33."
  },
  "search_history": "Segment 1: Established report timing (Sept 3, 2026 confirmed), consensus estimates ($0.39-$0.40, revised down 2.4% in 30 days), Q3 FY26 results (adj EPS $0.50 vs $0.48-0.49 est), Q2 FY26 (adj EPS $0.51 vs $0.57 est --- miss), Q1 FY26 (0.77 vs 0.74 --- beat), FY26 guidance $2.15-$2.25 implies Q4 ~$0.42 midpoint, Zacks ESP -4.22% with Rank 4, Zacks explicitly not expecting a beat, revenue -7.3% YoY expected, tariff/cost pressures. EarningsWhispers generic page showed $0.39 whisper but appears stale/unreliable.",
  "todo": [
    "Check Seeking Alpha's quarterly consensus for Q4 FY2026 (resolution fallback source)",
    "Check for any August 2026 pre-announcements or late analyst actions on CPB"
  ]
}
\end{Verbatim}
\end{msgbox}

\vspace{4pt}\noindent\colorbox{casegreen}{\parbox{\dimexpr\linewidth-2\fboxsep}{\rmfamily\scriptsize\bfseries Segment 2 of 4}}\vspace{1pt}

\begin{msgbox}{casegreen}
\casetag{Rendered belief state (after context reset)}
\noindent Current Notebook:\par
\noindent \# Notebook\par
\vspace{1.5pt}
\noindent \#\# Foresight\par
\noindent \#\#\# Conclusions (with influence)\par
\noindent - Campbell's Q4 FY2026 report is confirmed for Sept 3, 2026 (quarter ended \textasciitilde{}Aug 2, 2026), so timing risk is minimal (influence: Neutral --- removes resolution-timing risk)\par
\noindent - Consensus adjusted EPS sits exactly at the \$0.39 threshold: Zacks/Yahoo \$0.40, MarketWatch average \$0.39 (high \$0.40, low \$0.37), EarningsWhispers \$0.39 (influence: High --- threshold is at consensus, making this near a coin flip centered on \$0.39-\$0.40)\par
\noindent - Zacks Earnings ESP is -4.22\% (Most Accurate Estimate \textasciitilde{}\$0.383, below threshold) with Zacks Rank 4 (Sell); Zacks explicitly says CPB is not a compelling beat candidate and expects a \textasciitilde{}4.2\% miss (influence: High --- bearish signal lowers P(actual \(> \$0.39\)))\par
\noindent - Consensus was revised DOWN 2.4\% over the last 30 days (from \$0.41 to \$0.39), indicating analysts turning bearish (influence: Medium-high --- downward revision trend favors miss)\par
\noindent - Recent beat/miss pattern vs consensus: Q1 FY26 beat (0.77 vs 0.74), Q2 FY26 miss (0.51 vs 0.57, -10.6\%), Q3 FY26 slight beat (0.50 vs 0.48) (influence: Medium --- mixed pattern, Q2 miss shows vulnerability)\par
\noindent - FY2026 guidance of adjusted EPS \$2.15-\$2.25 implies Q4 of \textasciitilde{}\$0.37-\$0.47 (midpoint \textasciitilde{}\$0.42) given actuals Q1-Q3 = \$1.78 (influence: Medium --- company guidance midpoint above threshold provides some offset)\par
\noindent - Fundamental headwinds: expected -35.5\% YoY EPS decline, -7.3\% revenue decline, tariff/cost pressures, weak consumer, organic sales -1-2\% (influence: Medium --- supports below-threshold outcome)\par
\noindent \#\#\# Rationale\par
\noindent Consensus sits exactly at \$0.39-\$0.40, so the outcome hinges on a small surprise. The -4.22\% Earnings ESP (most accurate estimate \textasciitilde{}\$0.383), downward 30-day revisions, Zacks Rank 4, whisper at \$0.39, and heavy YoY decline expectations all point to a result at or below \$0.39. Offsetting: company guidance midpoint implies \textasciitilde{}\$0.42 and CPB beat consensus in 2 of the last 3 quarters. On balance, P(\(> \$0.39\)) $\approx$ 0.33.\par
\noindent \#\#\# Current option probabilities\par
\noindent - YES (adjusted EPS \(> \$0.39\), i.e., $\geq$ \$0.40): 0.33\par
\noindent - NO (adjusted EPS $\leq$ \$0.39): 0.67\par
\vspace{1.5pt}
\noindent \#\# Evidence\par
\noindent (none yet)\par
\vspace{1.5pt}
\noindent \#\# Search history\par
\noindent **[Segment 1]** Segment 1: Established report timing (Sept 3, 2026 confirmed), consensus estimates (\$0.39-\$0.40, revised down 2.4\% in 30 days), Q3 FY26 results (adj EPS \$0.50 vs \$0.48-0.49 est), Q2 FY26 (adj EPS \$0.51 vs \$0.57 est --- miss), Q1 FY26 (0.77 vs 0.74 --- beat), FY26 guidance \$2.15-\$2.25 implies Q4 \textasciitilde{}\$0.42 midpoint, Zacks ESP -4.22\% with Rank 4, Zacks explicitly not expecting a beat, revenue -7.3\% YoY expected, tariff/cost pressures. EarningsWhispers generic page showed \$0.39 whisper but appears stale/unreliable.\par
\vspace{1.5pt}
\noindent \#\# To-do (remaining search directions)\par
\noindent - Check Seeking Alpha's quarterly consensus for Q4 FY2026 (resolution fallback source)\par
\noindent - Check for any August 2026 pre-announcements or late analyst actions on CPB\par
\end{msgbox}
\begin{msgbox}{casegreen}
\casetag{Thinking}
\noindent Let me review the current state. The notebook already has substantial information. The forecast is at 0.33 YES / 0.67 NO. The remaining to-do items are:\par
\noindent - Check Seeking Alpha's quarterly consensus for Q4 FY2026 (resolution fallback source)\par
\noindent - Check for any August 2026 pre-announcements or late analyst actions on CPB\par
\vspace{1.5pt}
\noindent Let me do one more segment to check these remaining directions, then finalize. Let me search for recent CPB news in August 2026 and Seeking Alpha consensus.\par
\end{msgbox}
\begin{msgbox}{casegreen}
\casetag{Response}
\noindent I'll pursue the remaining to-do items: Seeking Alpha consensus for Q4 FY2026 and any recent August 2026 news/pre-announcements on Campbell's.\par
\end{msgbox}
\begin{msgbox}{caseblue}
\casetag{Tool call}
\noindent\texttt{search}
\begin{Verbatim}[   fontsize=\scriptsize,   breaklines=true,   breakanywhere=true,   breaksymbolleft={},   xleftmargin=0pt ]
{"query": ["Campbell's CPB Q4 fiscal 2026 earnings consensus adjusted EPS September 3 2026", "Campbell's soup CPB news August 2026 analyst downgrade preannouncement", "Seeking Alpha Campbell's CPB earnings preview September 2026 Q4"]}
\end{Verbatim}
\end{msgbox}
\begin{msgbox}{casegray}
\casetag{Tool result}
\noindent A Google search for 'Campbell's CPB Q4 fiscal 2026 earnings consensus adjusted EPS September 3 2026' found 8 results:\par
\vspace{1.5pt}
\noindent \#\# Web Results\par
\noindent 1. [Campbell's Reports Fourth Quarter Fiscal 2025 Results](https://investor.thecampbellscompany.com/news-releases/news-release-details/campbells-reports-fourth-quarter-fiscal-2025-results-provides/)\par
\noindent Date published: Sep 3, 2025\par
\vspace{1.5pt}
\noindent Earnings Per Share (EPS) increased to \$0.48 . Adjusted EPS decreased 2\% to \$0.62 . For the full year: Net Sales increased 6\% to \$10.3 billion ...\par
\vspace{1.5pt}
\noindent 2. [Campbell's Q4 Earnings on the Horizon](https://finance.yahoo.com/markets/stocks/articles/campbells-q4-earnings-horizon-investors-144700799.html)\par
\noindent Date published: 2 days ago\par
\vspace{1.5pt}
\noindent The Campbell's Company CPB is likely to witness a top and bottom-line decline when it reports fourth-quarter fiscal 2026 earnings on Sept. 3.\par
\vspace{1.5pt}
\noindent 3. [Campbell's Past Earnings Performance](https://simplywall.st/stocks/us/food-beverage-tobacco/nasdaq-cpb/campbells/past)\par
\vspace{1.5pt}
\noindent Third quarter 2026 results: EPS: US\$0.42 (up from US\$0.22 in 3Q 2025). Revenue: US\$2.37b (down 4.4\% from 3Q 2025). Net income: US\$124.0m (up 88\% from 3Q 2025).\par
\vspace{1.5pt}
\noindent 4. [The Campbell's Company to Report Fourth Quarter and ...](https://www.businesswire.com/news/home/20260813498006/en/The-Campbells-Company-to-Report-Fourth-Quarter-and-Full-Year-Fiscal-2026-Results-on-September-3-2026)\par
\noindent Date published: Aug 13, 2026\par
\vspace{1.5pt}
\noindent The Campbell's Company (NASDAQ:CPB) (Campbell's) announced it will report its financial results on September 3, 2026 for the fourth quarter ...\par
\vspace{1.5pt}
\noindent 5. [Campbell's (CPB)](https://www.trefis.com/data/companies/CPB)\par
\vspace{1.5pt}
\noindent Campbell's (CPB) reported adjusted earnings per share (EPS) of \$0.50 for its fiscal third quarter 2026 slightly surpassing analyst consensus estimates of \$0.48 ...\par
\vspace{1.5pt}
\noindent 6. [Exploring Analyst Estimates for Campbell (CPB) Q4 ...](https://www.zacks.com/stock/news/2982391/exploring-analyst-estimates-for-campbell-cpb-q4-earnings-beyond-revenue-and-eps)\par
\noindent Date published: 2 days ago\par
\vspace{1.5pt}
\noindent Revenues are expected to be \$2.15 billion, down 7.3\% from the year-ago quarter. The consensus EPS estimate for the quarter has been revised 2.4\% ...\par
\vspace{1.5pt}
\noindent 7. [Campbell Soup (CPB) Earnings: Latest Report, Earnings Call ...](https://public.com/stocks/cpb/earnings)\par
\vspace{1.5pt}
\noindent EPS, Actual EPS, Surprise. Q4 2026 Sep 3, 2026. \$0.39, Next earnings call Sep. 3, 2026. Q3 2026 Jun 8, 2026. \$0.48, \$0.50, +4.17\%. DeckReportListen. Q2 2026 Mar ...\par
\vspace{1.5pt}
\noindent 8. [The Campbell's Company to Report Fourth Quarter and ...](https://www.morningstar.com/news/business-wire/20260813498006/the-campbells-company-to-report-fourth-quarter-and-full-year-fiscal-2026-results-on-september-3-2026)\par
\noindent Date published: Aug 13, 2026\par
\vspace{1.5pt}
\noindent The Campbell's Company to Report Fourth Quarter and Full Year Fiscal 2026 Results on September 3, 2026. Provided by Business Wire Aug 13, 2026, ...\par
\noindent =======\par
\noindent A Google search for 'Campbell's soup CPB news August 2026 analyst downgrade preannouncement' found 7 results:\par
\vspace{1.5pt}
\noindent \#\# Web Results\par
\noindent 1. [The Campbell's Company (NASDAQ:CPB) Given ...](https://www.marketbeat.com/instant-alerts/the-campbells-company-nasdaqcpb-given-consensus-recommendation-of-reduce-by-analysts-2026-08-13/)\par
\noindent Date published: Aug 13, 2026\par
\vspace{1.5pt}
\noindent Analysts have a bearish-to-neutral outlook: Campbell's has a consensus ``Reduce'' rating from 19 brokerages, including seven sells and 12 holds.\par
\vspace{1.5pt}
\noindent 2. [The Campbell's Company (CPB) latest stock news and ...](https://sg.finance.yahoo.com/quote/CPB/news/)\par
\vspace{1.5pt}
\noindent Get the latest The Campbell's Company (CPB) stock news and headlines to help you in your trading and investment decisions.\par
\vspace{1.5pt}
\noindent 3. [CAMPBELL SOUP (\$CPB) Releases Q3 2026 Earnings](https://www.quiverquant.com/news/CAMPBELL+SOUP+\%28\%24CPB\%29+Releases\\+Q3+2026+Earnings)\par
\noindent Date published: Jun 8, 2026\par
\vspace{1.5pt}
\noindent The company reported earnings of \$0.50 per share, beating estimates of \$0.48 by \$0.02. The company also reported revenue of \$2,366,000,000, ...\par
\vspace{1.5pt}
\noindent 4. [Campbell's (CPB) Stock Trades Down, Here Is Why](https://stockstory.org/us/stocks/nasdaq/cpb/news/why-up-down/campbells-cpb-stock-trades-down-here-is-why)\par
\noindent Date published: Jan 16, 2026\par
\vspace{1.5pt}
\noindent Fitch Ratings downgraded Campbell's long-term debt rating to 'BBB-' from 'BBB'. The credit agency pointed to sustained high leverage and ...\par
\vspace{1.5pt}
\noindent 5. [Campbell Soup (CPB): Evaluating Value After Analysts Cut ...](https://simplywall.st/stocks/us/food-beverage-tobacco/nasdaq-cpb/campbells/news/campbell-soup-cpb-evaluating-value-after-analysts-cut-earnin)\par
\noindent Date published: Aug 31, 2025\par
\vspace{1.5pt}
\noindent According to community narrative, analysts see Campbell's as undervalued and highlight potential upside if management's strategies play out as ...\par
\vspace{1.5pt}
\noindent 6. [CPB Stock Slips Premarket After Earnings -- But Retail Still ...](https://stocktwits.com/news-articles/markets/equity/cpb-stock-slips-premarket-q3-earnings-retail-sees-campbell-soup-defensive-play/cZ0U4SaR7b8)\par
\noindent Date published: Jun 9, 2026\par
\vspace{1.5pt}
\noindent The company reaffirmed its fiscal 2026 outlook. The Campbell's Company (CPB) stock slipped in Tuesday's premarket after weaker sales and ...\par
\vspace{1.5pt}
\noindent 7. [Campbell's Stock Hit by Snacks Slump](https://www.wsj.com/livecoverage/cpi-inflation-data-stock-market-03-11-2026/card/campbell-s-stock-hit-by-snacks-slump-hQjrHJ2va3Crg4rixKNo)\par
\noindent Date published: Mar 11, 2026\par
\vspace{1.5pt}
\noindent The soup-and-snack maker cut its fiscal-year outlook Wednesday as it reported a 6\% decline in snack sales in its more recent quarter. CEO Mick ...\par
\noindent =======\par
\noindent A Google search for 'Seeking Alpha Campbell's CPB earnings preview September 2026 Q4' found 8 results:\par
\vspace{1.5pt}
\noindent \#\# Web Results\par
\noindent 1. [The Campbell's Company (CPB) Stock Price, Quote, News ...](https://seekingalpha.com/symbol/CPB)\par
\vspace{1.5pt}
\noindent A high-level overview of The Campbell's Company (CPB) stock. View (CPB) real-time stock price, chart, news, analysis, analyst reviews and more.\par
\vspace{1.5pt}
\noindent 2. [Campbell's signals Q4 margin improvement and cost ...](https://seekingalpha.com/news/4563303-campbells-signals-q4-margin-improvement-and-cost-controls-as-snacks-margin-stabilizes)\par
\noindent Date published: Mar 11, 2026\par
\vspace{1.5pt}
\noindent Cunfer projected some margin improvement in Q3 but anticipates a stronger recovery in Q4 due to bakery stabilization and increased Goldfish ...\par
\vspace{1.5pt}
\noindent 3. [Campbell's (CPB) Q4 Earnings Expected to Decline](https://finance.yahoo.com/markets/stocks/articles/earnings-preview-campbells-cpb-q4-140002239.html)\par
\noindent Date published: 6 days ago\par
\vspace{1.5pt}
\noindent Revenues are expected to be \$2.15 billion, down 7.3\% from the year-ago quarter. Estimate Revisions Trend. The consensus EPS estimate for the ...\par
\vspace{1.5pt}
\noindent 4. [Investor Relations - The Campbell's Company](https://investor.thecampbellscompany.com/investor-relations/)\par
\vspace{1.5pt}
\noindent Earnings Call Q4 Fiscal. The Campbell's Company to Report Fourth Quarter and Full Year Fiscal 2026 Results on September 3, 2026\par
\vspace{1.5pt}
\noindent 5. [Exploring Analyst Estimates for Campbell (CPB) Q4 ...](https://www.zacks.com/stock/news/2982391/exploring-analyst-estimates-for-campbell-cpb-q4-earnings-beyond-revenue-and-eps)\par
\noindent Date published: 2 days ago\par
\vspace{1.5pt}
\noindent Revenues are expected to be \$2.15 billion, down 7.3\% from the year-ago quarter. The consensus EPS estimate for the quarter has been revised 2.4\% ...\par
\vspace{1.5pt}
\noindent 6. [CPB The Campbell's Company Earnings Call Transcripts](https://seekingalpha.com/symbol/CPB/earnings/transcripts)\par
\vspace{1.5pt}
\noindent All earnings call transcripts on The Campbell\&\#39;s Company (CPB) stock. Q4 margin improvement and cost controls as snacks margin stabilizes\par
\vspace{1.5pt}
\noindent 7. [Campbell's Past Earnings Performance](https://simplywall.st/stocks/us/food-beverage-tobacco/nasdaq-cpb/campbells/past)\par
\vspace{1.5pt}
\noindent Third quarter 2026 results: EPS: US\$0.42 (up from US\$0.22 in 3Q 2025). Revenue: US\$2.37b (down 4.4\% from 3Q 2025). Net income: US\$124.0m (up 88\% from 3Q 2025).\par
\vspace{1.5pt}
\noindent 8. [5 Must-Read Analyst Questions From Campbell's's Q4 ...](https://stockstory.org/us/stocks/nasdaq/cpb/news/earnings-call/5-must-read-analyst-questions-from-campbellss-q4-earnings-call)\par
\noindent Date published: Mar 17, 2026\par
\vspace{1.5pt}
\noindent Campbell's (CPB) Q4 CY2025 Highlights: Revenue: \$2.56 billion vs analyst estimates of \$2.61 billion (4.5\% year-on-year decline, 1.6\% miss) ...\par
\end{msgbox}
\begin{msgbox}{casegreen}
\casetag{Thinking}
\noindent Key findings from this segment:\par
\noindent 1. Public.com shows Q4 2026 consensus EPS of \$0.39 (this is the "expected" consensus).\par
\noindent 2. Zacks article from 2 days ago: consensus EPS estimate revised 2.4\% down --- expected revenue \$2.15B, down 7.3\%.\par
\noindent 3. Yahoo earnings preview: "Campbell's Q4 Earnings Expected to Decline" --- top and bottom-line decline expected.\par
\noindent 4. MarketBeat Aug 13, 2026: consensus "Reduce" rating from 19 brokerages, bearish-to-neutral outlook.\par
\noindent 5. No pre-announcement found.\par
\vspace{1.5pt}
\noindent The Public.com \$0.39 consensus aligns with prior findings. Let me visit the Zacks article and maybe the Yahoo preview to get more details on the most accurate estimate / ESP. Actually, I already have Zacks ESP -4.22\% from a previous segment. Let me quickly visit the Zacks article to confirm the current most accurate estimate and any update since.\par
\vspace{1.5pt}
\noindent Actually, let me consider: the threshold is \(> \$0.39\), i.e., actual adjusted EPS must be $\geq$ \$0.40. Consensus is \$0.39 (Public.com, MarketWatch) to \$0.40 (Zacks/Yahoo). Zacks ESP -4.22\% implies Most Accurate Estimate \textasciitilde{}\$0.383.\par
\vspace{1.5pt}
\noindent Let me visit the Zacks article for the latest details to firm up my belief.\par
\end{msgbox}
\begin{msgbox}{caseblue}
\casetag{Tool call}
\noindent\texttt{visit}
\begin{Verbatim}[   fontsize=\scriptsize,   breaklines=true,   breakanywhere=true,   breaksymbolleft={},   xleftmargin=0pt ]
{"url": ["https://www.zacks.com/stock/news/2982391/exploring-analyst-estimates-for-campbell-cpb-q4-earnings-beyond-revenue-and-eps", "https://finance.yahoo.com/markets/stocks/articles/campbells-q4-earnings-horizon-investors-144700799.html"], "goal": "Find the latest Q4 FY2026 consensus EPS estimate, most accurate estimate, earnings ESP, revision trends, and any analyst commentary on whether Campbell's will beat or miss the $0.39-$0.40 EPS threshold."}
\end{Verbatim}
\end{msgbox}
\begin{msgbox}{casegray}
\casetag{Tool result}
\noindent The useful information in https://www.zacks.com/stock/news/2982391/exploring-analyst-estimates-for-campbell-cpb-q4-earnings-beyond-revenue-and-eps for user goal Find the latest Q4 FY2026 consensus EPS estimate, most accurate estimate, earnings ESP, revision trends, and any analyst commentary on whether Campbell's will beat or miss the \$0.39-\$0.40 EPS threshold. as follows:\par
\vspace{1.5pt}
\noindent Evidence in page:\par
\noindent Wall Street analysts expect Campbell's (CPB Quick QuoteCPB - Free Report) to post quarterly earnings of \$0.40 per share in its upcoming report, which indicates a year-over-year decline of 35.5\%. Revenues are expected to be \$2.15 billion, down 7.3\% from the year-ago quarter.\par
\vspace{1.5pt}
\noindent The consensus EPS estimate for the quarter has been revised 2.4\% lower over the last 30 days to the current level. This reflects how the analysts covering the stock have collectively reevaluated their initial estimates during this timeframe.\par
\vspace{1.5pt}
\noindent Prior to a company's earnings release, it is of utmost importance to factor in any revisions made to the earnings projections. These revisions serve as a critical gauge for predicting potential investor behaviors with respect to the stock. Empirical studies consistently reveal a strong link between trends in earnings estimate revisions and the short-term price performance of a stock.\par
\vspace{1.5pt}
\noindent While investors typically rely on consensus earnings and revenue estimates to gauge how the business may have fared during the quarter, examining analysts' projections for some of the company's key metrics often helps gain a deeper insight.\par
\vspace{1.5pt}
\noindent Bearing this in mind, let's now explore the average estimates of specific Campbell metrics that are commonly monitored and projected by Wall Street analysts.\par
\vspace{1.5pt}
\noindent Analysts' assessment points toward 'Net Sales- Meals \& Beverages' reaching \$1.19 billion. The estimate indicates a year-over-year change of -1.1\%.\par
\vspace{1.5pt}
\noindent The consensus among analysts is that 'Net Sales- Snacks' will reach \$954.75 million. The estimate indicates a change of -14.7\% from the prior-year quarter.\par
\vspace{1.5pt}
\noindent Analysts forecast 'Operating Earnings- Meals \& Beverages' to reach \$175.84 million. The estimate compares to the year-ago value of \$200.00 million.\par
\vspace{1.5pt}
\noindent The consensus estimate for 'Operating Earnings- Snacks' stands at \$99.38 million. Compared to the current estimate, the company reported \$159.00 million in the same quarter of the previous year.\par
\vspace{1.5pt}
\noindent Shares of Campbell have experienced a change of +6.4\% in the past month compared to the +3.9\% move of the Zacks S\&P 500 composite. With a Zacks Rank \#4 (Sell), CPB is expected to underperform the overall market in the near future. You can see the complete list of today's Zacks Rank \#1 (Strong Buy) stocks here >>>>\par
\vspace{1.5pt}
\noindent Summary:\par
\noindent The article reports the Q4 FY2026 consensus EPS estimate for Campbell's is \$0.40, down 35.5\% year-over-year, with revenues of \$2.15 billion, down 7.3\%. The EPS estimate has been revised 2.4\% lower over the last 30 days. The stock carries a Zacks Rank \#4 (Sell), indicating an expectation of underperformance relative to the market. However, the article does not provide the most accurate estimate, the earnings ESP, or any direct analyst commentary on whether the company will beat or miss the \$0.39-\$0.40 EPS threshold.\par
\vspace{1.5pt}
\noindent =======\par
\noindent The useful information in https://finance.yahoo.com/markets/stocks/articles/campbells-q4-earnings-horizon-investors-144700799.html for user goal Find the latest Q4 FY2026 consensus EPS estimate, most accurate estimate, earnings ESP, revision trends, and any analyst commentary on whether Campbell's will beat or miss the \$0.39-\$0.40 EPS threshold. as follows:\par
\vspace{1.5pt}
\noindent Evidence in page:\par
\noindent The consensus mark for earnings has remained unchanged over the past 30 days at 40 cents a share, which implies a decline of 35.5\% from the figure reported in the year-ago period. CPB has a trailing four-quarter earnings surprise of about 2\%, on average. ... Our proven model doesn't conclusively predict an earnings beat for Campbell's this time. The combination of a positive Earnings ESP and a Zacks Rank \#1 (Strong Buy), 2 (Buy) or 3 (Hold) increases the odds of an earnings beat, which is not the case here. Campbell's carries a Zacks Rank \#4 (Sell) and has an Earnings ESP of -4.22\%.\par
\vspace{1.5pt}
\noindent Summary:\par
\noindent The Zacks Consensus Estimate for Campbell's Q4 FY2026 EPS is \$0.40, unchanged over the past 30 days. The Earnings ESP is -4.22\%, and the stock carries a Zacks Rank \#4 (Sell). The model does not predict an earnings beat, and the negative ESP suggests the actual result could fall short of the \$0.40 consensus, potentially missing the \$0.39-\$0.40 threshold. No separate 'most accurate estimate' is provided in the article.\par
\end{msgbox}
\begin{msgbox}{casegreen}
\casetag{Thinking}
\noindent Confirmed details:\par
\noindent - Zacks consensus: \$0.40, unchanged over past 30 days\par
\noindent - Earnings ESP: -4.22\% (most accurate estimate \textasciitilde{}\$0.383)\par
\noindent - Zacks Rank \#4 (Sell)\par
\noindent - Trailing four-quarter earnings surprise \textasciitilde{}2\% on average\par
\noindent - Revenue expected \$2.15B, -7.3\% YoY\par
\noindent - Public.com shows consensus \$0.39\par
\vspace{1.5pt}
\noindent The threshold is \(> \$0.39\), i.e., actual adjusted EPS must be $\geq$ \$0.40 (rounded to nearest cent).\par
\vspace{1.5pt}
\noindent Let me think about the probability:\par
\noindent - Consensus is \$0.39-\$0.40. The exact threshold matters: actual adjusted EPS must exceed \$0.39.\par
\noindent - Zacks ESP of -4.22\% suggests the "most accurate" estimate is about \$0.383, i.e., likely a miss vs \$0.39.\par
\noindent - Downward revisions (2.4\% over 30 days per Zacks article, though Yahoo says consensus unchanged at \$0.40 over past 30 days --- slight contradiction; the Zacks article says revised 2.4\% lower to current level of \$0.40, so from \textasciitilde{}\$0.417 to \$0.40).\par
\vspace{1.5pt}
\noindent Historical distribution: for a stock with consensus at threshold, negative ESP, Rank 4, downward revisions, the probability of beating consensus is typically \textasciitilde{}35-40\%. But the question is whether actual \(> \$0.39\), and consensus is \$0.39-\$0.40.\par
\vspace{1.5pt}
\noindent If the "true" expected value is around \$0.385-\$0.40, then P(\(> \$0.39\)) is roughly 40-50\%. The negative ESP specifically points to the Most Accurate Estimate being \textasciitilde{}\$0.383, below \$0.39.\par
\vspace{1.5pt}
\noindent Also considering:\par
\noindent - Company guidance midpoint implied \textasciitilde{}\$0.42 for Q4 (from \$2.15-\$2.25 full year minus \$1.78 actual Q1-Q3). But guidance is often conservative.\par
\noindent - CPB beat consensus in 2 of last 3 quarters.\par
\noindent - Trailing 4-quarter surprise \textasciitilde{}2\% average (slightly positive).\par
\vspace{1.5pt}
\noindent But the Zacks model explicitly says not a beat candidate. Q2 FY26 was a significant miss (-10.6\%).\par
\vspace{1.5pt}
\noindent Also important: the resolution source is company official docs or SeekingAlpha. Adjusted EPS is definitely reported by Campbell's in its official earnings release, so no resolution risk.\par
\vspace{1.5pt}
\noindent One consideration: rounding. "All figures are rounded to the nearest cent." Campbell's reports adjusted EPS to the cent. So actual needs to be $\geq$ \$0.40 to resolve YES.\par
\vspace{1.5pt}
\noindent My current belief: 0.33 YES. The new evidence (consensus \$0.40 unchanged, still negative ESP -4.22\%, Rank 4, no pre-announcement) is roughly consistent with prior. The trailing 4-quarter surprise of +2\% is a mild positive offset.\par
\vspace{1.5pt}
\noindent Let me think about base rates more carefully. For companies with consensus at X and ESP negative:\par
\noindent - Zacks research: stocks with negative ESP and Rank 4-5 tend to miss consensus \textasciitilde{}60-70\% of the time.\par
\noindent - But "miss consensus" doesn't exactly equal "below \$0.39" since consensus is \$0.39-\$0.40 and the threshold is \(> \$0.39\).\par
\vspace{1.5pt}
\noindent If actual EPS distribution is centered around \textasciitilde{}\$0.385 with std \textasciitilde{}\$0.04, then P($\geq$\$0.40) $\approx$ $\Phi$((0.385-0.40)/0.04) $\approx$ $\Phi$(-0.375) $\approx$ 0.35.\par
\vspace{1.5pt}
\noindent Considering the guidance midpoint offset and beat history, maybe slightly higher. I'll settle around 0.33-0.35. Let me finalize at 0.34.\par
\vspace{1.5pt}
\noindent Actually, let me reconsider one thing: the public.com consensus of \$0.39 vs Zacks/Yahoo \$0.40. If the true consensus is \$0.39 (lower), then P(\(> \$0.39\)) would be lower, \textasciitilde{}30\%. If \$0.40, \textasciitilde{}40\%. Blending: \textasciitilde{}33-35\%.\par
\vspace{1.5pt}
\noindent I'll finalize at 0.34 YES / 0.66 NO. Let me update the notebook and give final answer.\par
\end{msgbox}
\begin{msgbox}{caseblue}
\casetag{Fold}
\noindent\texttt{notebook}
\begin{Verbatim}[   fontsize=\scriptsize,   breaklines=true,   breakanywhere=true,   breaksymbolleft={},   xleftmargin=0pt ]
{
  "evidence": [
    {
      "conclusion": "Consensus confirmed at $0.39-$0.40 with negative ESP and Sell rank; no beat expected",
      "support": [
        "Zacks (2 days ago): Q4 FY2026 consensus EPS $0.40, revised 2.4% lower over last 30 days; revenue $2.15B, -7.3% YoY; Zacks Rank #4 (Sell); 'not a compelling beat candidate'",
        "Yahoo/Zacks model: consensus unchanged at $0.40 over past 30 days; Earnings ESP -4.22%; trailing 4-quarter earnings surprise ~+2%; model does not predict a beat",
        "Public.com shows Q4 2026 (Sep 3, 2026) consensus EPS $0.39, Q3 2026 actual $0.50 vs $0.48 est (+4.17%)",
        "MarketBeat Aug 13, 2026: consensus 'Reduce' rating from 19 brokerages (7 sells, 12 holds)"
      ]
    },
    {
      "conclusion": "No pre-announcement or late shock; timing confirmed Sept 3, 2026",
      "support": [
        "Businesswire/Morningstar Aug 13, 2026: Campbell's will report Q4 and full-year FY2026 results on September 3, 2026; no pre-announcement of results found",
        "No analyst estimate changes or guidance updates surfaced in August 2026 news scan"
      ]
    }
  ],
  "foresight": {
    "conclusions": [
      {
        "conclusion": "Report timing confirmed for Sept 3, 2026; no pre-announcement; resolution-timing risk minimal",
        "influence": "Neutral"
      },
      {
        "conclusion": "Consensus at the exact threshold: $0.39 (Public.com/MarketWatch) to $0.40 (Zacks/Yahoo); threshold requires actual >= $0.40",
        "influence": "High --- near coin-flip centered on $0.39-$0.40"
      },
      {
        "conclusion": "Zacks Earnings ESP -4.22% (Most Accurate Estimate ~$0.383), Rank #4 (Sell), model explicitly not predicting a beat",
        "influence": "High --- bearish, lowers P(actual > $0.39)"
      },
      {
        "conclusion": "Estimates revised down 2.4% over 30 days; consensus 'Reduce' rating; -35.5% YoY EPS decline and -7.3% revenue decline expected",
        "influence": "Medium-high --- downward revision trend favors miss"
      },
      {
        "conclusion": "Offsetting: trailing 4-qtr surprise +2%, beat consensus in 2 of last 3 quarters, FY26 guidance midpoint implies ~$0.42 Q4",
        "influence": "Medium --- provides partial counterweight"
      }
    ],
    "rationale": "With consensus sitting exactly at the $0.39-$0.40 threshold and the market requiring >$0.39 (i.e., >=$0.40 after rounding), the outcome hinges on a small surprise. Bearish signals dominate: -4.22% ESP (~$0.383 most accurate), Rank 4, 2.4% downward 30-day revisions, weak YoY expectations, and Zacks explicitly ruling out a beat. Offsetting mildly: company guidance midpoint ~$0.42, +2% average trailing surprise, and 2-of-3 recent beats. A distribution centered ~$0.385-$0.40 with realistic std gives P(>=$0.40) ~ 0.33-0.35. Final: 0.34 YES / 0.66 NO.",
    "probabilities": [
      {
        "option": "YES (adjusted EPS > $0.39, i.e., >= $0.40)",
        "probability": 0.34
      },
      {
        "option": "NO (adjusted EPS <= $0.39)",
        "probability": 0.66
      }
    ]
  },
  "search_history": "**[Segment 2]** Verified latest estimates and news: Zacks/Yahoo consensus $0.40 (revised down 2.4% over 30 days, revenue $2.15B -7.3% YoY), Public.com consensus $0.39, Q3 2026 actual $0.50 vs $0.48 (+4.17%). Zacks model: ESP -4.22%, Rank #4 (Sell), explicitly not a beat candidate; trailing 4-quarter surprise ~+2%. MarketBeat: consensus 'Reduce' rating (19 brokerages, 7 sells). Businesswire Aug 13, 2026 reconfirms Sept 3, 2026 report date; no pre-announcement or late analyst shock found in August 2026.",
  "todo": []
}
\end{Verbatim}
\end{msgbox}

\vspace{4pt}\noindent\colorbox{casegreen}{\parbox{\dimexpr\linewidth-2\fboxsep}{\rmfamily\scriptsize\bfseries Segment 3 of 4}}\vspace{1pt}

\begin{msgbox}{casegreen}
\casetag{Rendered belief state (after context reset)}
\noindent Current Notebook:\par
\noindent \# Notebook\par
\vspace{1.5pt}
\noindent \#\# Foresight\par
\noindent \#\#\# Conclusions (with influence)\par
\noindent - Report timing confirmed for Sept 3, 2026; no pre-announcement; resolution-timing risk minimal (influence: Neutral)\par
\noindent - Consensus at the exact threshold: \$0.39 (Public.com/MarketWatch) to \$0.40 (Zacks/Yahoo); threshold requires actual $\geq$ \$0.40 (influence: High --- near coin-flip centered on \$0.39-\$0.40)\par
\noindent - Zacks Earnings ESP -4.22\% (Most Accurate Estimate \textasciitilde{}\$0.383), Rank \#4 (Sell), model explicitly not predicting a beat (influence: High --- bearish, lowers P(actual \(> \$0.39\)))\par
\noindent - Estimates revised down 2.4\% over 30 days; consensus 'Reduce' rating; -35.5\% YoY EPS decline and -7.3\% revenue decline expected (influence: Medium-high --- downward revision trend favors miss)\par
\noindent - Offsetting: trailing 4-qtr surprise +2\%, beat consensus in 2 of last 3 quarters, FY26 guidance midpoint implies \textasciitilde{}\$0.42 Q4 (influence: Medium --- provides partial counterweight)\par
\noindent \#\#\# Rationale\par
\noindent With consensus sitting exactly at the \$0.39-\$0.40 threshold and the market requiring \(> \$0.39\) (i.e., $\geq$\$0.40 after rounding), the outcome hinges on a small surprise. Bearish signals dominate: -4.22\% ESP (\textasciitilde{}\$0.383 most accurate), Rank 4, 2.4\% downward 30-day revisions, weak YoY expectations, and Zacks explicitly ruling out a beat. Offsetting mildly: company guidance midpoint \textasciitilde{}\$0.42, +2\% average trailing surprise, and 2-of-3 recent beats. A distribution centered \textasciitilde{}\$0.385-\$0.40 with realistic std gives P($\geq$\$0.40) $\approx$ 0.33-0.35. Final: 0.34 YES / 0.66 NO.\par
\noindent \#\#\# Current option probabilities\par
\noindent - YES (adjusted EPS \(> \$0.39\), i.e., $\geq$ \$0.40): 0.34\par
\noindent - NO (adjusted EPS $\leq$ \$0.39): 0.66\par
\vspace{1.5pt}
\noindent \#\# Evidence\par
\noindent - Conclusion: Consensus confirmed at \$0.39-\$0.40 with negative ESP and Sell rank; no beat expected\par
\noindent     - Zacks (2 days ago): Q4 FY2026 consensus EPS \$0.40, revised 2.4\% lower over last 30 days; revenue \$2.15B, -7.3\% YoY; Zacks Rank \#4 (Sell); 'not a compelling beat candidate'\par
\noindent     - Yahoo/Zacks model: consensus unchanged at \$0.40 over past 30 days; Earnings ESP -4.22\%; trailing 4-quarter earnings surprise \textasciitilde{}+2\%; model does not predict a beat\par
\noindent     - Public.com shows Q4 2026 (Sep 3, 2026) consensus EPS \$0.39, Q3 2026 actual \$0.50 vs \$0.48 est (+4.17\%)\par
\noindent     - MarketBeat Aug 13, 2026: consensus 'Reduce' rating from 19 brokerages (7 sells, 12 holds)\par
\noindent - Conclusion: No pre-announcement or late shock; timing confirmed Sept 3, 2026\par
\noindent     - Businesswire/Morningstar Aug 13, 2026: Campbell's will report Q4 and full-year FY2026 results on September 3, 2026; no pre-announcement of results found\par
\noindent     - No analyst estimate changes or guidance updates surfaced in August 2026 news scan\par
\vspace{1.5pt}
\noindent \#\# Search history\par
\noindent **[Segment 1]** Segment 1: Established report timing (Sept 3, 2026 confirmed), consensus estimates (\$0.39-\$0.40, revised down 2.4\% in 30 days), Q3 FY26 results (adj EPS \$0.50 vs \$0.48-0.49 est), Q2 FY26 (adj EPS \$0.51 vs \$0.57 est --- miss), Q1 FY26 (0.77 vs 0.74 --- beat), FY26 guidance \$2.15-\$2.25 implies Q4 \textasciitilde{}\$0.42 midpoint, Zacks ESP -4.22\% with Rank 4, Zacks explicitly not expecting a beat, revenue -7.3\% YoY expected, tariff/cost pressures. EarningsWhispers generic page showed \$0.39 whisper but appears stale/unreliable.\par
\noindent **[Segment 2]** **[Segment 2]** Verified latest estimates and news: Zacks/Yahoo consensus \$0.40 (revised down 2.4\% over 30 days, revenue \$2.15B -7.3\% YoY), Public.com consensus \$0.39, Q3 2026 actual \$0.50 vs \$0.48 (+4.17\%). Zacks model: ESP -4.22\%, Rank \#4 (Sell), explicitly not a beat candidate; trailing 4-quarter surprise \textasciitilde{}+2\%. MarketBeat: consensus 'Reduce' rating (19 brokerages, 7 sells). Businesswire Aug 13, 2026 reconfirms Sept 3, 2026 report date; no pre-announcement or late analyst shock found in August 2026.\par
\vspace{1.5pt}
\noindent \#\# To-do (remaining search directions)\par
\noindent (none)\par
\end{msgbox}
\begin{msgbox}{casegreen}
\casetag{Thinking}
\noindent Let me review the current state. The notebook is already well-developed. The question: Will Campbell's (CPB) report non-GAAP EPS greater than \$0.39 for Q4 FY2026 (report estimated Sept 3, 2026)?\par
\vspace{1.5pt}
\noindent Current belief: YES 0.34 / NO 0.66.\par
\vspace{1.5pt}
\noindent Key facts:\par
\noindent - Consensus \$0.39-\$0.40\par
\noindent - Zacks ESP -4.22\% (most accurate estimate \textasciitilde{}\$0.383), Rank \#4 Sell\par
\noindent - Estimates revised down 2.4\% over 30 days\par
\noindent - Trailing 4-quarter surprise +2\%, beat in 2 of last 3\par
\noindent - Guidance midpoint implies \textasciitilde{}\$0.42 Q4\par
\vspace{1.5pt}
\noindent Today is 2026-09-02, one day before the report. The notebook says to-do is none. Should I do one more segment to check for any last-minute news (e.g., analyst preview articles, whisper numbers)? The report is tomorrow, so there might be fresh preview coverage. Let me do one quick final segment to check for any last-minute updates --- that could meaningfully change belief. Then finalize.\par
\vspace{1.5pt}
\noindent Let me search for the latest previews.\par
\end{msgbox}
\begin{msgbox}{casegreen}
\casetag{Response}
\noindent The notebook is nearly complete, with the report coming tomorrow (Sept 3, 2026). Let me do one final check for any last-minute analyst previews or whisper-number updates before finalizing.\par
\end{msgbox}
\begin{msgbox}{caseblue}
\casetag{Tool call}
\noindent\texttt{search}
\begin{Verbatim}[   fontsize=\scriptsize,   breaklines=true,   breakanywhere=true,   breaksymbolleft={},   xleftmargin=0pt ]
{"query": ["Campbell's earnings preview September 3 2026 analyst expectations", "CPB Q4 2026 earnings whisper number", "Campbell's stock August 2026 analyst preview EPS"]}
\end{Verbatim}
\end{msgbox}
\begin{msgbox}{casegray}
\casetag{Tool result}
\noindent A Google search for 'Campbell's earnings preview September 3 2026 analyst expectations' found 8 results:\par
\vspace{1.5pt}
\noindent \#\# Web Results\par
\noindent 1. [Investor Relations - The Campbell's Company](https://investor.thecampbellscompany.com/investor-relations/)\par
\vspace{1.5pt}
\noindent September 3, 2026 Earnings Call. Latest Quarterly Results Show all Q3 Fiscal 2026. Fiscal 2026 Results on September 3, 2026\par
\vspace{1.5pt}
\noindent 2. [CPB - Campbell Soup Company Stock Price](https://www.perplexity.ai/finance/CPB?comparing=CPB,MDLZ,POST,FLO,BRF,VITL)\par
\vspace{1.5pt}
\noindent September 3, when analysts expect a steep 35.5\% year-over-year EPS decline to \$0.40 and a 7.3\% revenue drop to \$2.15 billion.\par
\vspace{1.5pt}
\noindent 3. [The Campbell's Company to Report Fourth Quarter and ...](https://finance.yahoo.com/markets/stocks/articles/campbell-company-report-fourth-quarter-120000062.html)\par
\noindent Date published: Aug 13, 2026\par
\vspace{1.5pt}
\noindent The Campbell's Company to Report Fourth Quarter and Full Year Fiscal 2026 Results on September 3, 2026. A press release, slide presentation, ...\par
\vspace{1.5pt}
\noindent 4. [Campbell Soup (CPB) Earnings: Latest Report, Earnings Call ...](https://public.com/stocks/cpb/earnings)\par
\vspace{1.5pt}
\noindent For Q3 2026, Campbell Soup (CPB) reported an EPS of \$0.50, exceeding analysts' estimate of \$0.48 by 4.17\%. Did Campbell Soup (CPB) beat earnings expectations ...\par
\vspace{1.5pt}
\noindent 5. [Exploring Analyst Estimates for Campbell (CPB) Q4 ...](https://www.zacks.com/stock/news/2982391/exploring-analyst-estimates-for-campbell-cpb-q4-earnings-beyond-revenue-and-eps)\par
\noindent Date published: 2 days ago\par
\vspace{1.5pt}
\noindent Wall Street analysts expect Campbell's ( CPB Quick Quote CPB - Free Report) to post quarterly earnings of \$0.40 per share in its upcoming ...\par
\vspace{1.5pt}
\noindent 6. [Campbell's Past Earnings Performance](https://simplywall.st/stocks/us/food-beverage-tobacco/nasdaq-cpb/campbells/past)\par
\vspace{1.5pt}
\noindent First quarter 2026 earnings: EPS misses analyst expectations. First quarter 2026 results: EPS: US\$0.65 (down from US\$0.73 in 1Q 2025). Revenue: US\$2.68b (down ...\par
\vspace{1.5pt}
\noindent 7. [CPB Stock Quote Price and Forecast](https://www.cnn.com/markets/stocks/CPB)\par
\noindent Date published: 1 day ago\par
\vspace{1.5pt}
\noindent Campbell's price target lowered to \$22 from \$23 at Evercore ISI. Earnings Moves This Week, August 31 -- September 03, 2026 by TipRanks\par
\vspace{1.5pt}
\noindent 8. [CPB | Campbell's Co. Analyst Estimates \& Ratings](https://www.wsj.com/market-data/quotes/CPB/research-ratings)\par
\vspace{1.5pt}
\noindent 2026 Estimate Trends Current: \$0.39 1 month ago: \$0.39 3 months ago: \$0.41 Q1 2027 Estimate Trends Current: \$0.58 1 month ago: \$0.60 3 months ago: \$0.67\par
\noindent =======\par
\noindent A Google search for 'CPB Q4 2026 earnings whisper number' found 8 results:\par
\vspace{1.5pt}
\noindent \#\# Web Results\par
\noindent 1. [Earnings Whispers](https://www.earningswhispers.com/)\par
\vspace{1.5pt}
\noindent The Earnings Whisper number was \$0.97 per share. earnings of \$0.48 per share on revenue of \$1.17 billion for the fiscal fourth quarter ended June 2026. The ...\par
\vspace{1.5pt}
\noindent 2. [Earnings Whispers (@eWhispers) / X](https://x.com/eWhispers)\par
\vspace{1.5pt}
\noindent The most anticipated earnings releases for the week of August 31, 2026, are Broadcom 12. Aug 29 194 142K. Aug 27 No, \$4.62\par
\vspace{1.5pt}
\noindent 3. [The most anticipated earnings releases for the week ...](https://www.facebook.com/EarningsWhispers/posts/the-most-anticipated-earnings-releases-for-the-week-of-april-27-2026-are-microso/1382029863963067/)\par
\noindent Date published: 4 months ago\par
\vspace{1.5pt}
\noindent The most anticipated earnings releases for the week of April 27, 2026, are Microsoft \#MSFT, Amazon \#AMZN, Apple \#AAPL, Meta Platforms \#META, ...\par
\vspace{1.5pt}
\noindent 4. [Campbell's (CPB) Q4 Earnings Expected to Decline](https://finance.yahoo.com/markets/stocks/articles/earnings-preview-campbells-cpb-q4-140002239.html)\par
\noindent Date published: 6 days ago\par
\vspace{1.5pt}
\noindent reports results for the quarter ended July 2026. earnings of \$0.40 per share. Revenues are expected to be \$2.15 billion, down 7.3\% from the ...\par
\vspace{1.5pt}
\noindent 5. [r/EarningsWhisper](https://www.reddit.com/r/EarningsWhisper/)\par
\vspace{1.5pt}
\noindent The Most Anticipated Earnings Releases for the Week of August 24, 2026. 113 $\cdot$ 18 ; Earnings Whisper Grades during the Second Quarter 2026. 18 $\cdot$ 0 ; Just a reminder:.\par
\vspace{1.5pt}
\noindent 6. [Earnings Whispers | Earnings Expectations, Calendar ...](https://beta.earningswhispers.com/)\par
\vspace{1.5pt}
\noindent The whisper number of \$3.35, modestly above consensus, suggests some optimism embedded in trading circles, but it is not a dramatic gap. Given that Broadcom's ...\par
\vspace{1.5pt}
\noindent 7. [Campbell's (CPB) Q4 Earnings Expected to Decline](https://www.zacks.com/stock/news/2981033/earnings-preview-campbells-cpb-q4-earnings-expected-to-decline)\par
\noindent Date published: 6 days ago\par
\vspace{1.5pt}
\noindent post quarterly earnings of \$0.40 per share. Revenues are expected to be \$2.15 billion, down 7.3\% from the year-ago quarter.\par
\vspace{1.5pt}
\noindent 8. [Campbell Soup (CPB) Earnings: Latest Report, Earnings Call ...](https://public.com/stocks/cpb/earnings)\par
\vspace{1.5pt}
\noindent Campbell Soup (CPB) Earnings History. Earnings. Est. EPS, Actual EPS, Surprise. Q4 2026 Sep 3, 2026. \$0.39, Next earnings call Sep. 3, 2026. Q3 2026 Jun 8, 2026.\par
\noindent =======\par
\noindent A Google search for 'Campbell's stock August 2026 analyst preview EPS' found 8 results:\par
\vspace{1.5pt}
\noindent \#\# Web Results\par
\noindent 1. [Investor Relations - The Campbell's Company](https://investor.thecampbellscompany.com/investor-relations/)\par
\vspace{1.5pt}
\noindent The Campbell's Company Investor relations (IR) website contains information about our business for stockholders, potential investors, and financial ...\par
\vspace{1.5pt}
\noindent 2. [Exploring Analyst Estimates for Campbell (CPB) Q4 ...](https://www.zacks.com/stock/news/2982391/exploring-analyst-estimates-for-campbell-cpb-q4-earnings-beyond-revenue-and-eps)\par
\noindent Date published: 2 days ago\par
\vspace{1.5pt}
\noindent Wall Street analysts expect Campbell's ( CPB Quick Quote CPB - Free Report) to post quarterly earnings of \$0.40 per share in its upcoming ...\par
\vspace{1.5pt}
\noindent 3. [Campbell's (NasdaqGS:CPB) Stock Forecast \& Analyst ...](https://simplywall.st/stocks/us/food-beverage-tobacco/nasdaq-cpb/campbells/future)\par
\noindent Date published: 3 days ago\par
\vspace{1.5pt}
\noindent Campbell's is forecast to grow earnings and revenue by 11.2\% and 1.4\% per annum respectively. EPS is expected to grow by 12.3\% per annum. Return ...\par
\vspace{1.5pt}
\noindent 4. [Campbell's (NASDAQ:CPB) Price Target Raised to \$22.00 ...](https://www.marketbeat.com/instant-alerts/campbells-nasdaqcpb-price-target-raised-to-2200-at-jpmorgan-chase-co-2026-08-18/)\par
\noindent Date published: Aug 18, 2026\par
\vspace{1.5pt}
\noindent Campbell's has set its FY 2026 guidance at 2.150-2.250 EPS. Equities analysts predict that Campbell's will post 2.18 earnings per share for ...\par
\vspace{1.5pt}
\noindent 5. [The Campbell's Company (CPB)](https://finance.yahoo.com/quote/CPB/analysis/)\par
\vspace{1.5pt}
\noindent REVENUE\_ESTIMATE $\cdot$ NUM\_ANALYSTS, 15 ; EARNINGS\_ESTIMATE $\cdot$ NUM\_ANALYSTS, 18 ; EARNINGS\_HISTORY $\cdot$ EPS\_ESTIMATE, 0.57 ; EPS\_TREND $\cdot$ CURRENT\_ESTIMATE, 0.39 ; EPS\_REVISIONS.\par
\vspace{1.5pt}
\noindent 6. [CPB Stock Forecast 2026-2027 - The Campbell's Co. Price ...](https://tickernerd.com/stock/cpb-forecast/)\par
\noindent Date published: 5 days ago\par
\vspace{1.5pt}
\noindent The Campbell's Co. (CPB) has a median analyst price target of \$21 with a forecast range of \$14-\$56 from 20 Wall St analysts.\par
\vspace{1.5pt}
\noindent 7. [CPB 2026 Earnings \& EPS History](https://fullratio.com/stocks/nasdaq-cpb/earnings)\par
\vspace{1.5pt}
\noindent As of August 2026, CPB's current EPS stands at \$2.05, according to the latest financial reports. For the year 2025, the company recorded an earnings per share ( ...\par
\vspace{1.5pt}
\noindent 8. [CPB Stock Quote Price and Forecast](https://www.cnn.com/markets/stocks/CPB)\par
\noindent Date published: 1 day ago\par
\vspace{1.5pt}
\noindent Campbell's Company ; Price Momentum. CPB is trading in the middle ; Price change. The price of CPB shares has increased \$0.03 ; Closed at \$23.71.\par
\end{msgbox}
\begin{msgbox}{casegreen}
\casetag{Thinking}
\noindent New info from this final segment:\par
\noindent - WSJ: Q4 2026 estimate trends: Current \$0.39, 1 month ago \$0.39, 3 months ago \$0.41 --- confirms downward revision trend (from \$0.41 to \$0.39 over 3 months).\par
\noindent - Yahoo analysis page: EPS trend current estimate \$0.39, 15 analysts for revenue, 18 for earnings.\par
\noindent - Zacks preview (6 days ago): expected \$0.40, revenue \$2.15B -7.3\%.\par
\noindent - Public.com: Q4 2026 est EPS \$0.39.\par
\noindent - CNN: Evercore lowered price target to \$22 from \$23 (Aug \textasciitilde{}31); stock closed at \$23.71.\par
\noindent - MarketBeat Aug 18: JPMorgan raised price target to \$22; FY26 guidance 2.150-2.250, consensus \$2.18.\par
\vspace{1.5pt}
\noindent Nothing dramatically new. The picture remains: consensus \$0.39-\$0.40, downward revisions, negative ESP. The WSJ data shows current \$0.39 (more sources now at \$0.39 than \$0.40).\par
\vspace{1.5pt}
\noindent Key consideration: threshold is "greater than \$0.39" --- so actual needs to be $\geq$ \$0.40 (rounded to nearest cent). With consensus at \$0.39-0.40 and ESP -4.22\% (most accurate \textasciitilde{}\$0.383), P($\geq$\$0.40) is around 0.33-0.36.\par
\vspace{1.5pt}
\noindent Let me consider the distribution more carefully. Zacks "Most Accurate Estimate" \textasciitilde{} \$0.40 $\times$ (1 - 0.0422) $\approx$ \$0.383. Trailing surprise +2\%. If we center the distribution at \textasciitilde{}\$0.39 with a std of \textasciitilde{}\$0.025-0.03, P(actual $\geq$ \$0.40) $\approx$ 0.30-0.35.\par
\vspace{1.5pt}
\noindent Historical base rate: companies beat non-GAAP consensus \textasciitilde{}60-70\% of the time, but the ESP signal here is strongly negative (-4.22\%, Rank 4 Sell, "not a compelling beat candidate"). Downward revisions throughout the quarter. But note: guidance midpoint implied Q4 \textasciitilde{}\$0.42, and company guidance is usually conservative (sandbagged). Q3 beat by 4.17\%.\par
\vspace{1.5pt}
\noindent Combining: I'll keep \textasciitilde{}0.34 YES / 0.66 NO. That seems right. Maybe 0.33-0.35. I'll finalize at 0.34.\par
\vspace{1.5pt}
\noindent Let me fold this segment into the notebook and then give the final answer.\par
\end{msgbox}
\begin{msgbox}{caseblue}
\casetag{Fold}
\noindent\texttt{notebook}
\begin{Verbatim}[   fontsize=\scriptsize,   breaklines=true,   breakanywhere=true,   breaksymbolleft={},   xleftmargin=0pt ]
{
  "foresight": {
    "conclusions": [
      {
        "conclusion": "Report timing confirmed for Sept 3, 2026; no pre-announcement; resolution-timing risk minimal",
        "influence": "Neutral"
      },
      {
        "conclusion": "Consensus at the exact threshold: $0.39 (WSJ/Yahoo/Public.com) to $0.40 (Zacks); threshold requires actual >= $0.40 after rounding",
        "influence": "High --- near coin-flip centered on $0.39-$0.40"
      },
      {
        "conclusion": "Zacks Earnings ESP -4.22% (Most Accurate Estimate ~$0.383), Rank #4 (Sell), model explicitly not predicting a beat",
        "influence": "High --- bearish, lowers P(actual > $0.39)"
      },
      {
        "conclusion": "Estimates revised down over time: WSJ shows $0.41 (3 mo ago) -> $0.39 (current); Zacks 2.4% 30-day downward revision; consensus 'Reduce' rating",
        "influence": "Medium-high --- downward revision trend favors miss"
      },
      {
        "conclusion": "Offsetting: trailing 4-qtr surprise +2%, beat consensus in 2 of last 3 quarters, FY26 guidance midpoint implies ~$0.42 Q4, JPMorgan raised PT to $22",
        "influence": "Medium --- provides partial counterweight"
      }
    ],
    "probabilities": [
      {
        "option": "YES (adjusted EPS > $0.39, i.e., >= $0.40)",
        "probability": 0.34
      },
      {
        "option": "NO (adjusted EPS <= $0.39)",
        "probability": 0.66
      }
    ],
    "rationale": "Final check one day before the report confirms the existing picture with no late-breaking positive news (Evercore actually cut PT to $22). Consensus now sits at $0.39 on WSJ/Yahoo/ Public.com and $0.40 on Zacks, with the threshold requiring >=$0.40. The distribution of likely outcomes is centered ~$0.385-$0.40 with realistic std ~$0.025-$0.03, giving P(>=$0.40) ~ 0.33-0.35. Bearish drivers (ESP -4.22%, Rank 4 Sell, 'not a compelling beat candidate', 3-month revision from $0.41->$0.39, -7.3% YoY revenue decline) are weighed against mild offsets (guidance midpoint ~$0.42, +2% trailing surprise, recent beat frequency). Final: 0.34 YES / 0.66 NO."
  },
  "evidence": [
    {
      "conclusion": "Latest estimates confirm threshold-level consensus with continued downward drift; no late positive catalyst",
      "support": [
        "WSJ CPB estimate trends: Q4 2026 current $0.39, 1 month ago $0.39, 3 months ago $0.41 (downward drift confirmed)",
        "Yahoo analysis page: EPS trend current estimate $0.39, 18 analysts on earnings",
        "Zacks preview (6 days ago): expected EPS $0.40, revenue $2.15B -7.3% YoY; 'not a compelling beat candidate'",
        "CNN (1 day ago): Evercore lowered CPB price target to $22 from $23; stock closed $23.71",
        "MarketBeat Aug 18, 2026: JPMorgan raised CPB price target to $22; FY26 guidance 2.150-2.250, consensus $2.18"
      ]
    }
  ],
  "search_history": "**[Segment 3]** Final pre-report check (one day before Sept 3, 2026 report): WSJ shows Q4 2026 estimate trending down from $0.41 (3 months ago) to $0.39 (current); Yahoo current estimate $0.39 with 18 analysts; Zacks preview reiterates $0.40 consensus and negative beat candidacy; Evercore cut PT to $22 (from $23) as of ~Aug 31; JPMorgan raised PT to $22 on Aug 18. No pre-announcement, no last-minute estimate pops. Picture unchanged: threshold requires >=$0.40 with consensus right at $0.39-$0.40 and negative ESP.",
  "todo": []
}
\end{Verbatim}
\end{msgbox}

\vspace{4pt}\noindent\colorbox{casegreen}{\parbox{\dimexpr\linewidth-2\fboxsep}{\rmfamily\scriptsize\bfseries Segment 4 of 4}}\vspace{1pt}

\begin{msgbox}{casegreen}
\casetag{Rendered belief state (after context reset)}
\noindent Current Notebook:\par
\noindent \# Notebook\par
\vspace{1.5pt}
\noindent \#\# Foresight\par
\noindent \#\#\# Conclusions (with influence)\par
\noindent - Report timing confirmed for Sept 3, 2026; no pre-announcement; resolution-timing risk minimal (influence: Neutral)\par
\noindent - Consensus at the exact threshold: \$0.39 (WSJ/Yahoo/Public.com) to \$0.40 (Zacks); threshold requires actual $\geq$ \$0.40 after rounding (influence: High --- near coin-flip centered on \$0.39-\$0.40)\par
\noindent - Zacks Earnings ESP -4.22\% (Most Accurate Estimate \textasciitilde{}\$0.383), Rank \#4 (Sell), model explicitly not predicting a beat (influence: High --- bearish, lowers P(actual \(> \$0.39\)))\par
\noindent - Estimates revised down over time: WSJ shows \$0.41 (3 mo ago) $\rightarrow$ \$0.39 (current); Zacks 2.4\% 30-day downward revision; consensus 'Reduce' rating (influence: Medium-high --- downward revision trend favors miss)\par
\noindent - Offsetting: trailing 4-qtr surprise +2\%, beat consensus in 2 of last 3 quarters, FY26 guidance midpoint implies \textasciitilde{}\$0.42 Q4, JPMorgan raised PT to \$22 (influence: Medium --- provides partial counterweight)\par
\noindent \#\#\# Rationale\par
\noindent Final check one day before the report confirms the existing picture with no late-breaking positive news (Evercore actually cut PT to \$22). Consensus now sits at \$0.39 on WSJ/Yahoo/ Public.com and \$0.40 on Zacks, with the threshold requiring $\geq$\$0.40. The distribution of likely outcomes is centered \textasciitilde{}\$0.385-\$0.40 with realistic std \textasciitilde{}\$0.025-\$0.03, giving P($\geq$\$0.40) $\approx$ 0.33-0.35. Bearish drivers (ESP -4.22\%, Rank 4 Sell, 'not a compelling beat candidate', 3-month revision from \$0.41$\rightarrow$\$0.39, -7.3\% YoY revenue decline) are weighed against mild offsets (guidance midpoint \textasciitilde{}\$0.42, +2\% trailing surprise, recent beat frequency). Final: 0.34 YES / 0.66 NO.\par
\noindent \#\#\# Current option probabilities\par
\noindent - YES (adjusted EPS \(> \$0.39\), i.e., $\geq$ \$0.40): 0.34\par
\noindent - NO (adjusted EPS $\leq$ \$0.39): 0.66\par
\vspace{1.5pt}
\noindent \#\# Evidence\par
\noindent - Conclusion: Consensus confirmed at \$0.39-\$0.40 with negative ESP and Sell rank; no beat expected\par
\noindent     - Zacks (2 days ago): Q4 FY2026 consensus EPS \$0.40, revised 2.4\% lower over last 30 days; revenue \$2.15B, -7.3\% YoY; Zacks Rank \#4 (Sell); 'not a compelling beat candidate'\par
\noindent     - Yahoo/Zacks model: consensus unchanged at \$0.40 over past 30 days; Earnings ESP -4.22\%; trailing 4-quarter earnings surprise \textasciitilde{}+2\%; model does not predict a beat\par
\noindent     - Public.com shows Q4 2026 (Sep 3, 2026) consensus EPS \$0.39, Q3 2026 actual \$0.50 vs \$0.48 est (+4.17\%)\par
\noindent     - MarketBeat Aug 13, 2026: consensus 'Reduce' rating from 19 brokerages (7 sells, 12 holds)\par
\noindent - Conclusion: No pre-announcement or late shock; timing confirmed Sept 3, 2026\par
\noindent     - Businesswire/Morningstar Aug 13, 2026: Campbell's will report Q4 and full-year FY2026 results on September 3, 2026; no pre-announcement of results found\par
\noindent     - No analyst estimate changes or guidance updates surfaced in August 2026 news scan\par
\noindent - Conclusion: Latest estimates confirm threshold-level consensus with continued downward drift; no late positive catalyst\par
\noindent     - WSJ CPB estimate trends: Q4 2026 current \$0.39, 1 month ago \$0.39, 3 months ago \$0.41 (downward drift confirmed)\par
\noindent     - Yahoo analysis page: EPS trend current estimate \$0.39, 18 analysts on earnings\par
\noindent     - Zacks preview (6 days ago): expected EPS \$0.40, revenue \$2.15B -7.3\% YoY; 'not a compelling beat candidate'\par
\noindent     - CNN (1 day ago): Evercore lowered CPB price target to \$22 from \$23; stock closed \$23.71\par
\noindent     - MarketBeat Aug 18, 2026: JPMorgan raised CPB price target to \$22; FY26 guidance 2.150-2.250, consensus \$2.18\par
\vspace{1.5pt}
\noindent \#\# Search history\par
\noindent **[Segment 1]** Segment 1: Established report timing (Sept 3, 2026 confirmed), consensus estimates (\$0.39-\$0.40, revised down 2.4\% in 30 days), Q3 FY26 results (adj EPS \$0.50 vs \$0.48-0.49 est), Q2 FY26 (adj EPS \$0.51 vs \$0.57 est --- miss), Q1 FY26 (0.77 vs 0.74 --- beat), FY26 guidance \$2.15-\$2.25 implies Q4 \textasciitilde{}\$0.42 midpoint, Zacks ESP -4.22\% with Rank 4, Zacks explicitly not expecting a beat, revenue -7.3\% YoY expected, tariff/cost pressures. EarningsWhispers generic page showed \$0.39 whisper but appears stale/unreliable.\par
\noindent **[Segment 2]** **[Segment 2]** Verified latest estimates and news: Zacks/Yahoo consensus \$0.40 (revised down 2.4\% over 30 days, revenue \$2.15B -7.3\% YoY), Public.com consensus \$0.39, Q3 2026 actual \$0.50 vs \$0.48 (+4.17\%). Zacks model: ESP -4.22\%, Rank \#4 (Sell), explicitly not a beat candidate; trailing 4-quarter surprise \textasciitilde{}+2\%. MarketBeat: consensus 'Reduce' rating (19 brokerages, 7 sells). Businesswire Aug 13, 2026 reconfirms Sept 3, 2026 report date; no pre-announcement or late analyst shock found in August 2026.\par
\noindent **[Segment 3]** **[Segment 3]** Final pre-report check (one day before Sept 3, 2026 report): WSJ shows Q4 2026 estimate trending down from \$0.41 (3 months ago) to \$0.39 (current); Yahoo current estimate \$0.39 with 18 analysts; Zacks preview reiterates \$0.40 consensus and negative beat candidacy; Evercore cut PT to \$22 (from \$23) as of \textasciitilde{}Aug 31; JPMorgan raised PT to \$22 on Aug 18. No pre-announcement, no last-minute estimate pops. Picture unchanged: threshold requires $\geq$\$0.40 with consensus right at \$0.39-\$0.40 and negative ESP.\par
\vspace{1.5pt}
\noindent \#\# To-do (remaining search directions)\par
\noindent (none)\par
\end{msgbox}
\begin{msgbox}{casegreen}
\casetag{Thinking}
\noindent The notebook is complete. All to-do items are exhausted. The report is tomorrow (Sept 3, 2026), and I've confirmed:\par
\noindent - Consensus at \$0.39 (WSJ/Yahoo/Public.com) to \$0.40 (Zacks)\par
\noindent - Threshold requires $\geq$ \$0.40\par
\noindent - Zacks ESP -4.22\%, Rank \#4 Sell, not a beat candidate\par
\noindent - Estimates revised down from \$0.41 to \$0.39 over 3 months\par
\noindent - Offsetting: guidance midpoint \textasciitilde{}\$0.42, trailing +2\% surprise, 2 of last 3 quarters beat\par
\vspace{1.5pt}
\noindent Final probability: 0.34 YES / 0.66 NO.\par
\vspace{1.5pt}
\noindent The to-do list is empty, so I should give my final answer now. The final line must be exactly \textbackslash{}boxed\{0.34\}.\par
\end{msgbox}
\begin{msgbox}{casegreen}
\casetag{Response}
\noindent Based on my completed research, all key directions have been explored and the notebook is current as of one day before the report. The evidence picture is stable and well-established:\par
\vspace{1.5pt}
\noindent **Key findings:**\par
\noindent - **Consensus sits exactly at the threshold**: \$0.39 (WSJ/Yahoo/Public.com) to \$0.40 (Zacks), meaning the market requires an actual result of $\geq$ \$0.40 to resolve YES\par
\noindent - **Bearish signals dominate**: Zacks Earnings ESP of -4.22\% (Most Accurate Estimate \textasciitilde{}\$0.383), Zacks Rank \#4 (Sell), explicit "not a compelling beat candidate" assessment, downward estimate revisions from \$0.41 $\rightarrow$ \$0.39 over 3 months, and expected revenue -7.3\% YoY\par
\noindent - **Partial offsets**: FY26 guidance midpoint implies \textasciitilde{}\$0.42 for Q4, trailing 4-quarter surprise of +2\%, beats in 2 of the last 3 quarters, and JPMorgan's PT raise to \$22 (though Evercore cut to \$22 one day ago)\par
\noindent - **Timing confirmed** for Sept 3, 2026 with no pre-announcement, so resolution-timing risk is minimal\par
\vspace{1.5pt}
\noindent **Probabilistic reasoning:** With consensus centered at \$0.39-\$0.40 and a realistic earnings std of \textasciitilde{}\$0.025-\$0.03, the probability of landing at or above \$0.40 is roughly 0.33-0.35. The negative ESP and downward revision drift push toward the lower end, while the guidance midpoint and recent beat frequency provide some counterweight.\par
\vspace{1.5pt}
\noindent Final belief: YES 0.34 / NO 0.66.\par
\vspace{1.5pt}
\noindent \textbackslash{}boxed\{0.34\}\par
\end{msgbox}